Habitation à Diriger des Recherches

# Contributions to Label-Efficient Learning in Computer Vision and Remote Sensing

Presented by

## Minh-Tan PHAM

Maître de Conférences at Université Bretagne Sud, UMR IRISA 6074

Defended on **June 6, 2025**. Committee:

| | |
|---|---|
| Céline HUDELOT, Full Professor at CentraleSupélec, France | Reviewer |
| Jocelyn CHANUSSOT, Research Director at INRIA, France | Reviewer |
| Begüm DEMIR, Full Professor at Technische Universität Berlin, Germany | Reviewer |
| Alexandre BENOIT, Full Professor at Université Savoie Mont Blanc, France | Examinator |
| Olivier LÉZORAY, Full Professor at Université de Caen Normandie, France | Examinator |
| Sébastien LEFÈVRE, Full Professor at Université Bretagne Sud, France | Guarantor |

# Abstract

This manuscript presents a series of my selected contributions to the topic of label-efficient learning in computer vision and remote sensing. The central focus of this research is to develop and adapt methods that can learn effectively from limited or partially annotated data, and can leverage abundant unlabeled data in real-world applications. The contributions span both methodological developments and domain-specific adaptations, in particular addressing challenges unique to Earth observation data such as multi-modality, spatial resolution variability, and scene heterogeneity. The manuscript is organized around four main axes including (1) weakly supervised learning for object discovery and detection based on anomaly-aware representations learned from large amounts of background images; (2) multi-task learning that jointly trains on multiple datasets with disjoint annotations to improve performance on object detection and semantic segmentation; (3) self-supervised and supervised contrastive learning with multimodal data to enhance scene classification in remote sensing; and (4) few-shot learning for hierarchical scene classification using both explicit and implicit modeling of class hierarchies. These contributions are supported by extensive experimental results across natural and remote sensing datasets, reflecting the outcomes of several collaborative research projects. The manuscript concludes by outlining ongoing and future research directions focused on scaling and enhancing label-efficient learning for real-world applications.

# Contents

# Chapter 1

# Introduction

## 1.1 Context and motivations

Over the past decade, deep learning has rapidly emerged, achieving state-of-the-art performance across a wide range of applications. In computer vision and remote sensing for Earth observation, deep learning models have demonstrated remarkable success in pivotal tasks such as image classification, object detection, semantic segmentation, change detection and super-resolution (Dimitrovski et al., 2023; Tuia et al., 2024; X. X. Zhu et al., 2017). These advances have unlocked new potentials in critical research areas in Earth and environment observation including forest and ocean monitoring, urban planning, agriculture, and disaster management (Persello et al., 2022). However, the success of deep learning hinges on the availability of large labeled datasets, which are expensive and time-consuming to build. This is particularly challenging in the domain of Earth observation, where annotation requires expert knowledge and is often hampered by the complexity and variability of remote sensing data (Rolf et al., 2024; Schmitt et al., 2023).

The need for extensive labeled data has motivated research in label-efficient learning techniques, which aim to mitigate the dependence on large annotated datasets while maintaining or even enhancing model performance. These techniques include methods such as semi-supervised learning, weakly supervised learning, self-supervised learning, multi-task learning, active learning and few/zero-shot learning (Hosseiny et al., 2023; Sarker, 2021). In computer vision, these approaches have already shown considerable potential, allowing models to learn effectively from limited labeled data by leveraging vast amounts of unlabeled data or pre-existing knowledge (Chai et al., 2021). In remote sensing for Earth observation, where huge amounts of data are collected from various sensors under varying conditions, label-efficient learning approaches can be crucial to overcome the inherent challenges associated with label scarcity. These techniques enable more scalable, cost-effective, and adaptable solutions for critical applications, including land cover mapping, change detection, and disaster monitoring (Hosseiny et al., 2023).

In this manuscript of Habilitation à Diriger des Recherches (HDR), I present a series of my selected contributions to the topic of label-efficient learning in computer vision and remote sensing over the past few years. Some of these research studies focus on developing novel architectures and frameworks that allow us to learn from limited labeled samples while taking full advantage of the abundant unlabeled or weakly-labeled data available in these domains. Others include adapting state-of-the-art label-efficient methods to the unique characteristics of Earth observation data, such as multi-modality, spatial resolution variability and scene heterogeneity. As a result, most of my research presented in this document seek to contribute novel methodologies that enhance the performance and efficiency of general computer vision tasks such as image classification, object detection and semantic segmentation, under the context of limited or partially annotated data. In addition, they aim to foster the scalability and robustness of deep learning models in real-world remote sensing applications including land-cover mapping, animal discovery and monitoring, wildfire delineation and methane source recognition.

## 1.2   Overview of featured contributions

In this manuscript, I focus on **four key research axes** related to label-efficient learning applied to both computer vision and remote sensing. Each of them will be presented in a chapter, outlined as follows.

**Chapter 2**  presents our research related to **weakly supervised learning for object discovery**. The key idea is to leverage a huge amount of empty image data to learn representations of "normal data" within a latent space. Then, objects and regions of interest will be detected and localized due to their "anomalous" representations based on some predefined distance metrics. The work covered in this chapter was published in two journal papers and a series of four conference papers. Most of them were conducted during the postdoc of Hugo Gangloff that I co-supervised with Sébastien Lefèvre (2021-2022). The others were done during the Master internships of Paul Berg (2021) and Oscar Narvaez-Luces (2023).

**Chapter 3**  investigates the **multi-task partially supervised learning from multiple datasets**. Motivated from the fact that small datasets annotated for their own task could semantically share complementary information to other tasks in a joint learning framework, we proposed a single model to learn object detection and semantic segmentation from multiple datasets where each training example is labeled for a single task. This work was done during the postdoc of Hoàng-Ân Lê supervised by myself within the ROMMEO (Robust multi-task learning with mutual knowledge distillation for Earth observation) research project (2022-2024). The outcomes were a series of five papers published in both computer vision and remote sensing venues.

**Chapter 4**  is devoted to our study on **self-supervised learning with multimodal image data** applied to remote sensing scene classification. In this work, we first implemented joint self-supervised pre-training framework to learn representations from unlabeled remote sensing images, from mono-modality to multi-modality scenarios. We then proposed a multimodal supervised contrastive loss, allowing us to better perform remote sensing scene classification task applied to methane source recognition. This work was published in two journal papers and two conferences papers during the PhD thesis of Paul Berg that I co-supervised with Nicolas Courty (2021-2024).

**Chapter 5**  is dedicated to our research on **few-shot learning with hierarchical image data** applied to scene classification in remote sensing. Built upon the prototypical network for few-shot learning, we introduced two frameworks to tackle few-shot scene classification. The first one explicitly integrated the class hierarchy to build hierarchical prototypes. The second implicitly discovered such information within a non-Euclidean space. These were done within the PhD thesis of Manal Hazaoui that I co-supervised with Laetitia Chapel and Sébastien Lefèvre (2019-2023), with a series of three publications including one journal and two conference papers.

For each chapter, I will highlight our motivation, describe methodological developments and provide experimental results on natural and remote sensing images. Finally, **Chapter 6** presents my ongoing and future research directions, from short- to mid-term perspectives.

## 1.3   Other contributions

In addition to the four key directions outlined above, this section briefly presents my contributions and research activities (after my PhD defense in 2016) that will be not detailed in this manuscript to favor conciseness and consistence. They are regrouped into four following topics.

**Multilevel image analysis using morphological attributes profiles**   Beside the main research projects related to deep learning applied to computer vision and remote sensing tasks, I conducted methodological studies on multilevel image analysis and processing based on tree-based image representation and morphological attribute profiles (APs). Some contributions were published during my postdoc (2016-2019) including the local feature-based APs **(J07)**, **(C08)**, the feature profiles **(J08)**, **(C12)** and the texture-derived APs **(J09)**. Most of them were proposed to improve the performance of the standard APs (Dalla Mura et al., 2010) and the histogram-based APs (Demir and Bruzzone, 2015) by considering statistical and texture information of each pixel's local patch within the tree-based filtering, in particular when dealing with highly textured remote sensing images. Then, during the postdoc of Deise Santana Maia that I co-supervised with Sébastien Lefèvre (2019-2021), we continued to develop the watershed attribute profiles by generalizing the tree-based AP paradigm to the hierarchical watersheds with semantic prior knowledge **(J16)**, **(C20)**. Finally, we conducted a survey paper on a decade of advances in APs and their extensions applied to remote sensing image classification **(J14)**.

**Small object detection under complex backgrounds**   During my postdoc working in the ANR DEEPDETECT project (2018-2019), I proposed the YOLO-fine framework to detect very small objects from aerial and satellite images under various backgrounds within operational contexts **(J12)**, **(C16)**. With only 7% the size of the baseline which was YOLOv3 (a state-of-the-art one-stage detector in 2018), the proposed YOLO-fine achieved much better performance on multi-class small vehicle detection from aerial and satellite images. As another approach, we then leveraged super-resolution as an auxiliary task to assist small object detection in **(J13)**, **(C18)**. In these studies, we proposed to focus the super-resolution only on objects of interest and not on the entire image from which most of the scene involve heterogeneous backgrounds. Recently, we have adopted a mutual guidance strategy coupling with contrastive learning **(J18)** as well as a multimodal fusion approach **(C26)** **(J25)** to tackle small object detection from remote sensing scenes. Then, during the Master internship of Anh-Kiet Duong (2024) supervised by myself in collaboration with the Luxembourg Institute of Science and Technology (LIST), we leveraged multi-source satellite image data from medium- to high-resolution sensors to perform deep learning-based small ship detection **(J22)**, **(C37)**, **(C41)**.

**Deep learning applied to non-visual modality**   Throughout different research projects, I contributed to develop and adapt various deep learning models specific to different sources of non-visual (non-RGB) remote sensing data such as ground penetrated radar (GPR) data **(C12)**, passive acoustic data **(J24)**, polarimetric radar images **(J10)**, **(C19)**, Lidar point clouds **(J11)** and digital terrain models (DTM) **(J17)**. For the specific case of GPR which was the main data source explored in the DELORA project (2016-2018), I proposed to combine simulated radargrams (using the gprMax software) with those generated from the real acquired B-scan GPR data to improve the performance of buried object detection using the Faster R-CNN detector **(C11)**. This work has become one of the tailored studies on deep learning applied to this data type. For the polarimetric radar images, we proposed to exploit both polarimetric features from the coherency matrix combined with structural tensors to feed the segmentation deep network (SegNet) to perform pixelwise classification **(J10)**, **(C19)**. This helped to significantly improve the performance, in particular on heterogeneous texture areas. Regarding the 3D Lidar point clouds, by leveraging various rasterization techniques, we trained deep models to address semantic segmentation task **(J11)**, as well as to perform DTM extraction from airbone laser scanning (ALS) point clouds **(J17)**. To promote the data-driven DTM extraction using deep learning, we collected and released the large-scale ALS2DTM dataset in **(J17)**.

**Learning with Optimal transport**   Within the PhD thesis of Paul Berg co-supervised by Nicolas Courty and myself (2021-2024), I had the opportunity to work on the topic of optimal transport for representation learning with two contributions in the context of the OT-TOPIA project. First, we leveraged the Gromov-Wasserstein distance and class hierarchy to perform smart initialization and optimization of prototypes within hyperbolic spaces, published in **(C40)**. Such an approach allows us to better learn data representations with low-dimensional embedding to tackle computer vision tasks such as classification and segmentation of natural images and 3D point clouds. Then, we have extended this work by implementing a multi-prototype version to generalize the prototypical learning guided by class hierarchy within hyperbolic spaces **(J26)**. Regarding the second contribution, built upon the work of contrastive representation learning through alignment and uniformity on the hypersphere (T. Wang and Isola, 2020), we leveraged the Spherical Sliced-Wasserstein (SSW) distance to perform the uniformity objective in **(C29)**. The proposed loss allowed us to preserve maximal information of the feature distribution on the hypersphere, avoiding the collapsing representations in self-supervised learning.

# Chapter 2

# Weakly supervised learning for object discovery

While supervised learning is effective, it often requires a substantial amount of human-labeled data to achieve high accuracy. In contrast, unsupervised learning generally tends to underperform due to the absence of labeled data to guide the learning process. In this context, weakly supervised learning offers a promising middle ground by utilizing limited or noisy labels, thereby reducing the dependency on extensive manual annotations and potentially bridging the performance gap between fully supervised and unsupervised methods. This chapter is dedicated to the task of **weakly supervised object discovery** based on image-level annotation, *i.e., empty* vs *non-empty* images. Such an approach is highly relevant in the context of scarce object detection from images where the objects of interest are often missing, hard to annotate and of varying aspects, *e.g.,* animal detection from aerial surveys from which more than 95% of captured images are empty.

Under the computer vision point of view, such a task is similar to the **unsupervised anomaly detection (AD)** task where deep networks are first trained only on empty images (*i.e.,* normal data), then able to detect and localize objects as anomalies at inference time. Indeed, anomalies are often unknown in advance, hence it is impossible to gather labeled anomalous data to train the deep model. From now on in this manuscript, the two terms unsupervised AD and weakly supervised object discovery are used indifferently. Following (L. Wang et al., 2020), there are three main approaches (which are often combined in practice) for unsupervised AD, including (1) feature extraction-based, relying on a distance metric defined in the feature space; (2) probability-based, making use of distributions or statistical tests to detect anomalies; and (3) reconstruction-based, computing the distance between input and reconstructed data produced by a network. Among them, reconstruction-based methods appear to be the most explored ones in the literature. They are often based on deep generative latent variable models such as Variational Autoencoders (VAE), Generative Adversarial Networks (GAN) or diffusion models (X. Zhang et al., 2023).

In this chapter[1], a focus will be given on the VAE-based unsupervised AD with our **two main contributions**. The first one involves *a VAE model with Gaussian Random Fields prior (VAE-GRF)*, described in Section 2.2. The second contribution is dedicated to *a Vector-Quantized VAE (VQ-VAE) model with discrete latent space*, presented in Section 2.3. It should be noted that we also conducted another research study on weakly supervised detection of marine animals from aerial images using the feature extraction-based approach (J15), which will not be described in this document. Before going to the details of our propositions, let us first remind some backgrounds and related works in VAE-based AD in Section 2.1.

---

[1]This chapter is built upon the research presented in the five articles (C21), (C28), (C30), (C31) and (J23) during the postdoc of Hugo Gangloff (2021-2022) and the Master internship of Oscar Narvaez Luces (2023).

## 2.1 VAE-based anomaly detection

In the unsupervised and weakly supervised AD contexts, deep generative models are a popular choice (Bond-Taylor et al., 2021). Among these models, the particular deep latent variable models called VAEs have been widely used. They are defined in a probabilistic framework detailed in (Kingma and Welling, 2019). In a nutshell, VAEs transform an input image $\boldsymbol{x}$ into a compressed representation $\boldsymbol{z}$ through a stochastic encoder network $q_{\boldsymbol{\varphi}}(\boldsymbol{z}|\boldsymbol{x})$. The image is then reconstructed to form the reconstruction $\hat{\boldsymbol{x}}$, by sampling from a stochastic decoder network $p_{\boldsymbol{\theta}}(\boldsymbol{x}|\boldsymbol{z})$. The model is trained, for both network parameters $\boldsymbol{\varphi}$ and $\boldsymbol{\theta}$, by maximizing the Evidential Lower Bound (ELBO) which reads:

$$\mathcal{L}_{\boldsymbol{\theta},\boldsymbol{\varphi}}(\boldsymbol{x}) = \underbrace{\mathbb{E}_{q_{\boldsymbol{\varphi}}(\boldsymbol{z}|\boldsymbol{x})}[\log p_{\boldsymbol{\theta}}(\boldsymbol{x}|\boldsymbol{z})]}_{\text{reconstruction term}} - \underbrace{\mathrm{KL}(q_{\boldsymbol{\varphi}}(\boldsymbol{z}|\boldsymbol{x})||p_{\boldsymbol{\theta}}(\boldsymbol{z}))}_{\text{regularization term}}. \tag{2.1}$$

Intuitively, the first term encourages the reconstruction $\hat{\boldsymbol{x}}$ to be similar to the input $\boldsymbol{x}$ with the constraint of a regularization term under the form of Kullback-Leibler (KL) divergence. It can be seen that VAEs give access to several metrics computed from the continuous latent space and the reconstructions. Such metrics form the foundation of many image-wise and pixel-wise AD approaches. Some of these metrics are, for example: the residual images between $\boldsymbol{x}$ and $\hat{\boldsymbol{x}}$, the evaluation of the reconstruction term and/or of the KL term of Eq. (2.1), the derivative of Eq. (2.1) with respect to $\boldsymbol{x}$, *etc.* Many combinations of these ideas have been successfully applied to unsupervised AD giving rise to an important number of recent studies such as (Baur et al., 2021; W. Liu et al., 2020; Venkataramanan et al., 2020). In Figure 2.1, we illustrate the classical VAE and the proposed VAE-GRF model (which will be described in the next section) with an encoder-decoder architecture.

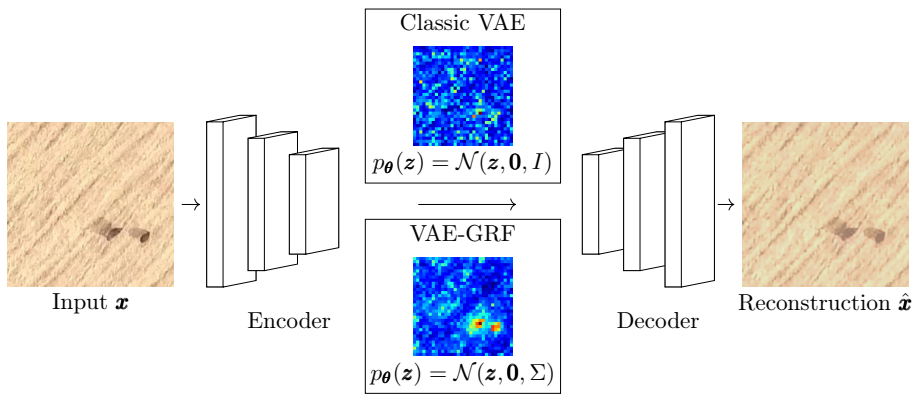

**Figure 2.1:** The classical VAE and the VAE-GRF architectures. We illustrate the pixel-wise $MAD$ (Mean Absolute Deviation) metric computed from the latent space in both cases.

**VAEs for anomaly detection**    In the context of unsupervised AD, VAE-like models are classically trained on a dataset of normal samples to learn the latent representation of the normality. Then, anomalies are detected at inference time using a combination of an intrinsic metric defined in the latent space, coupled with a classical reconstruction-based metric.

Let $MAD$ (Mean Absolute Deviation) be the $n_x \times n_y$ anomaly map that we compute from the latent space. Each pixel of the anomaly map is computed as the mean absolute deviation from the mean of the same location of the output of the convolutional encoder $\boldsymbol{z}$, i.e., $\forall x \in \{1, \ldots, n_x\}, \forall y \in \{1, \ldots, n_y\}$,

$$MAD_{x,y} = \frac{1}{n_z} \sum_{k=1}^{n_z} \left| z_{x,y,k} - \frac{1}{n_z} \sum_{k=1}^{n_z} z_{x,y,k} \right|. \tag{2.2}$$

Let us also remind the reconstruction-based anomaly map, called $SM$, which uses the Structural Similarity Index Measure (SSIM) (Z. Wang et al., 2004). For each pixel $i$:

$$SM(x_i) = \text{SSIM}(\boldsymbol{p}_i, \boldsymbol{q}_i) = \frac{(2\mu_{\boldsymbol{p}}\mu_{\boldsymbol{q}} + c_1)(2\sigma_{\boldsymbol{pq}} + c_2)}{(\mu_{\boldsymbol{p}}^2 + \mu_{\boldsymbol{q}}^2 + c_1)(\sigma_{\boldsymbol{p}}^2 + \sigma_{\boldsymbol{q}}^2 + c_2)}, \tag{2.3}$$

where $\boldsymbol{p}_i$ (resp. $\boldsymbol{q}_i$) is a patch around pixel $i$ of $\boldsymbol{x}$ (resp. $\hat{\boldsymbol{x}}$). $\mu_p$, $\sigma_p$ and $\sigma_{pq}$ represent, respectively, the mean, the standard deviation and the covariance of the patches. The scalars are set to $c_1 = 0.01$ and $c_2 = 0.03$ according to the original paper.

Finally, a combined metric can be formed based on the two previous anomaly maps using an element-wise multiplication, denoted as $MAD \odot SM$. To do that, $MAD$ is first upsampled to the image dimension.

## 2.2 VAE with Gaussian Random Fields prior

### 2.2.1 Model definition

The VAE-GRF model we propose is first composed of a stochastic encoding network, with input $\boldsymbol{x}$, which maps to a convolutional latent space associated to the realizations of a random variable $\boldsymbol{z}$, following the ideas from (Venkataramanan et al., 2020). In this context, $\boldsymbol{z}$ has dimension $N = n_x \times n_y \times n_z$ (width×height×depth). The outputs of the encoder, $L = \text{diag}(\sigma_1^2, \ldots, \sigma_N^2)$ and $\boldsymbol{m} \in \mathbb{R}^N$, parameterize a variational posterior distribution, $q_{\boldsymbol{\varphi}}(\boldsymbol{z}|\boldsymbol{x})$, chosen as independent Gaussian random variables; we then have $q_{\boldsymbol{\varphi}}(\boldsymbol{z}|\boldsymbol{x}) = \mathcal{N}(\boldsymbol{z}; \boldsymbol{m}, L)$. For a reason that will be clarified later, we factorize on the depth dimension such that $q_{\boldsymbol{\varphi}}(\boldsymbol{z}|\boldsymbol{x}) = \prod_{k=1}^{n_z} \mathcal{N}(\boldsymbol{z}_k; \boldsymbol{m}_k, L_k)$, and then $L_k = \text{diag}((\sigma_k^2)_1, \ldots, (\sigma_k^2)_{n_x n_y})$. The model is then composed of a stochastic decoder network whose output $\boldsymbol{\lambda} = (\lambda_1, \ldots, \lambda_{l_x l_y})$ parametrizes a product of independent Continuous Bernoulli random variables (Loaiza-Ganem and Cunningham, 2019), $p_{\boldsymbol{\theta}}(\boldsymbol{x}|\boldsymbol{z}) = \prod_{k=1}^{l_x l_y} \mathcal{CB}(x_k, \lambda_k)$. A realization of this stochastic decoder corresponds to a reconstruction, denoted $\hat{\boldsymbol{x}}$, of the input image $\boldsymbol{x}$ by the model.

While traditional VAEs consider a Gaussian prior with diagonal covariance matrix, for the GRF prior we relax this assumption. We consider a zero-mean stationary and toroidal GRF prior on the $n_x \times n_y$ dimension and we consider the components of $\boldsymbol{z}$ to remain independent on the depth dimension $n_z$. Hence $p_{\boldsymbol{\theta}}(\boldsymbol{z}) = \mathcal{N}(\boldsymbol{z}; \boldsymbol{0}, \Sigma) = \prod_{k=1}^{n_z} \mathcal{N}(\boldsymbol{z}_k; \boldsymbol{0}, \Sigma_k)$, where $\Sigma_k$ are Symmetric Positive Definite (SPD) matrices, and the computations can still be done in a parallel manner on this dimension because the computations fall back to $n_z$ parallel computations involving bi-dimensional GRFs which shared parameters. We go one step further by sharing the parameters between these GRFs, thus, $p_{\boldsymbol{\theta}}(\boldsymbol{z}) = \prod_{k=1}^{n_z} \mathcal{N}(\boldsymbol{z}_k; \boldsymbol{0}, \Sigma)$. Note that such a parameter sharing is also proposed in the one dimensional case of the factorized Gaussian Process VAEs of (Jazbec et al., 2020). Note also that, since the diagonal covariance matrix used in the standardized Gaussian prior in the classical VAE model is comprised in the set of the SPD matrices yielding a GRF, the VAE-GRF model is a strict generalization of the classical VAE model.

A critical point of the model resides in the estimation of the covariance matrix $\Sigma$ for the more complex prior we introduce. This estimation needs to yield a SPD matrix. To ensure this requirement, the covariance matrix $\Sigma$ is assumed to be generated by a Matern correlation function (Y. Liu et al., 2019). Such a GRF with Matern correlation function has parameter $\bar{\boldsymbol{\theta}} = \{\bar{r}, \bar{\sigma}^2\}$ (correlation range and variance), which needs to be learned. Thus, the GRF assumption over the input image yields a log-likelihood of the form:

$$\log p_{\bar{\boldsymbol{\theta}}}(\boldsymbol{x}) = -\frac{l_x l_y}{2} \log 2\pi - \frac{1}{2} \log|\bar{\Sigma}| - \frac{1}{2}\boldsymbol{x}^T \bar{\Sigma}^{-1} \boldsymbol{x}. \tag{2.4}$$

**Training the model** We make use of the $\beta$-ELBO (Higgins et al., 2017), with an additional stop gradient operator, which reads

$$\mathcal{E}_{\boldsymbol{\theta},\boldsymbol{\varphi},\beta}(\boldsymbol{x}) = \mathbb{E}_{q_{\boldsymbol{\varphi}}(\boldsymbol{z}|\boldsymbol{x})}[\log p_{\boldsymbol{\theta}}(\boldsymbol{x}|\boldsymbol{z})] - \beta \mathrm{KL}(q_{\boldsymbol{\varphi}}(\boldsymbol{z}|\boldsymbol{x})||\mathrm{sg}[p_{\boldsymbol{\theta}}(\boldsymbol{z})]), \tag{2.5}$$

where $\mathrm{sg}$ refers to the stop gradient operator and $\mathrm{KL}$ refers to the Kullback-Leibler divergence. The stop gradient operator has the role of preventing the KL term to drag $\Sigma$ towards $L$, a Gaussian distribution with diagonal covariance matrix. Then, the VAE-GRF is trained by maximizing the $\beta$-ELBO (with stop gradient) plus the log-likelihood prior over the observed image. The total loss is then defined by:

$$\mathcal{L}_{\boldsymbol{\theta},\boldsymbol{\varphi}}(\boldsymbol{x}) = \mathcal{E}_{\boldsymbol{\theta},\boldsymbol{\varphi},\beta}(\boldsymbol{x}) + \log p_{\bar{\boldsymbol{\theta}}}(\boldsymbol{x}). \tag{2.6}$$

It is noted that the choice of the correlation function and the hyperparameter $\beta$ is empirically made in our study. Fig. 2.1 illustrates the model and $MAD$ anomaly map computed within the latent space. The encoder-decoder architecture is identical to that of a classical VAE. In this example, the proposed VAE-GRF yields a remarkable improvement in the anomaly map, only based on an original refinement of the probabilistic model.

## 2.2.2 Experiments

To evaluate the VAE-GRF model, we consider comparable generative VAE-based models which also rely on model refinement techniques from the literature. That is why, we compare our $MAD$ and $MAD \odot SM$ metrics to the already existing $\ell$-2 and $SM$ metrics introduced in (Bergmann et al., 2019). An interesting idea quite similar to ours is the Visually Explained VAE (VEVAE) proposed by (W. Liu et al., 2020), which will be used as baseline for comparison. AD metrics similar to ours have also been introduced in (Zimmerer et al., 2019). They correspond to the computation $|\mathcal{E}_{\boldsymbol{\theta},\boldsymbol{\varphi}}(\boldsymbol{x})| \odot SM$. We also test the iterative procedure proposed by (Dehaene et al., 2020). It consists in a refinement of the VAE reconstruction. Regarding the network architecture, we adopt the similar encoder and decoder as proposed in (W. Liu et al., 2020), to ensure a fair comparison.

**Experiments on computer vision datasets** We exploit the popular MVTec dataset with normal and abnormal (defective) RGB images of industrial goods from 15 different categories (Bergmann et al., 2019). It is known as one of the standard datasets to benchmark AD methods in computer vision. Table 2.1a summarizes the results in terms of pixel-wise ROCAUC (Receiver Operating Characteristic/Area Under the Curve), computed with the final anomaly map and the ground truth. We observe that on the images which respect the most the stationary GRF assumption, *i.e.* texture images, the VAE-GRF performs better, especially when it comes to the new metrics we introduced (up to $8\%$ of improvement on the *Tile* texture). On images where the hypothesis of a stationary GRF clearly does not hold, such as *Carpet*, *Grid* and *Hazelnut* possibly because the stationary GRF assumption is clearly wrong, the $MAD$ metric seems to become useless. Moreover, it can be observed that, on such images, the VAE-GRF results fall back to the results of the classical VAE. Thus, we can see the interest of the refinement in VAE modeling and its limitations when the stationary GRF prior assumption is violated. Next, as shown in Table 2.1b, the VAE-GRF provides always on par or better results than the three state-of-the-art methods, with the best gains coming from the texture images. For more qualitative results and the sensitivity analysis of the hyperparameter $\beta$, readers are referred to our articles (J23), (C30).

| Category | | VAE | | | Our VAE-GRF | | |
|---|---|---|---|---|---|---|---|
| | | $SM$ | $MAD$ | $MAD \odot SM$ | $SM$ | $MAD$ | $MAD \odot SM$ |
| Stationary textures | Leather ($\beta = 100$) | 0.97 | 0.92 | 0.94 | 0.95 | 0.95 | **0.98** |
| | Tile ($\beta = 0.01$) | 0.93 | 0.69 | 0.93 | 0.93 | 0.87 | **0.95** |
| | Wood ($\beta = 100$) | 0.70 | 0.75 | 0.76 | 0.78 | **0.86** | 0.80 |
| Non stationary textures | Carpet ($\beta = 0.01$) | 0.91 | 0.60 | **0.92** | 0.90 | 0.61 | 0.91 |
| | Grid ($\beta = 0.01$) | 0.96 | 0.66 | 0.96 | 0.95 | 0.76 | **0.97** |
| | Hazelnut ($\beta = 1$) | 0.94 | 0.94 | 0.96 | **0.98** | 0.91 | **0.98** |

(a)

| | Category | Our VAE-GRF | (W. Liu et al., 2020) | (Zimmerer et al., 2019) | | (Dehaene et al., 2020) |
|---|---|---|---|---|---|---|
| | | $MAD \odot SM$ | VEVAE | $\lvert \nabla_{\boldsymbol{x}} \mathcal{E}_{\boldsymbol{\theta,\varphi}}(\boldsymbol{x}) \rvert$ | $\lvert \nabla_{\boldsymbol{x}} \mathcal{E}_{\boldsymbol{\theta,\varphi}}(\boldsymbol{x}) \rvert \odot SM$ | $SM$ grad |
| Stationary textures | Leather | **0.98** | 0.95 | 0.55 | 0.94 | 0.95 |
| | Tile | **0.95** | 0.80 | 0.62 | 0.78 | 0.78 |
| | Wood | **0.80** | 0.77 | 0.54 | 0.76 | 0.74 |
| Non stationary textures | Carpet | **0.91** | 0.78 | 0.61 | 0.89 | 0.89 |
| | Grid | **0.97** | 0.73 | 0.54 | 0.92 | 0.90 |
| | Hazelnut | **0.98** | **0.98** | 0.80 | **0.98** | 0.92 |

(b)

**Table 2.1:** ROCAUC scores for pixel-wise AD on the texture images from the MVTec dataset, provided by the proposed VAE-GRF compared to state-of-the-art methods. (a) VAE-GRF compared to the baseline VAE using different AD metrics. (b) VAE-GRF compared to three state-of-the-art methods. Best scores appear in bold.

**Experiments on remote sensing datasets** We conduct experiments on weakly supervised animal discovery using two remote sensing datasets. The first one is the Livestock dataset (Han et al., 2019) which regroups aerial images of livestock over grassland areas. It contains a total of $3430$ empty images for training and $890$ images containing at least one animal for testing. The second one is the Semmacape dataset introduced in (Berg, Santana Maia, et al., 2022) for marine animal detection from aerial surveys including $345$ images with animals and about $138K$ empty images. Such a large number of empty images enables us to learn the normality of the sea surface and then to detect animals as anomalies. The pixel-wise ROCAUC results for Livestock are shown in Table 2.2a. Again, the improvements brought by the GRF prior can be expected since the images really fit the stationary GRF hypothesis. Our model provides better scores compared to the classical VAE, as well as two other reference methods. For Semmacape, Table 2.2b gives the results in terms of precision, recall and F1 score. Because of the dataset complexity (many images are corrupted by the sun glare), the $SM$ anomaly map seems particularly unreliable. Among all models, the VAE-GRF with $MAD$ yields the best performance, probably by using information from the latent space. It is thus less insensitive to the reconstructed images, which are of rather poor quality because of the noise corruption (a low $SM$ metric was observed in practice).

|  | VAE | Our VAE-GRF | (Zimmerer et al., 2019) | | (Dehaene et al., 2020) |
|---|---|---|---|---|---|
|  | $MAD \odot SM$ | $MAD \odot SM$ | $\lvert\nabla_{\boldsymbol{x}}\mathcal{E}_{\boldsymbol{\theta},\boldsymbol{\varphi}}(\boldsymbol{x})\rvert$ | $\lvert\nabla_{\boldsymbol{x}}\mathcal{E}_{\boldsymbol{\theta},\boldsymbol{\varphi}}(\boldsymbol{x})\rvert \odot SM$ | $SM$ grad |
| ROCAUC | 0.77 | **0.80** | 0.77 | 0.52 | 0.68 |

(a)

|  | VAE | Our VAE-GRF | (Zimmerer et al., 2019) | | (Dehaene et al., 2020) | (Berg, Santana Maia, et al., 2022) |
|---|---|---|---|---|---|---|
|  | $MAD$ | $MAD$ | $\lvert\nabla_{\boldsymbol{x}}\mathcal{E}_{\boldsymbol{\theta},\boldsymbol{\varphi}}(\boldsymbol{x})\rvert$ | $\lvert\nabla_{\boldsymbol{x}}\mathcal{E}_{\boldsymbol{\theta},\boldsymbol{\varphi}}(\boldsymbol{x})\rvert \odot SM$ | $SM$ grad | PaDiM+NF |
| F1 | 0.514 | **0.714** | 0.281 | 0.138 | 0.099 | 0.530 |
| Recall | 0.677 | 0.600 | 0.349 | 0.134 | 0.130 | **0.757** |
| Precision | 0.415 | **0.881** | 0.236 | 0.143 | 0.080 | 0.408 |

(b)

**Table 2.2:** AD detection scores for pixel-wise AD on the images from (a) the Livestock dataset and (b) the Semmacape dataset, provided by the proposed VAE-GRF compared to state-of-the-art methods. Best scores appear in bold.

## 2.3 Vector-Quantized VAE with codebook inner metrics

### 2.3.1 Proposed method

Our second contribution presented in this chapter is based on the VQ-VAEs with discrete latent variables, first introduced in (Van Den Oord and Vinyals, 2017), which have been proved to produce much sharper reconstruction than traditional VAEs. This makes VQ-VAEs interesting models for reconstruction-based AD by suggesting less noisy residual images. To cope with the discrete latent space, VQ-VAEs are trained differently from standard VAEs. In particular, the encoder becomes deterministic while the decoder remains stochastic. We now denote the encoder by $\text{Enc}_{\boldsymbol{\varphi}}$. Let $M$ be the number of possible states for the latent variable $z_k, \forall k \in \{1, \ldots, K\}$, $K$ being the dimension of the latent space. VQ-VAEs integrate a codebook, *i.e.*, a set of vectors $(e_1, \ldots, e_M)$, each one in $\mathbb{R}^D$, with $D$ a positive integer. From the encoder output, $\boldsymbol{z}_{\text{Enc}_{\boldsymbol{\varphi}}(\boldsymbol{x})}$, we choose the closest codebook vector for $z_k, \forall k$, following a deterministic decision, *i.e.*,

$$z_k = \underset{m \in \{1, \ldots, M\}}{\arg\min} \|(z_{\text{Enc}_{\boldsymbol{\varphi}}(\boldsymbol{x})})_k - e_m\|_2. \tag{2.7}$$

This can be associated to a deterministic and categorical encoding distribution as follows:

$$q_{\boldsymbol{\varphi}}(z_k = m | \boldsymbol{x}) = \begin{cases} 1 \text{ if } m = \underset{m \in \{1, \ldots, M\}}{\arg\min} \|(z_{\text{Enc}_{\boldsymbol{\varphi}}})_k - e_m\|_2, \\ 0 \text{ otherwise.} \end{cases} \tag{2.8}$$

The inputs of the decoder, denoted by $\boldsymbol{z}_{\text{Dec}_{\boldsymbol{\theta}}}$, are then also deterministically set as $(z_{\text{Dec}_{\boldsymbol{\theta}}})_k = e_{z_k}, \forall k$. Note that when processing images with convolutional encoders and decoders, the latent space $\boldsymbol{z}$ can also be convolutional, *i.e.*, it consists of a latent image.

The loss is then composed of the same reconstruction term as in standard VAEs but also of a squared $\ell 2$ term to ensure that the codebook is learnt. Due to the deterministic operations, the gradient can not flow from the decoder input to the encoder output: it has to be automatically copied; thus bypassing the codebook which cannot be updated. This second term is called the codebook *alignment term*. A third regularizing term is also added for stability of the training procedure, which leads to the VQ-VAE loss, for an image $\boldsymbol{x}$:

$$\mathcal{L}_{\boldsymbol{\theta}, \boldsymbol{\varphi}, \boldsymbol{e}}^{VQ-VAE}(\boldsymbol{x}) = \log p_{\boldsymbol{\theta}}(\boldsymbol{x} | \boldsymbol{z}_{\text{Dec}_{\boldsymbol{\theta}}(\boldsymbol{x})}) + \underbrace{\|\text{sg}[\boldsymbol{z}_{\text{Enc}_{\boldsymbol{\varphi}}(\boldsymbol{x})}] - \boldsymbol{e}\|_2^2}_{\text{alignment term}} + \beta \|\boldsymbol{z}_{\text{Enc}_{\boldsymbol{\varphi}}(\boldsymbol{x})} - \text{sg}[\boldsymbol{e}]\|_2^2, \tag{2.9}$$

with $\beta$ a scalar parameter and sg the stop gradient operator.

**Anomaly Detection with VQ-VAEs**    Similar to the metrics used for VAEs, we now define metrics for AD with VQ-VAEs. Besides the SSIM measure (denoted by $SM$) defined previously, we also define a new metric producing a latent space anomaly map, intrinsic to VQ-VAEs, which we call the Alignment Map ($AM$):

$$AM(\boldsymbol{x}) = \|\boldsymbol{z}_{\text{Enc}_{\boldsymbol{\varphi}}(\boldsymbol{x})} - \boldsymbol{e}\|_2^2. \tag{2.10}$$

The intuition behind the $AM$ anomaly map is as follows. During training time, the codebook vectors are trained to be close to the output of the encoder, and reciprocally, in virtue to the last two terms in Eq. (2.9). Therefore, at testing time, the anomalies (encoded in $\boldsymbol{z}_{\text{Enc}_{\boldsymbol{\varphi}}(\boldsymbol{x})}$), which have not been seen yet by the model, will be far from the codebook vectors, relatively to the normal features (also encoded in $\boldsymbol{z}_{\text{Enc}_{\boldsymbol{\varphi}}(\boldsymbol{x})}$). Since $AM$ is a small image with same dimension as the latent space where some pixels stand out, those pixels can be seen as *markers*. They correspond to the latent variables with high alignment loss, *i.e.*, anomalies. To be used in addition to the $SM$ anomaly map, the $AM$ is first upsampled. Then it undergoes a morphological grey dilation which aims at emphasizing the markers. However, none of these markers correctly represent the anomaly since they are just upsampled pixels. As previously proposed in AD with VAE and VAE-GRF when combining $MAD$ metric with $SM$, we calculate a more realistic shape for the anomaly by multiplying the $AM$ with the $SM$, denoted by $AM \odot SM$.

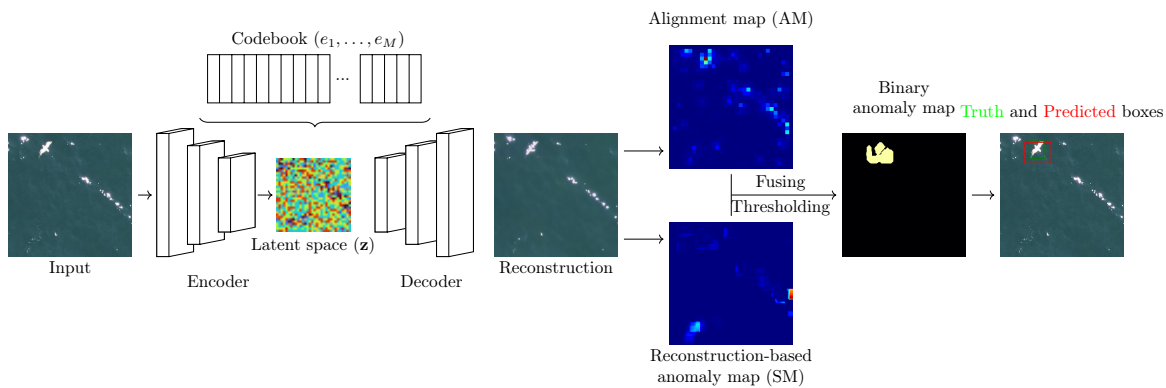

**Figure 2.2:** Overview of the proposed anomaly detection framework using VQ-VAE. In this figure, the framework is applied to weakly supervised marine animal detection from aerial images. Both AM and SM are thresholded before being fused to produce the final anomaly map.

### 2.3.2 Experiments

The VQ-VAE architecture used in our experiments is mainly based on the original VQ-VAE architecture implemented in (Van Den Oord and Vinyals, 2017). The codebook size is set to $M = 512$ and each vector of the codebook has dimension $D = 256$. For more details about the model setting, readers are referred to our papers (C21), (C28). To perform a fair comparative study, our VQ-VAE model is tested against comparable approaches, *i.e.*, those which are designed in a similar way. We now present our experiments conducted on both computer vision and remote sensing datasets.

**Experiments on vision datasets**    Table 2.3 shows the detection scores over all 15 texture categories of the MVTec dataset, yielded by different models. The proposed VQ-VAE with $SM \odot AM$ metric performs similarly as the FCDD method (Liznerski et al., 2020) which represents the state-of-the-art result on MVTec for this family of approaches. Our method also gives better results than AE (with $SM$) (Bergmann et al., 2019) and VEVAE (W. Liu et al., 2020). This suggests that VQ-VAE represents a more robust model than standard AEs and VAEs. VQ-VAE with only $SM$ metric also performs worse than with $SM \odot AM$, showing the interest of using the additional information provided by the proposed alignment map to improve the final anomaly map. More discussions about the AD performance on each texture category as well as some qualitative illustrations can be found in our paper (C21). In addition, experiments conducted on two other computer vision datasets including the CIFAR-10 and the UCSD-Ped1, done in (C21), also demonstrate that the proposed VQ-VAE can enhance anomaly detection results compared to other baselines.

**Experiments on remote sensing datasets**    We conduct experiments on two aerial image datasets dedicated to marine animal detection. One of them is the Semmacape dataset mentioned previously. The other one is the Kelonia dataset for marine turtle study, which contains 1983 aerial images with and without turtles. More details about both datasets can be found in (Berg, Santana Maia, et al., 2022). Table 2.2 shows better behavior of our VQ-VAE model on the two marine animal image datasets compared to three reference methods including PaDIM (Defard et al., 2021), OrthoAD (Kim et al., 2021) and PaDiM+NF (Berg, Santana Maia, et al., 2022). From the table, a high gain of F1-score is achieved for both datasets ($10\%$ to $12\%$ better than the second best method). It should be noted that since VQ-VAEs could provide highly accurate reconstructions, they reduce the noise in the residual images. Therefore, the number of false detections is reduced, which yields very high precision but may affect the recall, in particular on the Semmacape dataset.

| Category | AE | VEVAE | FCDD | Our VQ-VAE | |
| | (Bergmann et al., 2019) | (W. Liu et al., 2020) | (Liznerski et al., 2020) | $SM$ | $SM \odot AM$ |
|---|---|---|---|---|---|
| Carpet | 0.87 | 0.78 | **0.96** | 0.92 | 0.94 |
| Grid | 0.94 | 0.73 | 0.91 | **0.99** | **0.99** |
| Leather | 0.78 | 0.95 | **0.98** | **0.98** | **0.98** |
| Tile | 0.59 | 0.80 | **0.91** | 0.70 | 0.75 |
| Wood | 0.73 | 0.77 | **0.88** | 0.82 | 0.84 |
| Bottle | 0.93 | 0.87 | **0.97** | 0.94 | 0.95 |
| Cable | 0.82 | **0.90** | **0.90** | 0.87 | 0.87 |
| Capsule | **0.94** | 0.74 | 0.93 | 0.93 | **0.94** |
| Hazelnut | 0.97 | 0.98 | 0.95 | 0.98 | **0.99** |
| Metal Nut | 0.89 | 0.90 | **0.94** | 0.89 | 0.90 |
| Pill | **0.91** | 0.83 | 0.81 | 0.86 | 0.90 |
| Screw | 0.96 | 0.97 | 0.86 | **0.98** | **0.98** |
| Toothbrush | 0.92 | 0.94 | 0.94 | 0.96 | **0.97** |
| Transistor | 0.90 | **0.93** | 0.88 | 0.77 | 0.78 |
| Zipper | 0.88 | 0.78 | 0.92 | 0.97 | **0.98** |
| Mean | 0.86 | 0.86 | **0.92** | 0.90 | **0.92** |

**Table 2.3:** Performance in pixelwise ROCAUC on the MVTec dataset provided by the proposed VQ-VAE model compared to state-of-the-art methods. Best scores are marked in bold.

| Method | Semmacape | | | Kelonia | | |
| | F1-score | Recall | Precision | F1-score | Recall | Precision |
|---|---|---|---|---|---|---|
| PaDiM (Defard et al., 2021) | 0.383 | 0.434 | 0.343 | 0.504 | 0.443 | 0.586 |
| OrthoAD (Kim et al., 2021) | 0.458 | 0.373 | 0.594 | 0.571 | 0.514 | 0.643 |
| PaDiM+NF (Berg, Santana Maia, et al., 2022) | 0.530 | **0.757** | 0.408 | 0.568 | 0.559 | 0.578 |
| Our VQ-VAE | **0.636** | 0.497 | **0.884** | **0.726** | **0.754** | **0.701** |

**Table 2.4:** Performance in animal detection from two remote sensing datasets provided by the proposed VQ-VAE model compared to state-of-the-art methods. Best scores are in bold.

**Application to unsupervised burnt area extraction** As another use-case, we adopt the VQ-VAE framework to perform burnt area extraction from high resolution satellite images in our work (C31). The entire framework is depicted in Figure 2.3, from which burnt regions appear as anomalous areas. We remind that during the training phase, only images without burnt regions (usually *pre-event* images) are fed into the network. In addition, the proposed framework also involves an intensive post-processing step using some indexes related to vegetation, water and brightness to filter and remove false positives. Our experiments conducted on SPOT-6/7 images show that the proposed inner metric of VQ-VAE (based on the alignment map) has a huge potential for this challenging task. Nevertheless, further investigations should be devoted to improve the performance of such an approach, particularly in the scenario where training and test images exhibit significant domain shifts since they come from different sensors or different geographical areas.

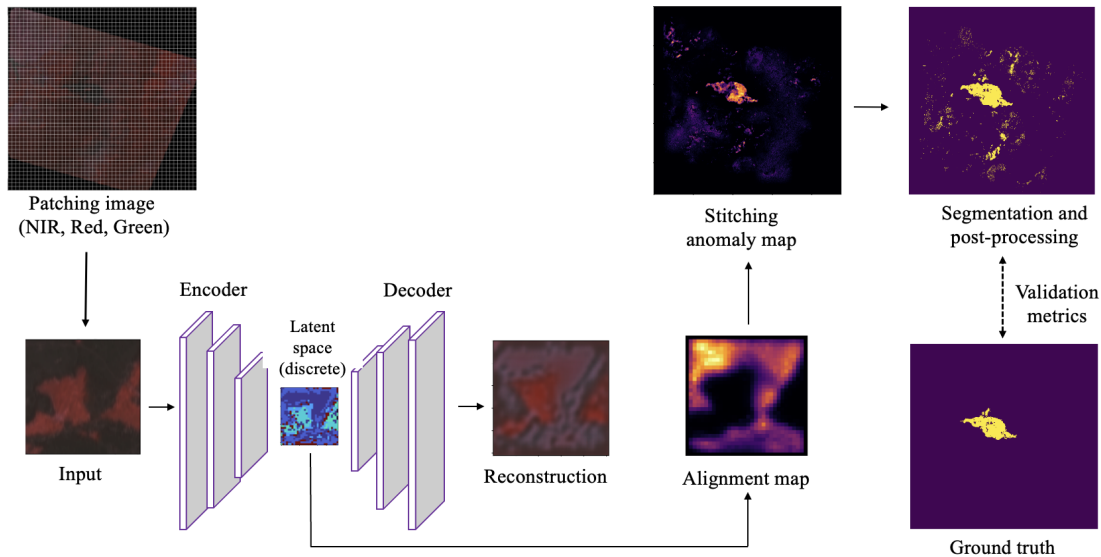

**Figure 2.3:** The proposed framework for burnt area extraction from satellite image based on anomaly detection. Here, the final anomaly score is derived based on the alignment map computed within the discrete latent space of the VQ-VAE.

## 2.4 Conclusion

This chapter has presented two contributions related to weakly supervised object discovery (or unsupervised anomaly detection) using VAE-based models. We have shown that the developed VAE-GRF and VQ-VAE can provide significant improvements when applied to both computer vision and remote sensing datasets. Our results confirm that the VAE-GRF might replace the classical VAE for many tasks, as we show how the stationary assumption does not introduce any additional computational cost. Thus, VAE-GRF offers an efficient and relevant prior for many practical applications, especially for the processing of images that exhibit textures. Since the stationary assumption for the GRF prior remains a strong assumption, future work can consider relaxing this assumption and it is possible to study ways to introduce non-stationary GRF prior while preserving the tractability of the model. Regarding the VQ-VAE models, their inner metrics outperform those of standard VAEs, without introducing more complexity in the model. To go further in the direction of leveraging inner metrics of VQ-VAEs, one can consider the benefits of using several codebooks to address multi-class anomaly detection. Others can work on more efficient training process of the discrete latent representation, in particular to deal with large codebooks.

# Chapter 3

# Multi-task partially supervised learning from multiple datasets

**Multi-task learning (MTL)** is an active research area of computer vision, in which a shared model is used to optimize multiple targets together from the same input image. As such, the model is compelled to learn a shared representation from the task-specific information, thus becomes better generalized and achieves improving performance (Vandenhende et al., 2021). Sharing model in the deep learning era also means reducing the number of parameters, or memory footprint, and thus reducing the computational cost. As the improvement of multi-task learning resulted from the inter-relationships obtained by co-training the tasks, the training examples are commonly assumed to be fully annotated for all the targets. This strong assumption would impede scalability, both in the number of tasks and the number of training examples, as it is expensive to maintain synchronization among the tasks, besides the multiplied annotating efforts.

**Multi-task partially supervised learning (MTPSL)** introduced by (W.-H. Li et al., 2022) relaxes such requirements as it allows each training example to be labeled only for one of the target tasks. This means that adding a new task to an existing dataset would add only new training examples annotated for that task without the need for annotating all existing images. It can also be seen as potentially augmenting the existing task with more training examples if the task inter-relationship can be learned from such settings to improve all task members' performance. To this end, MTPSL can be considered as a mix of both supervised learning and weakly supervised learning, formulating a label-efficient learning process with a parallel computation-efficient benefit (Fontana et al., 2024).

Inspired by these insights, we hypothesize that providing *weak but relevant training signals* for one task derived from another task's label would allow learning better joint representation of both tasks and improve their performance. This chapter[1] presents our **two contributions** related to MTPSL from multiple datasets to jointly learn *object detection* and *semantic segmentation* tasks, which has not been addressed in the literature. Section 3.2 first introduces *a feature-imitation knowledge distillation (KD)* approach which is employed for cross-task optimization, initially not available from partially annotated data. Section 3.3 then presents the *BoMBo (Box-for-Mask and Mask-for-Box) framework*, from which mutual weak losses are developed to take advantage of one task's labels to create pseudo-labels for the other task. The two proposed methods are experimented on both computer vision and remote sensing datasets, showing their high potential in both domains. Before going to the details of each contribution, Section 3.1 first provides some basic notions of MTPSL, allowing us to formulate the problem mainly focused on in this chapter.

---

[1]This chapter is built upon the research presented in the five papers (C24), (C30), (C33), (C34) and (C39) published during the postdoctoral work of Hoàng-Ân Lê in the scope of the ROMMEO project (2022-2024).

## 3.1 Problem formulation

Different from semi-supervised learning setting in which limited training data are assumed but still with all task annotations (Z. Chen et al., 2020; Huang et al., 2020), MTPSL from partially annotated data requires each training example to be labeled only for a single task. We follow the conventional MTL architecture including a shared encoder network with backbone and neck subnet for extracting features and decoder heads outputting task-specific predictions. The overview scheme is shown in Figure 3.1 with a MTPSL framework for object detection and semantic segmentation. As each training example is annotated for only a single task, not all the losses can be optimized together. Therefore, the network is fed with data annotated for one task after another and the gradients are computed separately for each head but accumulated for the shared subnet. All parameters are updated once the data for both tasks have been passed through the network. The object detection loss $\mathcal{L}_{\text{det}}$ includes the classification Focal Loss (Lin et al., 2017) and the localization Balanced L1 Loss (Pang et al., 2019) while the semantic segmentation head uses the regular cross-entropy with softmax loss $\mathcal{L}_{\text{seg}}$. To make up for the lack of corresponding annotations of one task when training the other, a *weak detection* loss ($\mathcal{L}_{\text{weak-det}}$) and a *weak segmentation* loss ($\mathcal{L}_{\text{weak-seg}}$) will be proposed to provide pseudo training signals. The two task losses are balanced by the parameter $\lambda$ to form a total training loss as follows.

$$\mathcal{L} = (\mathcal{L}_{\text{det}} + \mathcal{L}_{\text{weak-det}}) + \lambda (\mathcal{L}_{\text{seg}} + \mathcal{L}_{\text{weak-seg}}) \tag{3.1}$$

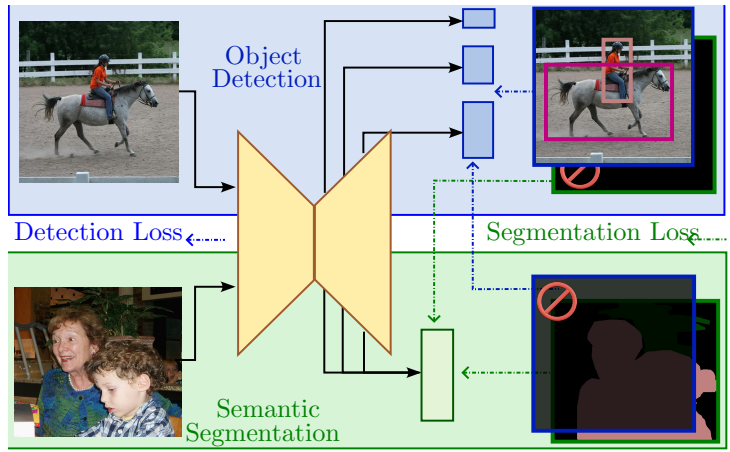

**Figure 3.1:** Multi-task partially supervised learning with two tasks, object detection (blue) and semantic segmentation (green). Each image is labeled for a single task, indicated by background colors, thus, can only train the respective head.

The main objective of our study is to design those two weak losses with an assumption that labels from one task could act as weak labels for the other, based on the cross-task correlation. In the following section, we first introduce weak losses by leveraging knowledge distillation from task-specific teacher networks.

## 3.2   Weak losses from knowledge distillation

### 3.2.1   Methodology

**Knowledge distillation**   (Hinton et al., 2015) showed that a network could benefit from a larger or an ensemble of models, called teachers, by mimicking the predicted logits or imitating the deep features (J. Guo et al., 2021) learned by them. Depending on the purpose, knowledge distillation (KD) can be first seen as a model compression technique which aims to reduce the model size with less performance sacrificing. Consequently, a network can learn from the outputs of a larger model and get improved without complicated modification in deployment. Since the pioneering publication (Hinton et al., 2015), KD has attracted various studies from which more understanding has been gained: (Beyer et al., 2022) with "a good teacher is patient and consistent"; (Mirzadeh et al., 2020) with an intermediate-sized teaching assistance network which better helps when the complexity gap is large between the teacher and student; (Xu et al., 2020) with their study on KD applied in a self-supervised or semi-supervised context by using contrastive loss, *etc.*

**Our proposition**   We propose to leverage KD technique by considering the main MTL network as the student, which will distil knowledge from two teacher networks, each corresponding to a specific task. An illustration of our framework is shown in Figure 3.2. We concatenate the features of all scale levels along the flattened spatial dimensions. The features of the student network are projected by a $1 \times 1$ convolution before being compared to the corresponding teachers'. The simple Mean Square Error (MSE) (H. Zhang et al., 2021) is applied for feature imitation distillation. Starting with Eq. (3.1), we formulate our MTLSL+KD loss function as follows.

$$\mathcal{L} = (\mathcal{L}_{\text{det}} + \mathcal{L}_{\text{KD-det}}) + \lambda \left( \mathcal{L}_{\text{seg}} + \mathcal{L}_{\text{KD-seg}} \right) \tag{3.2}$$

As each training image can optimize a single task, there are three cases for distilling the student features per iteration: (1) from the teacher whose task is annotated (1mse) so the student's features are forced to follow the teacher's while learning from the provided ground truths at the same time, (2) from the task teacher *without* annotations (0mse) so that the head is trained with one task (using ground truth) while the encoder is forced to follow the other's teacher, and (3) is the combination of both (2mse).

### 3.2.2   Experiments

**Experiments on vision datasets**   We first conduct experiments on the Pascal VOC (Everingham et al., 2015) containing 20 object categories with 8218 bounding-box annotated images for training, 8333 for validation, and 4952 for testing. Due to limited semantic segmentation annotations originally provided (1464/1449 for training/validation), the common practices use extra annotations provided by (Hariharan et al., 2011), resulting in 10582 training images. To simulate the partial supervision scenario, images are randomly sampled into 2 subsets, one for detection whose semantic annotations are held back and the other for semantic segmentation whose bounding-box annotations are kept out, resulting in 7558 and 7656, respectively. For validation, the originally provided validation set for semantic segmentation with both task annotations are used with 1,443 images. For setup, the ResNet50+PAFPN architecture is used as the teacher model and the ResNet18+FPN is used as the student. The task-specific heads are kept the same for the two networks. The teacher-student parameter ratio is 1.61. The results are shown in Table 3.1 with mAP metric for detection and mIOU for segmentation. By simply forcing the student encoders to imitate the output of the teachers, the corresponding results are improved, confirming the benefit of KD. Distilling the encoder neck features using one task's teacher while training the other task's head using provided ground truths (0mse) shows favorable results over distilling the same task that has annotations (1mse). The results are even higher when both tasks are optimized simultaneously, showing the benefit of multi-task data exploitation and joint-task optimization using KD.

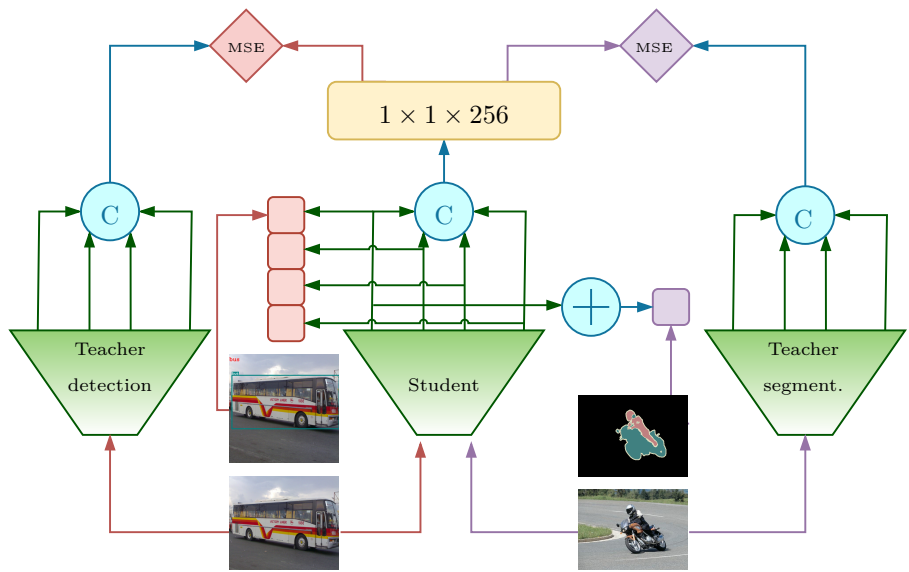

**Figure 3.2:** Overview of the proposed MTPSL+KD framework. Images and task-specific annotations are fed to the student and the respective teachers, with KD losses computed on teacher-student flattened and concatenated neck features.

**Experiments on remote sensing datasets** Experiments are also conducted on the IS-PRS 2D semantic labeling contest (Rottensteiner et al., 2012), consisting of aerial images collected from two different cities: Vaihingen and Potsdam. Vaihingen contains 33 tiles of 3K×2K pixels at 9-cm resolution whereas Potsdam contains 38 tiles of 6K×6K pixels at 5cm. The two subsets are divergent in scene appearance due to geographical differences: small and detached houses in Vaihingen and complex settlement structure with narrow streets in Potsdam. The tiles are cropped into $320 \times 320$-pixel image chips. The chips are, then, split into training and validation, resulting in 632/702 for Vaihingen and 2993/1515 for Potsdam. The ISPRS benchmark comes with pixel-wise annotations for 6 land-cover categories: *impervious surfaces, building, low vegetation, tree, car, and background*. For object detection task, we derive bounding-box targets from the provided pixel-wise semantic labels by taking a blob of connected pixels of the same class as an instance. To that end, the *building*, *tree*, and *car* class are selected for their interest in urban object detection. Table 3.2 shows the results on the ISPRS dataset with two settings. It can be seen that on remote sensing data, MTL does not gain much benefit over single-task learning (STL): the results are slightly higher for semantic segmentation and on par or slightly lower for object detection. This could be attributed to the domain shift between Potsdam and Vaihingen data due to the difference in geographic location and spatial resolution. By leveraging the use of KD, adding soft labels or feature distillation is generally helpful. This agrees to the discovery found in our experiments on the previous Pascal VOC dataset examined with the MSE setting. To this end, combining soft label with feature distillation, especially PDF-Distil (H. Zhang et al., 2021) results in the highest performance.

| Training | | Detection | Segmentation |
|---|---|---|---|
| Teacher | | 50.22 | 72.32 |
| Single task | | 42.907 | 65.291 |
| | + KD | 44.982 | 67.375 |
| Multi-task | | 45.678 | 67.310 |
| | + 1mse | 45.989 | 69.126 |
| | + 0mse | 47.337 | **70.056** |
| | + 2mse | **47.611** | 69.911 |

**Table 3.1:** Performance of MTPSL+KD on the Pascal VOC dataset. For multi-task learning, the distilled features can be on the task with (1mse), or without annotations (0mse), or both (2mse). The performances are in favor for (0mse) and (2mse).

| | Setting 1 | | Setting 2 | |
|---|---|---|---|---|
| **Model** | Detection (Potsdam) | Segmentation (Vaihingen) | Detection (Vaihingen) | Segmentation (Potsdam) |
| Teacher | 51.90 | 70.96 | 48.16 | 66.60 |
| Single-task | 45.37 | 66.15 | 45.07 | 59.24 |
| Multi-task | 45.33 | 68.60 | 44.13 | 60.56 |
| + Soft | 46.23 | 68.63 | 44.83 | 61.60 |
| + MSE | 45.47 | 68.29 | 44.44 | 60.93 |
| + PDF | 46.49 | 67.99 | 45.22 | **61.76** |
| + Soft + MSE | 46.34 | 68.60 | 44.61 | 61.70 |
| + Soft + PDF | **46.63** | **68.97** | **45.49** | 61.66 |

**Table 3.2:** Performance of MTPSL+KD on the ISPRS dataset with two scenarios. Without KD, MTLSP is on par with STL for object detection and slightly better for semantic segmentation. The distances are more pronounced when KD is employed.

## 3.3 BoMBo: Box-for-Mask and Mask-for-Box weak losses

In this section, we present our second and main contribution to the MTPSL learning framework, by introducing the BoMBo weak losses including two modules: Mask-for-Box and Box-for-Mask, allowing the joint optimization from the two tasks.

### 3.3.1 Mask-for-Box

The Mask-for-Box (M4B) module is inspired by the observation that the circumscribed rectangles of the connected components in a semantic mask could be considered referenced boxes for training object detection. These boxes correctly identify the object's category, by definition, but might not its instance, thus introducing noise during training. On the other hand, predicted boxes of a well trained network have learned to localize object instances, despite with uncertainty, and may fail to recognize the correct object category. The refining idea, therefore, aims to take the best of both worlds by re-localizing the referenced boxes with the guidance of the predicted boxes' localization information.

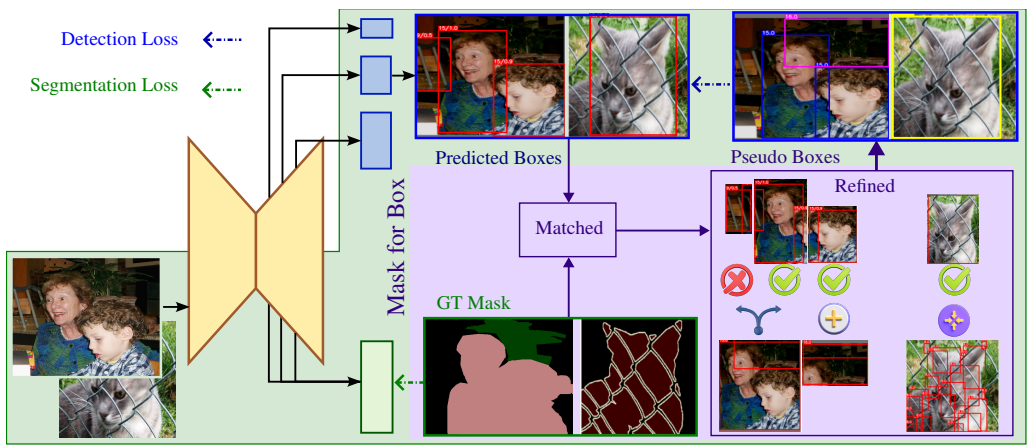

**Figure 3.3:** Overview of the Mask-for-Box module which uses predicted boxes to refine the circumscribed rectangles of the masks' connected components, by separating ⅄ multi-instance masks, merging 🌀 sub-instance masks, or using as ground truths ⊕. The good ✅ predicted boxes provide the instance cue while the wrong ⊗ are to be removed.

The process described in Figure 3.3 shows the M4B module applied to two input images. The circumscribed boxes of the connected components in the ground truth masks can be of one of the three cases. They may (1) contain multiple object instances, which should be separated (indicated by ⅄), (2) cover only a part of an object, which should be merged with other boxes (🌀), otherwise (3) they can be directly used for training (⊕). The referenced boxes are, first, matched with the predicted boxes following regular object detection process. The predicted boxes not overlapping with any reference box of the same class are removed (indicated by ⊗) while those with confidence higher than $\theta_1$ threshold are used for refinement (✅). Reference boxes that cannot be matched to any predicted box are merged together before being re-matched. For refinement, as it can be observed that a predicted box should have at least 2 sides touching those of a reference box, a refined box will have the two touching sides from the reference box and the other two from the predicted one. Although the same supervised losses can be used for training with refined boxes, only the localization loss will be used, with regard to our experiments described later.

### 3.3.2 Box-for-Mask

To train the semantic segmentation head from ground truth bounding boxes, the Box-for-Mask (B4M) module is proposed. The general scheme is showed in Figure 3.4. Let $\mathcal{C} = \{0, 1, \ldots, n_c\}$ be the set of categories where $0$ denotes the background, and $N = w \times h$ is the spatial dimension of the input image. Following (Song et al., 2023), two types of pseudo-masks are generated from provided ground truth boxes, a box-shaped mask $M^b \in \mathcal{C}^N$ by filling a box-enclosed area with its category, with priority given to the smaller box, and a coarse pseudo-mask $M^c \in \mathcal{C}^N$ using an unsupervised method such as GrabCut (Rother et al., 2004) or dense CRF (Krähenbühl and Koltun, 2011). As it can be observed that a ground truth box provides the upper limit of the object's extent, the coarse mask $M^c$ is filtered by $M^b$ so that $M_i^c = 0, \forall i : M_i^b = 0$, where $M_i^b = 0$ denoting a pixel $i$ in the box-shaped mask not enclosed in any ground truth box, thus surely a background pixel.

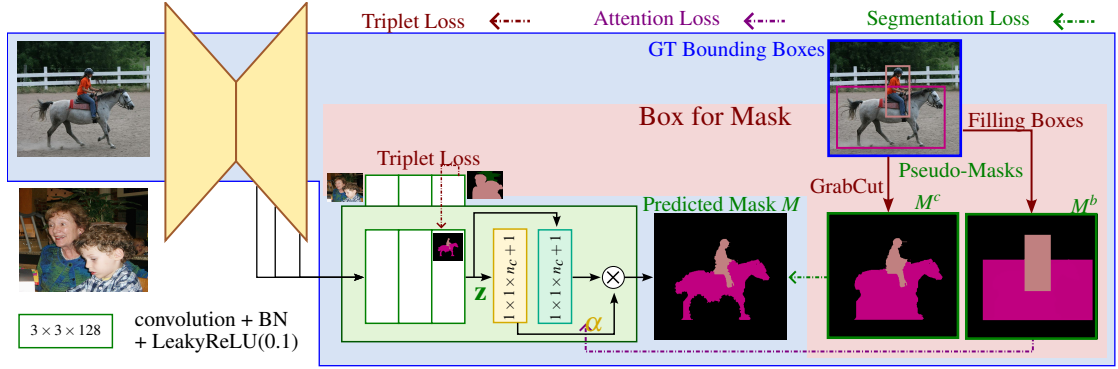

**Figure 3.4:** Overview of the Box-for-Mask module which generates pseudo-masks by filling the ground truth boxes with the same category and by using an unsupervised-learning method like GrabCut. The box-shaped pseudo-masks are used to train the attention map $\alpha$ while the others are predicted masks. The triplet loss constrains the embeddings to follow those with annotations.

**Semantic segmentation Loss**  The final predicted semantic mask $M$ is optimized using the coarse mask $M^c$ following

$$\mathcal{L}_S = -\frac{1}{N} \sum_i^N \sum_{j=0}^{n_c} \mathbb{1}_j \left(M_i^c\right) \log \left(\sigma \left(M_i\right)_j\right),  \tag{3.3}$$

where $\mathbb{1}_j(x)$ is the indicator function and $\sigma(\mathbf{x})_j = \frac{\exp(x_j)}{\sum_k^{n_c} \exp(x_k)}$, for $j \in \mathcal{C}$ is the softmax.

**Attention Loss**  Inspired by the previous work where the box-shaped masks are used to train attention masks $\alpha$, the segmentation head is modified to include an identical and parallel convolution layer (yellow block in Figure 3.4) to the last layer. Therefore, the attention map $\alpha = [\alpha^j]_{j \in \mathcal{C}} \in \mathbb{R}^{(n_c+1) \times N}$ shares the same dimension with the logit map $M^l$ and $\alpha^j \in \mathbb{R}^N, j \in \mathcal{C}$ is the attention map for class $j$. The predicted semantic mask $M$ is obtained by modulating the logits as $M = M^l \otimes \alpha$, where $\otimes$ is the Hadamard product. The attention map is optimized using the mean-squared error as shown in Eq. 3.4.

$$\mathcal{L}_\alpha = \frac{1}{N} \sum_i^N \left\| M_i^b - \alpha_i \right\|^2 . \tag{3.4}$$

**Triplet Loss**  To enforce consistency between predicted representations of unlabeled images, a triplet loss is proposed between the currently predicted ones, called queries, and those with annotations from previous batches, called keys.

Let $\mathcal{B}_k \subset \mathbb{N}^2$ be the *index set* of the pixels within a ground truth box $B_k$ and $\mathcal{M}_k \subset \mathcal{B}_k$ is the *index set* of the pixels having the same class prediction with $B_k$, or $\text{argmax}\,(M_i) = c_k, \forall i \in \mathcal{M}_k$. The unit-length mean embedding of the features vectors of the correctly predicted pixels in the bounding box $B_k$ is given by $\bar{\mathbf{z}}_{\mathcal{M}_k} = \frac{1}{|\mathcal{M}_k|} \sum_{j \in \mathcal{M}_k} \mathbf{z}_j$, where $\|\bar{\mathbf{z}}_{\mathcal{M}_k}\| = 1$ and $\mathbf{z}$ is the feature vector input to the attention and logit layers (Figure 3.4).

The triplet loss of a 3-tuple $\left(\bar{\mathbf{z}}_{\mathcal{M}_k}, \bar{\mathbf{z}}_{\mathcal{M}_k}^+, \bar{\mathbf{z}}_{\mathcal{M}_k}^-\right)$ is defined as

$$\mathcal{L}_{\text{object}} = \sum_{j \in [k]} \max \left(0, \gamma + d_{\mathbb{E}}\left(\bar{\mathbf{z}}_{\mathcal{M}_k}, \bar{\mathbf{z}}_{\mathcal{M}_k}^+\right) - d_{\mathbb{E}}\left(\bar{\mathbf{z}}_{\mathcal{M}_k}, \bar{\mathbf{z}}_{\mathcal{M}_k}^-\right)\right), \tag{3.5}$$

where $\bar{\mathbf{z}}_{\mathcal{M}_k}^+$ denotes the embeddings of the same class $c_k$ with $\mathbf{z}$ and $\bar{\mathbf{z}}_{\mathcal{M}_k}^-$ denotes the embeddings of different classes, $d_{\mathbb{E}}$ is the Euclidean distance and $\gamma$ is the margin, which is set to 0.1 The B4M loss is the sum of all three loses, i.e. $\mathcal{L}_{B4M} = \mathcal{L}_S + \mathcal{L}_\alpha + \mathcal{L}_{\text{object}}$.

### 3.3.3  Experiments

**Data and Setup**  Most of the experiments are conducted on the Pascal VOC with data preparation and annotation similar to the previous section. We also exploit the COCO dataset (Lin et al., 2014) for the final result report. To simulate data scarcity from COCO, one-eighth of the training set (14655 images) is uniformly sampled for training the detection task and another one-eighth for semantic segmentation task. The results are reported on the provided validation set of 5000 images. For backbone architectures, the ResNet (-18 and -50) and the SwinTransformer (Z. Liu et al., 2022) (-T, -B, -L) families are used. The FPN neck (Lin et al., 2017) is used with ResNet18 and PAFPN neck for the rest. The networks are trained for 70 epochs using SGD with learning rate of $1e-3$. We use an EMA-updated network to collect trained parameters from one being trained with SGD.

**Mask-for-Box** The first experiment is conducted to confirm the benefit of M4B loss for object detection. We compare two cases, when the circumscribed rectangles of the ground truth masks are used directly as referenced boxes for training (GT Masks) and when they are refined using the M4B. The results with classification loss (C) and both localization and classification (L+C) for pseudo-targets are also shown for ablation study. The results are shown in Table 3.3. It can be seen that directly using the ground truth masks' circumscribed rectangles would improve over the baseline due to more information being used and using M4B-refined boxes furthers the performance. Surprisingly, involving the classification loss when training with pseudo-targets (C and L+C) results in sub-optimal performance given that the pseudo-boxes' categories are correct by definition. The setting with only localization loss (L) consistently attains high performances.

| Training | $\mathcal{L}_{\text{M4B}}$ | ResNet18 | ResNet50 | SwinT | SwinB |
|---|---|---|---|---|---|
| Baseline | | 50.959 | 56.286 | 54.522 | 59.986 |
| GT Masks | C | 49.611 | 54.843 | 54.236 | 58.380 |
| | L | 50.168 | **56.588** | **55.296** | **59.473** |
| | L+C | **50.617** | 54.540 | 54.546 | 58.567 |
| M4B Refined | C | 50.396 | 55.187 | 55.625 | 60.036 |
| | L | **52.101** | **56.486** | **56.392** | **61.351** |
| | L+C | 48.031 | 51.787 | 56.283 | 60.508 |

**Table 3.3:** Performance of M4B losses, including a localization (L) and/or a classification loss (C), when combining mask-annotated and supervised images (baseline). M4B refinement with only localization loss outperforms using directly GT Masks' circumscribed rectangles.

**Box-for-Mask** This section confirms the design choices for B4M losses. Table 3.4 shows the performance when combining box-annotated with mask-annotated images (baseline) for training semantic segmentation. It can be seen that simply using cross-entropy loss with the pseudo-masks results in negative effects, even with more training data and half of them fully annotated. The pseudo-masks, even when confined to only areas within respective ground truth boxes, are noisy and confuse the training process when wrong pixels are used as targets. The attention maps, trained with the box-shaped masks $M^b$, soften cross-entropy loss' strong imposition on all pixels, thus help to take advantage of extra training data. Comparing to Shifting Rate (Song et al., 2023), which requires fixed statistics pre-computed for each dataset and pseudo-labels, and comparing to Affinity loss (Kulharia et al., 2020), the triplet object loss $\mathcal{L}_{\text{object}}$ provides improving performances.

**Combining Box-for-Mask and Mask-for-Box** In this section, we combine the two previous modules, Box-for-Mask and Mask-for-Box, or BoMBo, in one network for multi-task partially supervised learning. Table 3.5 shows the results with and without BoMBo on the VOC and COCO datasets. Although BoMBo outperforms all baselines on COCO, it only excels on both tasks for SwinL on VOC and has mixed results on the other architectures. The results show the benefit of the weak losses in using the extra data available to train one task with the other's annotations but also suggest an imbalance problem when combining training signals for both tasks.

|  | ResNet18 | ResNet50 | SwinT | SwinB |
|---|---|---|---|---|
| Baseline | 65.292 | 69.065 | 75.012 | 79.690 |
| $+\mathcal{L}_S$ | 64.740 | 67.582 | 73.830 | 78.510 |
| $+\mathcal{L}_\alpha$ | 67.623 | 71.763 | 76.269 | 80.654 |
| + B2S Affinity (Kulharia et al., 2020) | 66.954 | 71.562 | 76.266 | 80.880 |
| + Shifting Rate (Song et al., 2023) | 67.325 | 71.164 | 76.657 | **81.264** |
| $+\mathcal{L}_{\text{object}}$ | **68.545** | **72.306** | **76.981** | 81.236 |

**Table 3.4:** Performance of B4M losses when training box-annotated with supervised images (baseline). Simply applying cross-entropy loss ($\mathcal{L}_S$) on pseudo-masks (without attention maps) produces negative effects but adding the attention loss ($\mathcal{L}_\alpha$) provides a boost. The triplet object loss ($\mathcal{L}_{\text{object}}$) is generally more helpful than the other approaches.

| Training | Detection | | | | Segmentation | | | |
|---|---|---|---|---|---|---|---|---|
|  | ResNet50 | SwinT | SwinB | SwinL | ResNet50 | SwinT | SwinB | SwinL |
| VOC: MTL | **55.174** | 53.305 | **58.267** | 59.713 | **75.658** | **77.795** | **81.798** | 83.093 |
| + BoMBo | 54.885 | **54.696** | 58.259 | **60.687** | 74.861 | 77.433 | 81.205 | **83.310** |
| COCO: MTL | 17.198 | 15.158 | 14.914 | 19.766 | 54.535 | 56.280 | 63.802 | 67.788 |
| + BoMBo | **19.087** | **16.918** | **17.416** | **21.935** | **58.466** | **59.102** | **66.420** | **68.968** |

**Table 3.5:** Performance of BoMBo on the VOC and COCO datasets.

## 3.4 Conclusion

This chapter has presented two contributions in developing weak losses for multi-task partially supervised learning to jointly learn object detection and semantic segmentation from multiple datasets. The first approach is generally helpful by distilling knowledge from both soft labels and multilevel features of two big task-specific teachers. Yet, the performance improvement for both tasks remains limited compared to the two single-task teachers due to the challenging joint task optimization during training. In the second approach with the proposed Box-for-Mask and Mask-for-Box weak losses, our extensive experiments show that naively extracting shared information to train another task might result in negative impacts even when supervised data are also being used. The pseudo-boxes, despite having correct categories, can only help when being trained only for localization while pseudo-semantic masks should only constrain the attention-modulated predictions. One limitation of our study involves the assumption of the same data domain and shared class space between the two tasks. Future works can consider more complex scenarios from which partially annotated datasets come from different domains, for which domain adaptation techniques should be integrated to handle data domain shifts. Another perspective is to investigate the scalability of the proposed framework to more tasks. In this case, continual learning, which allows for the incremental integration of new tasks without retraining the entire model, would be useful.

# Chapter 4

# Self-supervised learning with multimodal image data

Deep learning methods based on supervised approaches often rely on large amounts of annotated images in order to learn efficient image features. Nonetheless, large datasets are very time-consuming and labor-intensive to annotate. As a practical approach in many vision-based research fields, exploiting supervised models pre-trained on ImageNet is a common way to boost the performance of deep neural networks when performing transfer learning or fine-tuning on smaller domain-specific image data. The pre-trained weights offer better representation capabilities than random parameters, in particular in the first layers of the network. Still, deeper layers should be fine-tuned on domain-specific data so that the network is able to extract features relevant to the new task and can perform predictions. Nevertheless, for certain tasks and datasets, suitable pre-trained models may not be available. In order to learn feature representations without relying on data annotations, **self-supervised learning (SSL)** has been introduced.

Earth observation using aerial and satellite remote sensing imagery produces terabytes of data in all forms everyday, thus making it nearly impossible to carefully annotate every image produced. If annotated, these data could be exploited to train supervised models for scene classification and serve as backbone models for other downstream tasks by leveraging neuronal activation from coarse to deeper layers as image-level or patch-level representations. One way to train generalized image representations **without relying on annotated data** is to perform SSL. In a nutshell, SSL basically learns deep feature representations that are invariant to sensible transformations, also called augmentations, of the input data. Such SSL models rely only on unlabeled data to create their own training objective (i.e. pretext task) without the need for time-consuming annotations. The representations learned by SSL methods should be discriminative for future downstream tasks, while being generalized enough to transfer across tasks without heavily retraining the model.

In this chapter[1], we investigate how these emergent developments in SSL can contribute to the field of remote sensing, with a particular focus on scene classification task. First, we remind some basic backgrounds in self-supervised representation learning in Section 4.1. In Section 4.2, we *conduct comparative experiments* to benchmark different SSL methods applied to scene classification in order to draw insights on their representation quality and transfer learning capacity. Section 4.3 then presents our main contribution in multimodal scene classification using optical and SAR images by developing *multimodal self-supervised pre-training* and *multimodal supervised contrastive learning* frameworks. Conclusions will be given in Section 4.4.

---

[1]This chapter is built upon the research presented in the three articles (J19), (J21) and (C25) during the PhD of Paul Berg (2021-2024).

# 4.1 Self-supervised representation learning

This section briefly reviews some significant state-of-the-art SSL methods, which are divided into four categories including *generative, predictive, contrastive* and *non-contrastive*. In the literature, contrastive and non-contrastive approaches can be regrouped into a single *joint-embedding* approach. Our choice is to distinguish these two, without any loss of generality. For more thorough background reviews of SSL approaches, we refer readers to (Jing and Tian, 2020; Ohri and Kumar, 2021).

**Generative Methods.** The first approach relies on generative models with the common pretext task of reconstructing an image using an Auto-encoder (AE). To minimize the reconstruction loss, the model has to learn to compress all significant information from the image into a latent space with a lower dimension, using the first network component called encoder. Then, a second component named decoder tries to reconstruct the image from the latent space. Variational Autoencoder (VAE) (Kingma and Welling, 2019) improves over the AE framework by encoding the parameters of the latent space distribution. They are trained to minimize both the reconstruction error and an additional term minimizing the Kullback-Leibler divergence between a known latent distribution (*i.e.,* the unit centered Gaussian distribution) and the one produced by the encoder. Recently, the use of vision transformers (Dosovitskiy et al., 2020) has enabled the development of large Masked Auto-encoders (MAE) (He et al., 2022) and Masked image modeling (MIM) (Xie et al., 2022) working at a patch level instead of pixel-wise to reconstruct entire patches with only a low number of visible patches. This reconstruction task produces robust image representations by appending a class token to the sequence of patches or simply by using a global average pooling on all the patch tokens. Since their appearance, MAE and MIM have remained among the most powerful SSL methods to date.

**Predictive Methods.** The second category involves models which are trained to predict the effect of an artificial transformation of the input image. Such an approach is motivated by the intuition that predicting the transformation requires learning relevant characteristics of semantic objects and regions within the image. By pre-training a model to predict the relative position of two image patches, (Doersch et al., 2015) managed to boost the performance of a model close to the one using ImageNet pre-trained weights. Another well-known predictive SSL method is the RotNet (Gidaris et al., 2018) which proposes to train a model to predict the rotation that was randomly applied to the input image. Solving this rotation prediction task requires the model to extract meaningful features to understand the image's semantic content. Similarly, another SSL model was developed to solve a jigsaw puzzle in (Noroozi and Favaro, 2016) to predict relative positions of image partitions that were previously shuffled. Also, by considering several types of augmentations, the Exemplar CNN (Dosovitskiy et al., 2014) is trained to predict the augmentations applied to images, including flipping, cropping, rotation, color jittering and contrast modification.

**Contrastive Methods.** A solution to learn effective representations is to design contrastive loss forcing the model to discriminate representations between views (augmentations) from the same image (*i.e.* positives) and those from different images (*i.e.* negatives). In other words, it aims to obtain similar feature representations for positive pairs while pushing away representations for negative pairs. Within this approach, the simplest objective is the triplet loss (Dong and Shen, 2018) from which a model is trained to provide a smaller distance between representations of an anchor and its positive than the distance between that anchor and a random negative. SimCLR (T. Chen et al., 2020) has then appeared to be one of the most popular contrastive SSL approaches. For each image in the training batch, two different views are created by sampling random augmentations. The model is trained to maximize the similarity between representations of an image and its positive views and to minimize the similarity between representations of that image and its negative views. In practice, contrastive methods often require a large batch size to provide enough negative samples to the loss and to prevent collapsing representations. To mitigate this, Momentum Contrast (MoCo) (He et al., 2020) was proposed to work with smaller batches with the same effective number of negative samples when computing the contrastive loss, by using a sample queued dictionary and a momentum encoder updating using Exponential Moving Average (EMA). Another method called SwAV (Caron et al., 2020) was designed to cluster representations to a set of prototypes and learn to match views to consistent clusters.

**Non-Contrastive Methods.** As part of joint-embedding approaches, other methods manage to train SSL models without using contrastive components in their loss. We categorize them as non-contrastive methods. As a popular one, BYOL (Grill et al., 2020) uses a teacher-student distillation network configuration, from which the teacher weights are defined as an EMA of the student weights. Another framework named DINO (Caron et al., 2021) uses a transformer-based self-distillation architecture where the teacher is defined as an EMA of the student network's weights. For a given positive pair, the student is trained to have the same centered and sharpened output of the teacher. Without requiring separate weights for each branch of the teacher-student pipeline as BYOL or DINO, another non-contrastive framework named Barlow Twins (Zbontar et al., 2021) leverages the information bottleneck theory to maximize the mutual information between two views by increasing the cross-correlation of their corresponding features while removing redundant information in their representations. To improve this framework, a method based on Variance, Invariance, Covariance regularization (VICReg) was proposed in (Bardes et al., 2022). Unlike Barlow Twins, the loss terms are independent for each branch except for the invariance which maximizes alignment between positives, enabling non-contrastive multimodal pre-training between text and image pairs.

## 4.2   SSL for remote sensing scene classification

This section presents our experiments to study the behavior and performance of some SSL models applied to remote sensing image classification task. Experiments are conducted on two popular public datasets: the high-resolution Resisc-45 (Cheng et al., 2017) with 45 classes and the EuroSAT (Helber et al., 2019) with 10 classes captured by the Sentinel-2 sensor. Since this study was done in 2021, we selected four state-of-the-art SSL methods at that time including two contrastive approaches: SimCLR (T. Chen et al., 2020) and MoCo-v2 (He et al., 2020), and two non-contrastive approaches: BYOL (Grill et al., 2020) and Barlow Twins (Zbontar et al., 2021). In this document, we only focus on two significant experiments to study the representation quality and transfer learning capacity of these SSL methods. For more extensive results and discussions, we refer readers to our article (J19).

**Quality of representation**   To evaluate the quality of feature representations, we use the linear evaluation protocol whereby models are first pre-trained in a self-supervised mode. Then, a linear classifier is trained on top of the frozen pre-trained models. Results can be seen in Table 4.1. As observed, SSL methods perform much better than the randomly initialized model: 82.55%-85.37% compared to 45.65% on Resisc-45 and 92.59%-95.59% compared to 63.48% on EuroSAT. Meanwhile, the supervised ImageNet initialization performs on par or better than self-supervised pre-training due to the fact that this model was pre-trained on a huge number of images ($14M$ labeled images) against using only around $16K$-$18K$ images of each studied dataset. On Resisc-45, it yields an accuracy about $5\%$ higher than the best performing SSL method (MoCo-v2). Nevertheless on EuroSAT, the two non-contrastive SSL models (BYOL and Barlow Twins), both perform better than the ImageNet initialization. One explanation could be the fact the small size of EuroSAT images is not well-suited for supervised ImageNet models (initially pre-trained on $224 \times 224$ pixel images) who tend to drop a lot of details through the use of pooling layers. In the meantime, the SSL models have been adapted to handle the small size of $64 \times 64$ pixels during our pre-training process. Among the four SSL methods, their behaviors are not the same on the two datasets. MoCo-v2 gives the best score on Resisc-45 but its performance is lower than BYOL and Barlows Twins on EuroSAT. Meanwhile, Barlow Twins performs very well on EuroSAT but stands behind MoCo-v2 and BYOL on Resisc-45.

| Pre-training method | Resisc-45 | | EuroSAT | |
|---|---|---|---|---|
| | Acc. | $100 \times \kappa$ | Acc. | $100 \times \kappa$ |
| Random initialization | 45.65±0.84 | 43.43±0.89 | 63.48±0.16 | 59.33±0.19 |
| ImageNet supervised | **90.32±0.00** | **89.93±0.00** | 94.46±0.00 | 93.84±0.00 |
| SimCLR | 82.55±0.68 | 81.84±0.71 | 92.59±0.05 | 91.76±0.05 |
| MoCo-v2 | 85.37±0.15 | 84.78±0.15 | 93.78±0.07 | 93.08±0.08 |
| BYOL | 85.13±0.07 | 84.52±0.31 | 94.92±0.12 | 94.34±0.13 |
| Barlow Twins | 83.14±0.30 | 82.44±0.31 | **95.59±0.17** | **95.08±0.19** |

**Table 4.1:** Classification performance on the Resisc-45 and the EuroSAT datasets with the linear evaluation protocol (3 runs).

While the linear protocol is a good measure of the representation quality of a frozen encoder, it is not the only strategy to evaluate the performance of SSL methods. Thus, we also investigate those SSL models using fine-tuning strategy for which the amount of labeled data are varied from $1\%$, $10\%$ and $100\%$ of the available labels. In general, we find that SSL models are more advantageous in the scenarios where the number of labels is limited. In addition, an ablation study on the role of different augmentation techniques including color jittering, flipping, cropping is conducted and analyzed in (J19).

**Transfer learning capacity**   Experiments are conducted to evaluate the transfer learning capability of SSL pre-training models, compared with those pre-trained in a supervised mode. To do this, we use the Resisc-45 dataset for SSL pre-training, then the EuroSAT for fine-tuning, and vice-versa. For these experiments, we also vary the amount of labeled data in order to highlight the relative performance gap between the different models, as experiments using less labeled data will put more importance to the model initialization for fine-tuning. The results can be observed in Figure 4.1. The orange, blue and green colors represent the performance using MoCo-v2, BYOL and supervised method for pre-training respectively. Figure 4.1a shows the results observed when transferring models pre-trained on Resisc-45 → EuroSAT, whereas Figure 4.1b shows transfer performance of EuroSAT → Resisc-45. We can observe that, when using a low amount of labeled data for fine-tuning, a performance difference between self-supervised and supervised models is significant. In fact, by pre-training without labels, SSL approaches could provide more generalized features which are more relevant in downstream tasks with a possible domain shift, whereas supervised models usually learn more domain-specific representations. This confirms that self-supervised models have a better transfer capability than their supervised counterparts. Yet, with higher numbers of labeled data, the gap between supervised and self-supervised model used as initialization shrinks as the fine-tuning reaches its optimal performance. We note that such a behavior is also confirmed by other research studies in the literature (Ericsson et al., 2021; Y. Wang et al., 2022).

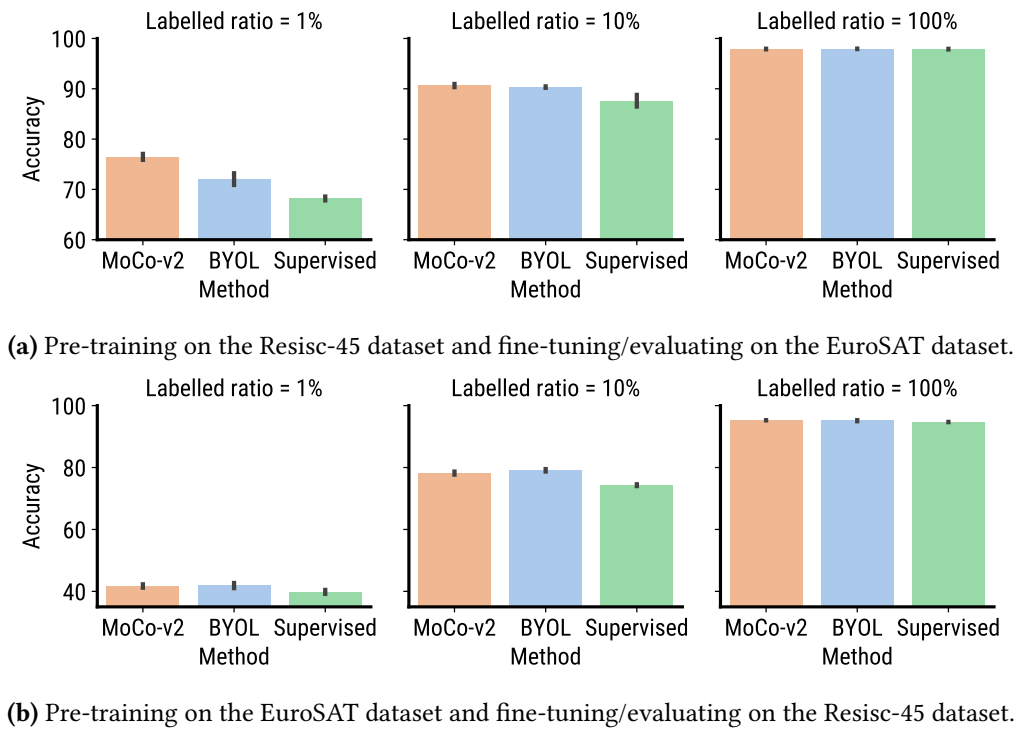

**(a)** Pre-training on the Resisc-45 dataset and fine-tuning/evaluating on the EuroSAT dataset.

**(b)** Pre-training on the EuroSAT dataset and fine-tuning/evaluating on the Resisc-45 dataset.

**Figure 4.1:** Performance in transfer learning of two SSL methods compared to supervised learning.

## 4.3 Multimodal SSL for scene classification

This section presents our main contributions on multimodal remote sensing image classification by developing joint multimodal SSL pre-training and multimodal supervised contrastive learning frameworks. Here, we focus on the use of multimodal datasets in which each sample involves multiple images captured by different sensors from the same scene.

### 4.3.1 Joint multimodal SSL pre-training

To adapt the contrastive learning framework to a multimodal dataset, we consider a multimodal sample as multiple views of the same scene. In this case, sensing with different sensors can be seen as a form of data augmentation which remains valid since the underlying ground truth label does not change. In addition, one can still use random augmentations such as color jittering but with caution since remote sensing images may not be as augmentable as natural images. Another consideration is that the multiple modalities composing a remote sensing dataset are heterogeneous (*e.g.*, optical and SAR). That is, they do not have the same resolution or channels. As such, they cannot be directly encoded using the same backbone model. Hence, we propose to use an encoder per modality and then learn a latent representation space which becomes domain and augmentation invariant.

In a setting with $M$ modalities, the joint learning pipeline proceeds as follows. After encoding each image using a modality-specific image encoder $f_m(\cdot), m \in [1, M]$ and projecting all feature representations into a hypersphere (*i.e.* feature space) $z_{i,m} = \frac{f_m(x_{i,m})}{\|f_m(x_{i,m})\|}$ (*i.e.* normalized to unit vectors), we consider $z_i = \text{cat}(z_{i,1}, \ldots, z_{i,M})$ the set of all fused representations in the batch. For each view representation $z_i$, we defines its set of multimodal positives $P_{\text{MM}}(i)$ as the set of representations corresponding to the same sample but from different modalities. The pre-training loss is formulated as follows:

$$\mathcal{L}_{\text{MM-SSL}} = \sum_{i=1}^{N} \frac{-1}{|P_{\text{MM}}(i)|} \sum_{j \in P_{\text{MM}}(i)}^{M} \log \frac{\exp(\langle z_i, z_j \rangle / \tau)}{\sum_{k \neq i} \exp(\langle z_i, z_k \rangle / \tau)}, \quad (4.1)$$

where $\tau > 0$ is a temperature parameter used to control the overall sharpening of the distribution produced by the softmax operator.

After pre-training the encoder models with $\mathcal{L}_{\text{MM-SSL}}$, the $f_m$ backbones can be used independently as discriminative initializations for downstream tasks. For each geographical location, a single feature representation is obtained by fusing the outputs from the different input modalities. We choose to generate these image-level representations by concatenating the modality-specific features. While this fusion strategy is effective, it could be further refined and improved in future work on multimodal representation learning.

**Experiments**   To evaluate our proposed framework, we conduct a series of multimodal SSL pre-training experiments using different combinations of modalities available in the Meter-ML (B. Zhu et al., 2022) dataset, created for methane source classification. Each methane-emitting facility in the dataset involves images captured by three sensors: Sentinel-1, Sentinel-2 and NAIP. The dataset contains facilities from six different classes. To experiment with both optical and SAR data, as well as varying spatial resolutions, we select sensor views from S1 (VH and VV) and S2 (RGB and NIR) at 10-m resolution, along with NAIP (RGB and NIR) at 1-m resolution. For backbone networks, we employ AlexNet for S1 and S2, and ResNet18 model for NAIP.

From Table 4.2, self-supervised pre-training consistently improves the performance compared to randomly initialized models. The multi-backbone architecture also scales with the number of modalities even when certain modalities are removed in the downstream task. The best performance ($73.68\%$) is obtained by combining all the three modalities for classification. Each modality therefore contains information relevant to the classification task that the methane source classifier is able to exploit. It is interesting to highlight that using fusion only during the downstream task and with a random initialization leads to a worst performance than using only S2 data during pre-training and fine-tuning. This means that this modality could provide the most important information for classification and that self-supervised pre-training allows a better discriminative initialization, which gives a better performance on the scene classification downstream task. Our experiments also show that the classification performance is generally better with artificial augmentations, since they offer in-modality positives.

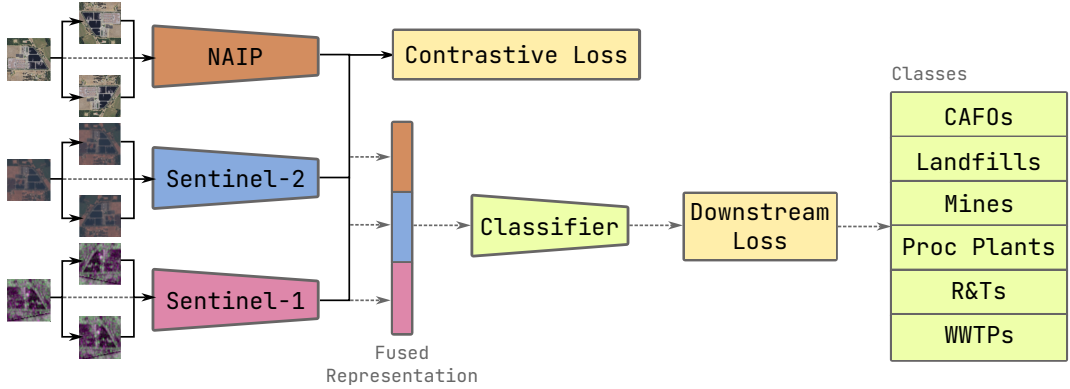

**Figure 4.2:** The proposed multimodal self-supervised pre-training and fine-tuning framework. Black arrows represent the forward data flow during pre-training while dashed arrows represent the flow during fine-tuning. Here, we illustrate a specific use-case of methane source classification (6 classes) from the Meter-ML dataset with 3 modalities: NAIP, Sentinel-2 and Sentinel-1 images.

| Pre-training | Downstream | | | | | | |
| --- | --- | --- | --- | --- | --- | --- | --- |
| | S1 | S2 | NAIP | S1 + S2 | S1 + NAIP | S2 + NAIP | S1 + S2 + NAIP |
| None | 47.37% | 64.29% | 62.03% | 65.04% | 63.16% | 68.42% | 65.79% |
| S1 | 51.13% | - | - | - | - | - | - |
| S2 | - | 70.30% | - | - | - | - | - |
| NAIP | - | - | 66.92% | - | - | - | - |
| S1 + S2 | 53.76% | 71.80% | - | 71.80% | - | - | - |
| S1 + NAIP | 56.39% | - | 70.68% | - | **72.18%** | - | - |
| S2 + NAIP | - | 71.43% | 68.42% | - | - | 72.18% | - |
| S1 + S2 + NAIP | **58.65%** | **72.56%** | **72.93%** | **72.18%** | 68.80% | **73.31%** | **73.68%** |
| S1 + S2 * | 50.00% | 69.55% | - | 65.04% | - | - | - |
| S1 + NAIP * | 55.26% | - | 70.30% | - | 60.90% | - | - |
| S2 + NAIP * | - | **72.56%** | 69.92% | - | - | 72.93% | - |
| S1 + S2 + NAIP * | 52.63% | 71.05% | 69.92% | 65.41% | 65.79% | 69.92% | 69.17% |

**Table 4.2:** Performance of the proposed multimodal SSL pre-training on the Meter-ML dataset using different pre-training and finetuning scenarios. When pre-training is None, it refers to randomly initialized baseline models. Pre-trainings annotated with * denote using no random augmentations and using only view (the original) by modality during pre-training.

### 4.3.2 Joint multimodal supervised contrastive learning

The final contribution in this chapter is a multimodal supervised contrastive learning framework for downstream classification. The proposed method is built upon the Supervised Contrastive (SupCon) learning (Khosla et al., 2020), which extends contrastive loss into supervised scenarios. This extension leverages label information to bring together samples from the same class while pushing apart those from different classes. Compared to self-supervised contrastive learning, the added supervision prevents the model from pushing samples from the same class apart, leading the model to extract more discriminative image representations for the downstream classification task.

Similar to the previous section, we extend the idea of SupCon to perform joint contrastive learning across multimodal image views during supervised fine-tuning. To do that, we define the the set of positives for a sample $i$ as $P_{\text{MM-S}}(i)$, consisting of all views (*i.e.,* different modalities and augmentations) of all the samples belonging to the same class. Using this positive set, we define the multimodal supervised contrastive loss by embedding the multimodal images and concatenating their representations into a single vector z, as follows:

$$\mathcal{L}_{\text{MM-S}} = \sum_{i=1}^{N} \frac{-1}{|P_{\text{MM-S}}(i)|} \sum_{j \in P_{\text{MM-S}}(i)} \log \frac{\exp(\langle z_i, z_j \rangle / \tau)}{\sum_{k \neq i} \exp(\langle z_i, z_k \rangle / \tau)}, \qquad (4.2)$$

where $\tau$ is again a temperature parameter.

Then, we propose to jointly perform the multimodal supervised-contrastive learning and train a linear classifier on top using the classical categorical cross-entropy (CE) in a single step. In addition to positives from other modalities, we also generate an augmented view for each image which acts as an in-modality positive. Therefore, in Eq. (4.2), each image representation will be pulled closer to those of its in-modality and cross-modality positives. Finally, only the non-augmented image representations are fed into the linear classifier to train the categorical CE objective. This prevents the linear classifier from encountering representations of out-of-distribution images, due to too aggressive random augmentations. Figure 4.3 illustrates an overview of our training pipeline for multimodal scene classification. Given multiple image encoders $f_m(\cdot)$, one for each modality, and a classifier $h_{\theta_h}(\cdot) : \mathbb{R}^d \to \mathbb{R}^C$, our training process uses a final loss combining the proposed multimodal SupCon loss and the categorical CE loss as follows:

$$\mathcal{L}_{\text{train}} = \alpha \mathcal{L}_{\text{MM-S}} + (1 - \alpha) \mathcal{L}_{\text{CE}}, \qquad (4.3)$$

where $0 \leq \alpha < 1$ is the weighting coefficient to balance the importance of the two loss terms. $\mathcal{L}_{\text{CE}}$ refers to the categorical cross-entropy loss:

$$\mathcal{L}_{\text{CE}} = \frac{-1}{N} \sum_{i=1}^{N} \log \left( \frac{\exp(h(z_i)_{y_i})}{\sum_{k=1}^{C} \exp(h(z_i)_k)} \right), \qquad (4.4)$$

where $y_i$ corresponds to the label of sample $i$ among the $C$ classes and $z_i$ is the fused representation of the multimodal sample $i$. This loss can be seen as a combination of a representation learning objective with a classification objective. In our experiments, we set $\alpha = 0.5$ to give equal importance to each term. Then, by setting $\alpha = 0$, one can ignore the multimodal SupCon term to only train using the standard CE loss.

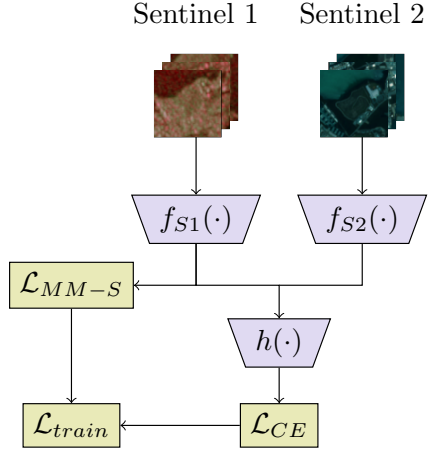

**Figure 4.3:** An overview of the proposed scene classification framework using multimodal SupCon loss combined with cross-entropy classification loss.

**Experiments**   We conduct experiments on multimodal scene classification using the public 2020 IEEE Data Fusion Contest (DFC2020) (Yokoya et al., 2020) dataset. Following the setting of (Scheibenreif et al., 2022), the pre-training phase is performed on the large-scale SEN12MS (Schmitt et al., 2019) dataset which contains more than $180K$ pairs of spatially-aligned Sentinel-1/Sentinel-2 observations without labels. Then, in the downstream task, $986$ and $5128$ paired Sentinel-1/2 from DFC2020 are used for fine-tuning and test, respectively. This dataset involves 8 land-use scene classes including *Forest, Shrubland, Grassland, Wetland, Cropland, Urban/Built-up, Barren and Water*. For a fair comparison to the baseline in (Scheibenreif et al., 2022), we adopt the ResNet-50 model as encoder and exploit the pre-trained weights provided by the authors using their multimodal SSL pre-training approach. With the fine-tuning, we also use the same hyper-parameters as the baseline.

The classification results on DFC2020 can be observed in Table 4.3. We also report the performance from the paper (Scheibenreif et al., 2022) for a comparison, although our reproduced results are marginally higher. The table shows that the proposed multimodal SupCon learning provides better performance than the standard use of categorical cross-entropy. The overall accuracy is significantly increased with a gain of $4.33\%$ ($72.62\%$ compared to $68.29\%$) and the average accuracy over all classes is also improved for about $1.5\%$ ($55.87\%$ compared to $54.39\%$). This confirms the effectiveness of the proposed frameworks, in particular the use of supervised contrastive loss to better leveraging class information and multiple data modalities during the fine-tuning phase.

In our article (J21), we also conduct experiments on the Meter-ML dataset to confirm the superior performance of the proposed multimodal SupCon method. In addition, a sensitivity study on the weighting coefficient $\alpha$ in Eq. (4.3) is performed to investigate its behavior. Finally, to demonstrate the consistency of the proposed method to other supervised classification loss functions, we carry out a comparative study by combining our proposed multimodal SupCon objective with different losses such as the weighted CE (WCE) loss and the Focal loss. Results on the Meter-ML dataset in Table 4.4 show that adding the proposed SupCon loss consistently improves the classification performance.

| Class | CE | CE | SupCon+CE |
|---|---|---|---|
| | (Scheibenreif et al., 2022) | (Our impl.) | (Proposed) |
| Forest | 65±8 | 63.49±5.90 | **75.54±5.01** |
| Shrubl. | 56±11 | **58.31±4.59** | 52.26±4.42 |
| Grassl. | 9±6 | **12.60±7.53** | 6.18±4.93 |
| Wetland | 15±8 | 14.43±11.55 | **20.90±7.11** |
| Croplan | 45±8 | 52.65±7.94 | **59.65±5.07** |
| Urban | **95±1** | 91.14±3.48 | 89.94±2.63 |
| Barren | 39±3 | 43.91±2.7 | **44.09±3.49** |
| Water | **99±1** | 98.63±0.41 | 98.42±1.25 |
| Overall | 67±2 | 68.29±1.66 | **72.62±1.31** |
| Average | 53±2 | 54.39±0.72 | **55.87±0.89** |

**Table 4.3:** Performance of the proposed multimodal SupCon loss on the DFC2020 dataset.

| Loss function | Accuracy |
|---|---|
| CE loss | 70.74 ± 1.74 |
| SupCon+CE (ours) | **73.83±2.00** |
| WCE loss | 71.23 ± 0.69 |
| SupCon+WCE (ours) | **74.74±0.55** |
| Focal loss | 71.50±0.93 |
| SupCon+Focal (ours) | **75.19±0.76** |

**Table 4.4:** Performance on the Meter-ML dataset yielded by combining the proposed multimodal SupCon loss with different classification loss functions.

## 4.4   Conclusion

This chapter has presented two contributions to the task of multimodal scene classification in remote sensing. First, we propose a self-supervised framework to learn discriminative representations for multimodal remote sensing datasets. Our experiments show that SSL pre-training consistently improves the downstream classification performance and scales with the number of image modalities. Secondly, we generalize the contrastive loss to supervised scenarios by leveraging multimodal supervised contrastive learning during classification fine-tuning. We have shown that combining such a multimodal SupCon loss with any classification loss would significantly improve the performance. Despite the existence of multimodal image views, we recommend to leverage artificial augmentations in both self-supervised and supervised contrastive frameworks.

The setting we study in this chapter concerns multi-modal datasets whose samples are co-registered images captured by different sensors. As perspective work, one can consider a more challenging setting related to modality missing, where some samples lack of certain modalities during the training. In this case, samples that contain all modalities could serve as anchors, providing a reference for aligning and learning from incomplete samples. Developing methods that can effectively leverage these anchors, while still learning robust representations from partially observed data, would be a valuable extension of the current framework.

# Chapter 5

# Few-shot learning with hierarchical image data

Inspired by the human ability to learn new abstract concepts from very few (or even one) examples and to quickly generalize to new instances, **few-shot learning (FSL)** appears as one of the alternative ways to deal with the "data-hungry" issue. FSL methods can be divided into three categories (Sun et al., 2021): metric learning, meta-learning and transfer learning. Metric learning methods learn a distance function that brings samples from the same category as close as possible in the feature space while pushing samples from other categories as far away as possible. Meta-learning, also known as "learning to learn", is the most common approach in FSL, which focuses on training models that can rapidly adapt to new tasks by leveraging experience from a diverse set of previous tasks. Transfer learning aims at using the knowledge gained from relevant tasks towards new tasks, *e.g.* fine-tuning the pre-trained models is a powerful transfer method.

In recent years, several approaches were proposed to tackle the problem of **few-shot remote sensing scene classification (FSRSSC)**, mostly based on the idea of prototypical network (Snell et al., 2017). The authors in (Cheng et al., 2021) used a Siamese ProtoNet with prototype self-calibration and inter-calibration to learn more discriminative prototypes. In (P. Zhang et al., 2020), the authors introduced a pre-training step on the base data to provide better initialization of the feature extractor and performed FSRSSC using cosine distance metric. However, to the best our knowledge, the majority of these methods have focused only on visual scene information to improve feature representations without considering semantic knowledge that may exist within these classes. Yet, this type of semantic knowledge about classes, which can consist of attributes, word embeddings or even a knowledge graph like WordNet (Miller, 1998), is commonly used in zero-shot learning and increasingly in few-shot natural image classification approaches.

In this chapter[1], we build on prototypical networks to tackle FSL in hierarchical scene classification with **two contributions**. Our first contribution, presented in Section 5.2, aims to define a *hierarchical variant of the ProtoNet*. In a nutshell, hierarchical prototypes are attached to each level of the class hierarchy, allowing us to first consider high-level aggregated information before making a fine prediction. Section 5.3 presents another framework to implicitly discover the hierarchical structure of remote sensing scene images within a *hyperbolic space*. Such a non-Euclidean space is well-known for its ability to embed hierarchical and tree-like data structures with low distortion, making it suitable for capturing semantic relationships in complex image datasets. Before describing each contribution, let us provide some backgrounds on FSRSSC with prototypical networks in Section 5.1.

---

[1]This chapter is mainly built upon the research presented in the three articles (C22), (C27) and (J20) during the PhD of Manal Hamzaoui (2019-2023).

## 5.1 Problem formulation

In few-shot classification, we assume to have two sets, a large labeled training set, referred to as the base set $D_{base}$, and a test set with few labeled images per class, the novel set $D_{novel}$. The classes that constitute the base and novel sets, denoted $C_{base}$ and $C_{novel}$ respectively, are disjoint $C_{base} \cap C_{novel} = \emptyset$. To mimic the sparsity of the test data in the training stage, we adopt the K-way N-shot strategy (an episodic learning strategy) used in various FSL studies (Snell et al., 2017; Vinyals et al., 2016), in which $K$ refers to the number of classes and $N$ (usually set to 1 or 5) is the number of labeled images per classes during a training/testing episode. For each training episode, we randomly sample a subset of $K$ classes out of $C_{base}$ which we denote $C_e$. We then randomly sample $N$ labeled images from $D_{base}$ for each class $k \in C_e$, resulting in the episode support set $S = \{(x_i, y_i)\}_{i=1}^{K \times N}$, where $x_i$ is an image and $y_i \in C_e$ its corresponding label. Furthermore, for the same $K$ selected classes, we sample $N'$ labeled images for each class $k \in C_e$ to form a set known as the query set $Q = \{(x_i, y_i)\}_{i=1}^{K \times N'}$. A training episode therefore has a total of $K \times (N + N')$ samples. In this training step, the support set $S$ and the query set $Q$ are used to learn the model that projects the input images into the feature space. The testing step is also carried out with the same episodic strategy where we have an unlabeled query set $Q$ (drawn from $D_{novel}$) for which we want to predict the class label of each query sample $x_i \in Q$ using the labeled support set $S$ (also drawn from $D_{novel}$). Fig. 5.1 shows a visualization of the K-way N-shot scene image classification episodes.

Prototypical networks (Snell et al., 2017) are metric learning-based methods which learn a distance function in order to bring samples within the same category as close as possible in the feature space, while pushing away samples from other categories. They adopt an episodic strategy to train the meta-learner classifier. Given an episode with a support set $S$ and a query set $Q$, we compute the representations of the images in both sets $S$ and $Q$ using the meta-learner feature extractor $f_\Phi$ (a neural network such as CNN) parameterised by $\Phi$. Thereafter, the support set representations are averaged to compute the prototypes $p^k$ for each class $k \in C_e$ as follows:

$$p^k = \frac{1}{N} \sum_{(x_i, y_i) \in S^k} f_\Phi(x_i), \tag{5.1}$$

where $S^k$ is the subset of the episode support set $S$ that contains the samples of class $k \in C_e$, $C_e$ is the set of classes sampled during episode $e$.

To optimise the feature extractor $f_\Phi$, we minimise the loss function:

$$\mathcal{L} = -\frac{1}{K \times N'} \sum_{k \in C_e} \sum_{(x_i, y_i) \in Q^k} \log p_\Phi(y_i = k \mid x_i), \tag{5.2}$$

where $Q^k$ is the subset of the episode query set $Q$ composed of samples from class $k$ and $p_\Phi(y_i = k \mid x_i)$ is the probability of predicting a query sample $(x_i, y_i) \in Q$ as class $k$ and is given as:

$$p_\Phi(y_i = k | x_i) = \frac{\exp(-d(f_\Phi(x_i), p^k)/\tau)}{\sum\limits_{k' \in C_e} \exp(-d(f_\Phi(x_i), p^{k'})/\tau)}, \tag{5.3}$$

where $d(.)$ is the Euclidean distance and $\tau$ is the temperature hyper-parameter.

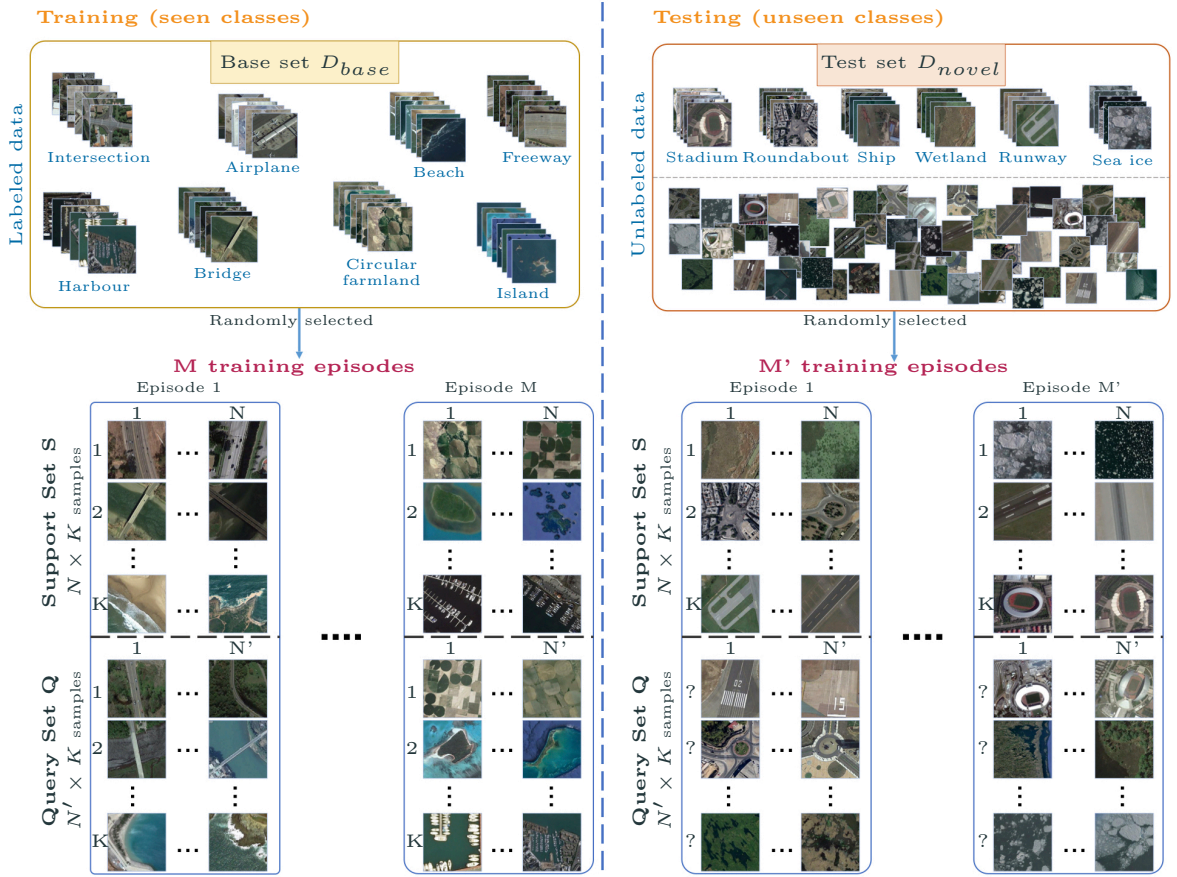

**Figure 5.1:** Illustration of K-way N-shot scene image classification. The left side shows the $M$ episodes of the training step; each episode consists of $K \times N$ support samples and $K \times N'$ query samples. The testing step is similarly defined on $M'$ episodes, as shown on the right.

## 5.2 Hierarchical prototypical networks

### 5.2.1 Overall framework

We propose a meta-learning framework whose complete pipeline is illustrated in Fig. 5.2 to solve the few-shot classification problem when a hierarchy that describes the organization between the classes is available. We train a meta-learner classifier by adopting an episodic training strategy. During training stage, using the support set $S$, we compute $N$ prototypes $\mathcal{P} = \{p^k\}_{k \in C_e}$ for each class in the current task (episode) and $K_h$ hierarchical prototypes for their super-classes. The query features are then compared to both the scene and the hierarchical prototypes, allowing us to compute an episodic error at different levels of the class hierarchy $\mathcal{T}$ to be minimised and used to fine-tune the parameters $\Phi$ of the feature extractor $f_\Phi$. At testing stage, the parameters $\Phi$ of $f_\Phi$ are fixed and the meta-learner classifier is evaluated on a set of episodes sampled from the novel classes in $D_{novel}$.

Here, we rely on the prototypical networks and introduce the hierarchy knowledge thanks to the definition of hierarchical prototypes. The overall idea is to regularize the latent space by putting closer classes that are in the same branch of the class hierarchy $\mathcal{T}$,

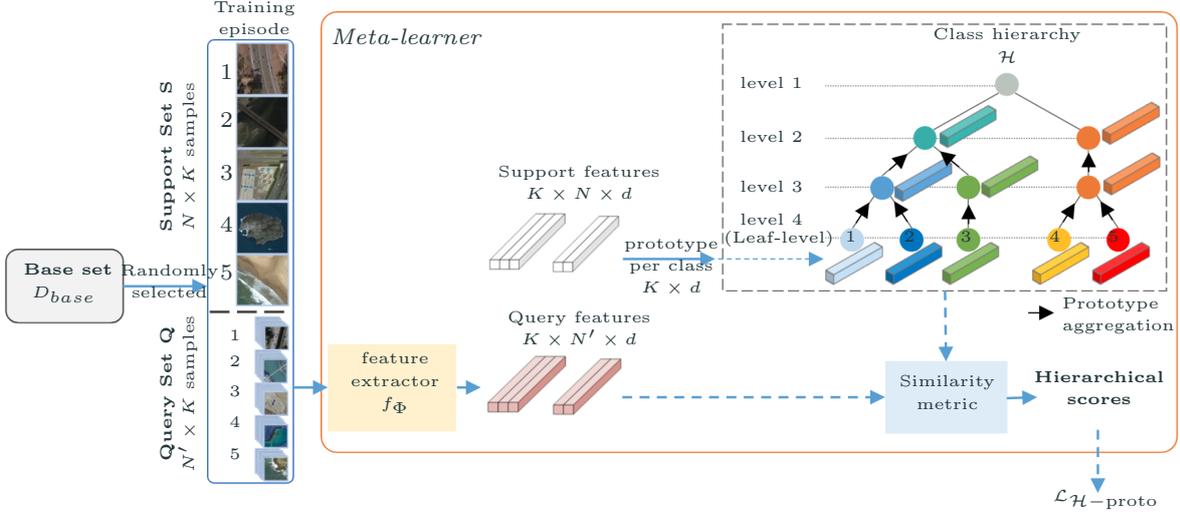

**Figure 5.2:** Overall framework of the proposed hierarchical prototypical network for few-shot image classification. In this example (one-shot), $K = 5$, $K_h = 5$, $N = 1$, $N' > 1$ (usually set to 15).

and pushing apart classes that have common ancestors in higher levels of $\mathcal{T}$.

To properly formulate our approach, given an episode, we first compute the prototypes per class which are prototypes at the leaf-level of $\mathcal{T}$ (following Eq. (5.1)). We then compute the hierarchical prototypes by aggregating the leaf-level prototypes according to $\mathcal{T}$.

The prototypes of the super-classes $k \in C_e^l$ (the hierarchical prototypes) at level ($1 < l < L$ with $l = 1$ the root node and $L = \text{height}(\mathcal{T})$) are denoted as $\mathcal{P}_l = \left\{ p_l^k \right\}_{k \in C_e^l}$ and computed as the mean of support samples of the super-class sub-tree $S_l^k$ similarly to Eq. (5.1):

$$p_l^k = \frac{1}{|S_l^k|} \sum_{(x_i, y_i) \in S_l^k} f_\Phi(x_i).$$ (5.4)

Note that when $l = L$, the prototypes at level $l$ are the prototypes at the lowest level of $\mathcal{T}$ (leaf-level prototypes). The hierarchical ProtoNet outputs a distribution over classes for each query sample $x_q \in Q$ at different levels of $\mathcal{T}$, based on a softmax over the distances to the prototypes of each level $l$ in $\mathcal{T}$. We then formulate the probability of predicting the query features $f_\Phi(x_q)$ and the prototype $p_l^k$ of its super-class $k$ at level $l$ in $\mathcal{T}$ as formulated in Eq. (5.3) as:

$$p_\Phi(y_i^l = k | x_i) = \frac{\exp(-d(f_\Phi(x_i), p_l^k)/\tau)}{\sum_{k' \in C_e^l} \exp(-d(f_\Phi(x_i), p_l^{k'})/\tau)},$$ (5.5)

where $y_i^l$ is the ancestor of $y_i$ at level $l$, $C_e^l$ represents the super-classes at level $l$ at the current episode.

We therefore optimise a new loss function given as

$$\mathcal{L}_{\mathcal{H}-\text{proto}} = \sum_{l=2}^{L} \lambda_l \mathcal{L}_l,$$ (5.6)

where $\lambda_l = \frac{\gamma^{l-1}}{\sum_{l'=2}^{L} \gamma^{l'-1}}$, $\gamma$ is a hyper-parameter that controls the importance of each level in the hierarchy and $\sum_{l=2}^{L} \lambda_l = 1$. $\mathcal{L}_l$ represents the prototypical network loss at level $l$ of the class hierarchy $\mathcal{T}$.

As such, we can tune the importance of each level of the hierarchy into the learning process: by choosing low values of $\gamma$, we put more importance into organising the higher levels of the hierarchy; a value close to one gives the same importance for all the levels; a high value tends to behave like the *flat* cross entropy loss formulation.

### 5.2.2 Experiments

We conduct experiments using the Resisc-45 dataset which is split into three disjoint subsets: meta-training $D_{base}$, meta-validation $D_{val}$, and meta-test $D_{novel}$ containing 25, 12, and 8 categories, respectively. We note that the meta-validation set is used for hyper-parameter selection in the meta-training step. The meta-training set is further divided into three subsets: training, validation, and test sets. Following recent FSRSSC studies (X. Li et al., 2021; P. Zhang et al., 2020; P. Zhang et al., 2021), we utilise ResNet-12 as a backbone for feature extraction and resize all the images to $80 \times 80$ pixels to fit our designed feature extractor. We also adopt the pre-training strategy as suggested in (P. Zhang et al., 2020) to better initialise the meta-learner feature extractor.

We train our meta-learning for 400 epochs, with the best model parameters chosen based on the best overall accuracy on the validation set. In standard deep learning, an epoch implies that the entire train set passes through the deep neural network once. However, in meta-learning, an epoch is a set of episodes randomly sampled from the base set $D_{base}$, which we set to 500 episodes per epoch. We optimise the model based on the average loss of 2 episodes, i.e. the batch size is set to 2 episodes. We use SGD optimiser to update the network parameters with a momentum set to 0.9 and a weight decay set to 0.0005. The learning rate is fixed to $10^{-3}$. After each training epoch, we test our model on a validation set $D_{val}$ by randomly sampling 500 episodes, the network weights with the highest validation overall accuracy are retained as the best results. For the hierarchical hyper-parameter $\gamma$, we assigned different values ($\gamma = 1$, $\gamma < 1$ and $\gamma > 1$) in order to observe its impact on the framework performances. For the meta-testing stage, we conduct a 5-way 1-shot and 5-way 5-shot classification following the widely used meta-learning protocol. We evaluate the best model on 1000 randomly sampled episodes from the test set $D_{novel}$. Following the FSL evaluation protocol (Snell et al., 2017), for K-way N-shot episode, we randomly sample 15 images per class to form the query set $Q$, making a total of $K \times 15$ query images per episode. More details about the experimental setup can be found in (C22).

Table 5.1 reports the classification performance of the different approaches in both 5-way N-shot configurations, N = 5 and N = 1 respectively. We re-implement the *flat* method (ProtoNet) according to (P. Zhang et al., 2020). Our proposed *h-ProtoNet* achieves the highest accuracy and outperforms both *flat* prototypes (ProtoNet) and the *soft-labels* hierarchical loss in the 5-shot setting. We obtain the best performance with $\gamma = 2$, that is to say when we put more weights on the prototypes that correspond to the lower level of the hierarchy (corresponding to the leaf nodes, $\gamma > 1$). In the 1-shot setting, we outperform

| Method | hyper-param | overall acc | L3-acc | L2-acc | $P_H$ |
|---|---|---|---|---|---|
| ProtoNet | 1 | $77.84 \pm 0.40$ | $80.89 \pm 0.37$ | $85.55 \pm 0.41$ | $81.43 \pm 0.35$ |
| Soft-labels | 4 | $76.77 \pm 0.41$ | $79.93 \pm 0.37$ | $85.39 \pm 0.42$ | $80.70 \pm 0.35$ |
| h-ProtoNet (ours) | 0.5 | $77.75 \pm 0.39$ | $80.98 \pm 0.35$ | $85.53 \pm 0.41$ | $81.42 \pm 0.34$ |
| h-ProtoNet (ours) | 1 | $\underline{78.41 \pm 0.40}$ | $\underline{81.60 \pm 0.36}$ | $\mathbf{85.99 \pm 0.40}$ | $\underline{82.00 \pm 0.34}$ |
| h-ProtoNet (ours) | 2 | $\mathbf{78.65 \pm 0.40}$ | $\mathbf{81.72 \pm 0.36}$ | $\mathbf{85.99 \pm 0.40}$ | $\mathbf{82.12 \pm 0.34}$ |

(a) 5-shot results

| Method | hyper-param | overall acc | L3-acc | L2-acc | $P_H$ |
|---|---|---|---|---|---|
| ProtoNet | 1 | $58.90 \pm 0.61$ | $62.56 \pm 0.63$ | $72.24 \pm 0.71$ | $64.57 \pm 0.57$ |
| Soft-labels | 4 | $\mathbf{61.69 \pm 0.62}$ | $\mathbf{65.66 \pm 0.62}$ | $\mathbf{74.88 \pm 0.70}$ | $\mathbf{67.41 \pm 0.57}$ |
| h-ProtoNet (ours) | 0.5 | $\underline{60.72 \pm 0.62}$ | $\underline{64.77 \pm 0.63}$ | $\underline{74.21 \pm 0.70}$ | $\underline{66.57 \pm 0.57}$ |
| h-ProtoNet (ours) | 1 | $60.20 \pm 0.62$ | $64.00 \pm 0.62$ | $72.83 \pm 0.73$ | $65.68 \pm 0.59$ |
| h-ProtoNet (ours) | 2 | $60.08 \pm 0.62$ | $64.09 \pm 0.62$ | $73.58 \pm 0.71$ | $65.92 \pm 0.58$ |

(b) 1-shot results

**Table 5.1:** Few-shot classification results on the Resisc-45 dataset. All accuracy results are averaged over 1000 test episodes and are reported with a 95% confidence interval. Reference methods include ProtoNet (Snell et al., 2017) and Soft-labels (Bertinetto et al., 2020).

the *flat* ProtoNet achieving better results with $\gamma = 0.5$ which corresponds to the higher level of the hierarchy (corresponding to nodes close to the root, $\gamma < 1$). We observe that incorporating information related to the class hierarchy leads to improved performance in our approach, as well as with *soft labels* method (Bertinetto et al., 2020). The performance of our hierarchical prototypes is comparable to that of *soft labels*, indicating that we are effectively taking hierarchical information into account. Further investigation is needed to understand why our performance is either superior or inferior to that of *soft labels*, which we plan to explore in future research.

We observe that our h-ProtoNet is more sensitive to the class hierarchy when few labeled data are available (the 1-shot setting), thus further enhancing the performance of the *flat* ProtoNet. In this case, prototypes at higher levels of the hierarchy allow for significant information transfer between leaf prototypes. Note that these values of $\gamma = 2$ for 5-shot and $\gamma = 0.5$ for 1-shot would have been selected if we perform a cross-validation on the validation set. We argue that the improvement observed in the case of the hierarchical prototypes is due to an efficient regularization of the latent space, with a loss that encourages leaves within the same branch of the level hierarchy to be closer. As such, the performances at level 2 and 3 are improved, but also the overall accuracy.

## 5.3 Hyperbolic prototypical networks

### 5.3.1 Overall framework

Our hyperbolic meta-learning framework for FSRSSC is illustrated in Figure 5.3. As it is not yet obvious how to implement a fully hyperbolic network (Peng et al., 2021), we follow the trend of hyperbolic works and adopt a hybrid "Euclidean-Hyperbolic" architecture. During the training stage, we first project the scene images of both support and query sets sampled from $D_{base}$ into the Euclidean embedding space using the feature extractor $f_\Phi$. As we use a hybrid "Euclidean-Hyperbolic" architecture, we suggest using the Euclidean feature clipping technique (Y. Guo et al., 2022) to ensure better numerical stability of the model. The clipped features of the support and query representations are then mapped into the hyperbolic space. Class prototypes are further calculated via the *Einstein midpoint*, which is an alternative way to calculate the mean in hyperbolic space. Subsequently, we follow the same steps as for the baseline prototypical network and compute the hyperbolic distances between the query hyperbolic representation and the hyperbolic prototypes, which will be used as a proxy for the class membership probabilities. The model parameters are finally optimised by minimising Eq. (5.2). At the test stage, the model parameters $\Phi$ are fixed and evaluated in the hyperbolic space on a set of episodes sampled from $D_{novel}$. We now detail further the different steps of our method.

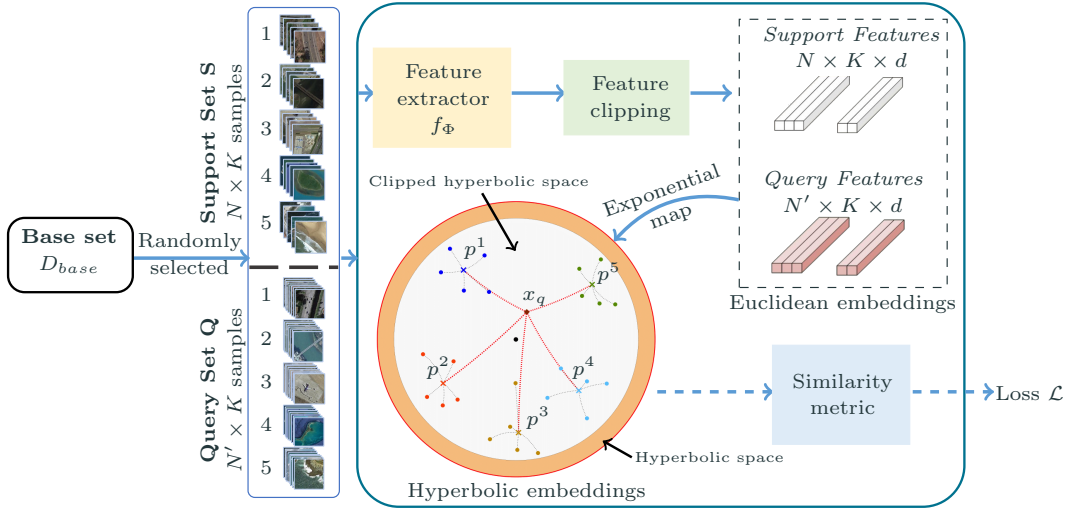

**Figure 5.3:** Overall framework of the hyperbolic prototypical network for few-shot image classification. In this example (5-shot), $K = 5$, $N = 5$, $N' \geq 1$ (usually set to 15) and the latent space dimension $d = 2$.

**Hyperbolic Geometry**  Similarly to (Khrulkov et al., 2020), we adopt here the Poincaré Ball model (Nickel and Kiela, 2017), a popular hyperbolic model. The Poincaré Ball model $(\mathbb{B}^d, g^\mathbb{B})$ is a Riemannian manifold defined by the open $d$-dimensional ball of radius $\frac{1}{\sqrt{c}}$, $\mathbb{B}_c^d = \{x \in \mathbb{R}^d : c \|x\|^2 < 1\}$, where $-c$ is the curvature of the hyperbolic space and $\|.\|$ is the Euclidean norm. The Poincaré Ball model is endowed with the Riemannian metric tensor $g^{\mathbb{B}_c}(x) = (\lambda_x^c)^2 g^\mathbb{E}$, where $x \in \mathbb{B}_c^d$, $\lambda_x = (1 - c\|x\|^2)^{-1}$ is the conformal factor and $g^\mathbb{E} = \mathbb{I}^d$ denotes the Euclidean metric tensor.

Furthermore, the geodesic distance between two points $x, y \in \mathbb{B}_c^d$ is defined as:

$$d_c(x, y) = \frac{2}{\sqrt{c}} (\sqrt{c} \|-x \oplus_c y\|),$$ (5.7)

where $\oplus$ is Möbius addition which has the closed-form expression:

$$x \oplus_c y = \frac{(1 + 2c \langle x, y \rangle + c \|y\|^2)x + (1 - c \|x\|^2)y}{1 + 2c \langle x, y \rangle + c^2 \|x\|^2 \|y\|^2}. \tag{5.8}$$

**Hyperbolic Prototypical Network**  To extend prototypical networks to the Poincaré Ball model and to perform different operations in this hyperbolic space, we first map the Euclidean features to the hyperbolic space, using the exponential map defined as:

$$\exp_z^c(x) = z \oplus_c \left( \tanh \left( \sqrt{c} \frac{\lambda_z^c \|x\|}{2} \right) \frac{x}{\sqrt{c} \|x\|} \right), \tag{5.9}$$

where $z \in \mathbb{B}_c^d$ and $x \in \mathcal{T}_z \mathbb{B}_c^d$. $\mathcal{T}_z \mathbb{B}_c^d \cong \mathbb{R}^d$ is the tangent space at $z$. Here $z$ is the origin of the Poincaré Ball model.

The prototype of class $k$ is then calculated as the average over hyperbolic embeddings of support samples from class $k$. However, there is no closed-form expression of the hyperbolic mean in the Poincaré ball model. We can nevertheless use the one defined in the Klein model since all hyperbolic models are isometric (Peng et al., 2021). The hyperbolic mean is then defined with the Klein coordinates, known as the *Einstein midpoint*, as:

$$\text{HypAve}(x_1, \ldots, x_n) = \sum_{i=1}^{n} \gamma_i x_i \Big/ \sum_{i=1}^{n} \gamma_i, \tag{5.10}$$

where $x_i$ are embeddings in Klein model and $\gamma_i = \left(1 - c \|x_i\|^2\right)^{-1/2}$ are the Lorentz factors.

Therefore, given the support embeddings in the Poincaré Ball $\mathbb{B}_c^d$, we can first map them to the Klein model $\mathbb{K}^d$ as Eq. (5.11), calculate the prototypes using Eq. (5.10), and map them back to the Poincaré model using Eq. (5.12):

$$x_{\mathbb{K}^d} = \frac{2x_{\mathbb{B}_c^d}}{1 + c \left\| x_{\mathbb{B}_c^d} \right\|} \tag{5.11}$$

$$x_{\mathbb{B}_c^d} = \frac{x_{\mathbb{K}^d}}{1 + \sqrt{1 - c \left\| x_{\mathbb{K}^d} \right\|^2}} \tag{5.12}$$

**Feature clipping**  Extending deep neural networks to hyperbolic spaces is a challenging task since generalizing basic operations such as vector addition or matrix-vector multiplication is not trivial (Peng et al., 2021).

Most studies on hyperbolic spaces use hybrid "Euclidean-Hyperbolic" architectures. However, passing between the Euclidean and hyperbolic layers of these hybrid architectures often positions the hyperbolic features close to the space boundary, which usually leads to numerical problems, especially the vanishing gradient phenomenon. In order to tackle this issue, Guo et al. (Y. Guo et al., 2022) proposed to *clip* Euclidean features before moving to the hyperbolic layers, allowing pushing the hyperbolic embeddings further away from the Poincaré boundary. Furthermore, it enhances the performance of hyperbolic networks and makes their behaviour steadier; this clipping is defined as:

$$x_r^E = \min \left\{ 1, \frac{r}{||x^E||} \right\} \cdot x^E, \tag{5.13}$$

with $x_r^E$ the clipped embedding of $x^E$ (in the Euclidean space) and $r$ the clipping value.

### 5.3.2 Experiments

We again conduct experiments on the Resisc-45 scene classification dataset as in the previous section. During training, we randomly sample $500$ episodes in each epoch to learn the model which parameters are updated every 2 episodes, i.e. the batch size is set to 2 and the average loss over the 2 episodes is used to learn the network parameters. The SGD optimiser is used to update the network parameters with momentum set to 0.9 and weight decay set to 0.0005. For the hyperbolic parameters, we use the Riemannian SGD optimiser (Bonnabel, 2013). The learning rate is set to $10^{-4}$. We evaluate the learned model on a validation set after each training epoch by randomly sampling $500$ episodes, the network weights with the highest accuracy over the validation set are retained as the best parameters. For the meta-testing stage, we evaluate the best model on $1000$ randomly sampled episodes from the test set $D_{novel}$. For the hyperbolic hyper-parameters, we follow (Khrulkov et al., 2020) and set the hyperbolic curvature $c$ to 0.01, in both pre-training and meta-training while the chipping value $r$ and the temperature $\tau$ are cross-validated. A sensitivity study of these hyper-parameters is further discussed in our article (J20).

In both 5-way $N$-shot configurations (with $N = 1$ or $5$), we compare our hyperbolic prototypical network (*H-ProtoNet*) not only to the reference model (Khrulkov et al., 2020) but also to the Euclidean prototypical network (Snell et al., 2017) counterpart (*E-ProtoNet*). We also compare with the cosine metric as a similarity function (*C-ProtoNet*) as advocated in a similar context (P. Zhang et al., 2020). We re-implement these methods and cross-validate their hyper-parameters for a fair comparison.

Table 5.2 reports the classification accuracy on various latent space dimensions for 1-shot and 5-shot respectively. We stop at dimension 32 as going further deteriorated the performance considerably. We report accuracy values and the hierarchical precision which is a global hierarchical metric to assess the model's ability to better reflect the semantic relationships between the novel scene categories. We observe that the baseline H-ProtoNet (Khrulkov et al., 2020), which was reported to perform well in the literature, actually performs worse than both Euclidean ProtoNets (E-ProtoNet and C-ProtoNet). However, our H-ProtoNet, which uses the clipping technique, outperforms not only the baseline H-ProtoNet but also both Euclidean ProtoNets. Interestingly, our H-ProtoNet also improves the hierarchical precision over different dimensions of the latent space, especially in the 5-shot scenario. This supports to some extent our hypothesis that hyperbolic geometry, if carefully tackled, better handles data with an underlying hierarchical structure and allows for more meaningful embeddings.

| Approach | Metric | Latent Space Dimension $d$ | | |
|---|---|---|---|---|
| | | 32 | 128 | 512 |
| E-ProtoNet | Acc | $\underline{78.96 \pm 0.36}$ | $80.24 \pm 0.37$ | $77.79 \pm 0.41$ |
| | $P_H$ | $82.19 \pm 0.34$ | $83.52 \pm 0.33$ | $81.26 \pm 0.37$ |
| C-ProtoNet | Acc | $76.76 \pm 0.41$ | $79.76 \pm 0.37$ | $\underline{78.75 \pm 0.40}$ |
| | $P_H$ | $80.43 \pm 0.36$ | $83.03 \pm 0.34$ | $\underline{82.09 \pm 0.36}$ |
| H-ProtoNet* | Acc | $74.31 \pm 0.41$ | $78.45 \pm 0.38$ | $78.24 \pm 0.38$ |
| | $P_H$ | $77.96 \pm 0.39$ | $81.95 \pm 0.36$ | $81.93 \pm 0.33$ |
| H-ProtoNet (ours) | Acc | $\mathbf{79.37 \pm 0.39}$ | $\mathbf{82.75 \pm 0.33}$ | $\mathbf{80.74 \pm 0.37}$ |
| | $P_H$ | $\mathbf{82.79 \pm 0.34}$ | $\mathbf{85.81 \pm 0.29}$ | $\mathbf{84.04 \pm 0.32}$ |

(a) 5-shot results

| Approach | Metric | Latent Space Dimension $d$ | | |
|---|---|---|---|---|
| | | 32 | 128 | 512 |
| E-ProtoNet | Acc | $\underline{62.40 \pm 0.62}$ | $\underline{64.15 \pm 0.65}$ | $63.89 \pm 0.59$ |
| | $P_H$ | $\underline{67.75 \pm 0.59}$ | $69.38 \pm 0.60$ | $\underline{69.27 \pm 0.55}$ |
| C-ProtoNet | Acc | $61.58 \pm 0.65$ | $63.85 \pm 0.63$ | $63.72 \pm 0.62$ |
| | $P_H$ | $67.15 \pm 0.60$ | $69.03 \pm 0.59$ | $69.05 \pm 0.57$ |
| H-ProtoNet* | Acc | $56.89 \pm 0.64$ | $61.69 \pm 0.63$ | $63.48 \pm 0.62$ |
| | $P_H$ | $62.87 \pm 0.62$ | $67.37 \pm 0.60$ | $69.07 \pm 0.57$ |
| H-ProtoNet (ours) | Acc | $\mathbf{63.30 \pm 0.63}$ | $\mathbf{66.05 \pm 0.64}$ | $\mathbf{65.09 \pm 0.63}$ |
| | $P_H$ | $\mathbf{68.69 \pm 0.59}$ | $\mathbf{71.26 \pm 0.58}$ | $\mathbf{70.39 \pm 0.58}$ |

(b) 1-shot results

**Table 5.2:** Few-shot classification results computed on the Resisc-45 dataset. All accuracy results are averaged over 1000 test episodes and are reported with a 95% confidence interval. * refers to the baseline model (Khrulkov et al., 2020).

## 5.4 Conclusion

This chapter has presented our two contributions to leverage the prototypical network for few-shot scene image classification. As the first contribution, we propose a hierarchical loss-based approach that leverages the explicit class hierarchy through hierarchical prototypes for few-shot scene classification. Experiments show that the proposed method outperforms flat baselines, demonstrating the effectiveness of using class hierarchy for remote sensing scene classification. The second contribution focuses on leveraging implicit hierarchical information through hyperbolic prototypical networks. Our experiments then confirm the suitability of hyperbolic space for remote sensing scene classification and emphasize the importance of the clipping technique to achieve its effectiveness in this domain.

The second contribution in this chapter serves as an opening towards the application of hyperbolic space to remote sensing images. Although challenging, the properties of remote sensing images align well with the geometric properties of hyperbolic space, presenting a promising avenue for future research in this field. As perspective, an extension of hierarchical prototypes to hyperbolic space seems to be an intriguing pursuit. However, an exploratory search for hyper-parameters that accentuates the hierarchical loss in the hyperbolic space and allows reaching the optimal performance should be carried out.

# Chapter 6

# Perspectives

The previous four chapters have highlighted my contributions to the topic of label-efficient learning applied to computer vision and remote sensing, with the primary goal to reduce the dependence of deep learning frameworks on large annotated datasets. Yet, such an approach still faces **several challenges**.

- Firstly, leveraging the abundance of unlabeled but highly varied and complex data remains challenging, particularly with the presence of domain shifts due to various data acquisition conditions. Within the SEMMACAPE research project (2019-2022), we observed that the developed VAE-based weakly-supervised object detection models presented in Chapter 2, while achieving strong performance on public vision benchmarks and small aerial datasets, struggled when applied to large-scale aerial image data acquired from different seasons or geographic areas. Likewise, the multi-task frameworks we developed to jointly learn from multiple partially annotated datasets presented in Chapter 3, despite their good results on the studied vision benchmarks such as Pascal VOC and COCO, might fail to maintain such positive behaviors when applied to remote sensing datasets with significant domain variability.

- The second challenge arises from the integration of multimodal data in label-efficient learning methods. Since adding more modalities also increases the complexity of the model, more advanced fusion techniques are required to effectively extract and combine information from each modality (Liang et al., 2024). Moreover, effectively balancing different modalities to ensure the model learns good and complementary representations from all modalities remains a challenging task to be explored. For instance, in our joint self-supervised learning of optical and SAR images presented in Chapter 4, the two data sources are still processed in a similar way, which may be not optimal due to their differences in sensor characterizations. Developing methodology to address the variations in resolution, noise levels and the intrinsic properties of each modality remains an open challenge.

- Thirdly, methods like self-supervised or semi-supervised learning often come with high computational costs to learn representations (Ericsson et al., 2022), potentially limiting the advantage of reducing the requirement of labeled data. Indeed, training large models typically involves processing vast amounts of unlabeled data with large batch sizes, thus requiring substantial computational resources and energy. To illustrate, during the PhD work of Paul Berg (2021-2024), it was measured that conducting self-supervised pre-training on the COCO dataset (328K images) using a dense contrastive approach (X. Wang et al., 2021) took several days to one week using four A100 GPUs. This raises a critical concern regarding the environmental impact, since the carbon footprint associated with training such large models is becoming a major challenge in AI and deep learning research.

In the coming years, I plan to continue my research on topics related to resource-efficient deep learning tailored to the domains of computer vision and remote sensing applied to Earth and environment observation. My focus will be on developing models that maintain high performance while minimizing the requirement on both labeled data and computational cost. This approach is particularly crucial for deploying deep learning solutions in scenarios where time and energy efficiency are critical, such as in remote sensing-based rapid mapping and edge computing. The goal is to create more sustainable and scalable models that are practical for real-world applications. In addition, my work will focus on improving model robustness in multi-modal and multi-sensor settings, with particular attention on privacy-preserving techniques to ensure that models are both effective and secure across diverse and sensitive applications. To this end, below are the three main research directions I will focus on as short- and mid-term research perspectives:

**Resource-efficient foundation models in Earth observation**    The rapid progress of self-supervised learning in remote sensing has initiated the advent of foundation models (FMs) in the field. FMs emerge as pre-trained generic models that excel in a wide range of downstream tasks. These models, characterized by extensive parameters and pretrained on huge volumes of large-scale datasets, have demonstrated remarkable performance across various Earth observation applications (Lu et al., 2024; Xiao et al., 2024). Most of them have focused on building huge-size pretrained models (billions of parameters) using an abundance of multi-modal RS data (i.e., aerial and satellite, optical and SAR, low to very high resolution). Consequently, they demand significant computational resources for training and deployment. For instance, SkySense (2.06 billion parameters) has been pretrained using 80 A100-80GB GPUs in 12.8 days (X. Guo et al., 2024). Therefore, there is an urgent need to develop methods capable of deploying these FMs in downstream Earth observation applications while simultaneously reducing their size and computational cost. One of my primary objectives is to develop models that can maintain high performance while minimizing the need for annotated data and computational costs. This approach is particularly crucial for deploying AI solutions in contexts where time and energy efficiency are essential, such as remote sensing or embedded applications. This line of research was initiated with the PhD work of Pierre Adorni (started in October 2024), which aims to develop efficient FMs for VHR satellite images, in collaboration with the French National Center for Space Studies (CNES). Furthermore, within the framework of the ANR JCJC call for projects (national equivalent to ERC Starting Grant program), I submitted a project named DREAMS, which aims to develop resource-efficient and scalable AI models for rapid mapping wildfires and landslides from satellite imagery. If selected, this project will enrich and strengthen my contributions which are well aligned with this research direction.

**Open-world understanding with multi-modal learning**    My second research focus is on designing and developing robust learning methods for data from multimodal sensors, particularly in low-supervision contexts. I am particularly interested in efficient fusion techniques to leverage the complementary information from different modalities, such as visible and thermal data in computer vision, or optical and radar data in remote sensing. Additionally, I plan to extend my research beyond image data by integrating other modalities such as text or sound, to address scenarios where the diversity of information sources can compensate for the lack of annotations in weakly supervised settings. This research path was initiated with the PhD work of Manuel Nkegoum (started in November 2024), focusing on object detection from few examples of visible and thermal images. With the evolution of current vision-language large models (J. Zhang et al., 2024), the integration of text modality is expected to go towards to zero-shot and few-shot learning of unseen objects in complex scenarios. Indeed, by leveraging the rich textual semantics to complement visual information, these models can generalize better to object categories without requiring extensive labeled data for fine-tuning. Furthermore, in the context of the European Horizon AXOLOTL project (2024-2027) dedicated to AI solutions for marine and maritime applications, we aim at developing deep learning-based vessel detection frameworks using data coming from various sources, including low- and high-resolution satellite images (e.g., Sentinel-2 and PlanetScope images), AIS (Automatic Identification Systems) signals, and passive acoustic data. For instance, one of the objectives of the project is to perform super-resolution on Sentinel-2 images (10 meters) to match the resolution of PlanetScope (3 meters), allowing us to combine the high revisit frequency and broad coverage of Sentinel-2 and the fine spatial details of PlanetScope to enhance vessel detection.

**Cross-domain and privacy-preserving federated learning**    In the context that the development of AI models involves an abundance of multiple data sources, it is essential to build approaches that not only achieve high performance but also ensure data confidentiality. My third research line focuses on exploring reliable and trustworthy AI methods that ensure data privacy and security in both training and deployment phases. Recently, we have started working with privacy-preserving federated learning (FL) methods (Yin et al., 2021) which allow training AI models without sharing data on a centralized server. This work was conducted within the Master internship of Anh-Kiet Duong (March to August 2024) with the goal to investigate and benchmark FL frameworks in two remote sensing scenarios: scene classification and ship detection from multi-source satellite images. With the aim to improve the privacy-preserving property of FL methods, we developed a feature communication strategy published in (C38) which allows us to not only reduce the amount of transmitted information (i.e., by sending features and not the model weights), but also limit the membership inference attacks (MIA) (Bai et al., 2024). Currently, I continue to collaborate with researchers at the Luxembourg Institute of Science and Technology (LIST) on this FL direction with a main focus on ship detection using image data acquired from different optical sensors (commercial or public), with various spatial resolutions (low to very high), metocean conditions, and regions of interest. A preliminary work published in (J22) has demonstrated the effectiveness of FL approaches, particularly in scenarios where data cannot be shared. As an on-going work, we are exploring other FL models and configurations, including varying the number of clients, local epochs and rounds, as well as asynchronous FL methods for more comprehensive insights. For short-term and mid-term perspectives, we plan to investigate domain adaptation techniques to better perform cross-domain and cross-sensor federated learning in order to address larger and more diverse remote sensing applications. To this end, we aim to submit a future European-scale project to strengthen our collaboration in this area.

# List of Publications

Here is the list of my publications from which the names of authors I directly supervised are <u>underlined</u>.

**Peer-reviewed journal papers**

**(J26)** Berg, P., Buecher, L., Michele, B., **Pham, M. T.**, Chapel, L., & Courty, N. (2025). Multi-prototype hyperbolic learning guided by class hierarchy. In International Journal of Computer Vision (IJCV). Accepted.

**(J25)** Belmouhcine, A., **Pham, M. T.**, & Lefèvre, S. (2025). YOLO-G3CF: Gaussian contrastive cross-channel fusion for multimodal object detection. IEEE Geoscience and Remote Sensing Letters, 22, 8002005. (Belmouhcine et al., 2025)

**(J24)** Hamard, Q., **Pham, M. T.**, Cazau, D., & Heerah, K. (2024) A deep learning model for detecting and classifying multiple marine mammal species from passive acoustic data. Ecological Informatics, 102906. (Hamard et al., 2024)

**(J23)** Gangloff, H., **Pham, M. T.**, Courtrai, L., & Lefèvre, S. (2024). Variational autoencoder with gaussian random field prior: application to unsupervised animal detection in aerial images. ISPRS Journal of Photogrammetry and Remote Sensing, 218: 600-609. (Gangloff et al., 2024)

**(J22)** Duong, A. K., La, T. V., Lê, H. Â., & **Pham, M. T.** (2024) FedShip: Federated learning for ship detection from multi-source remote sensing images. IEEE Geoscience and Remote Sensing Letters, 22, 6002005. (Duong, La, et al., 2024)

**(J21)** Berg, P., Uzun, B., **Pham, M. T.**, & Courty, N. (2024). Multimodal supervised contrastive learning in remote sensing downstream tasks. IEEE Geoscience and Remote Sensing Letters, 21, 6007405. (Berg, Uzun, et al., 2024)

**(J20)** Hamzaoui, M., Chapel, L., **Pham, M. T.**, & Lefèvre, S. (2024). Hyperbolic prototypical network for few shot remote sensing scene classification. Pattern Recognition Letters, 177, 151-156. (Hamzaoui et al., 2024)

**(J19)** Berg, P., **Pham, M. T.**, & Courty, N. (2022). Self-supervised learning for scene classification in remote sensing: Current state of the art and perspectives. Remote Sensing, 14(16), 3995. (Berg, Pham, et al., 2022)

**(J18)** Lê, H. Â., Zhang, H., **Pham, M. T.**, & Lefèvre, S. (2022). Mutual guidance meets supervised contrastive learning: Vehicle detection in remote sensing images. Remote Sensing, 14(15), 3689. (Lê et al., 2022)

**(J17)** Lê, H. Â., Guiotte, F., **Pham, M. T.**, Lefèvre, S., & Corpetti, T. (2022). Learning digital terrain models from point clouds: ALS2DTM dataset and rasterization-based gan. IEEE Journal of Selected Topics in Applied Earth Observations and Remote Sensing, 15, 4980-4989. (Le et al., 2022)

**(J16)** Maia, D., **Pham, M. T.**, & Lefèvre, S. (2022). Watershed-based attribute profiles with semantic prior knowledge for remote sensing image analysis. IEEE Journal of Selected Topics in Applied Earth Observations and Remote Sensing, 15, 2574-2591. (Maia et al., 2022)

**(J15)** Berg, P., Santana Maia, D., **Pham, M. T.**, & Lefèvre, S. (2022). Weakly supervised detection of marine animals in high resolution aerial images. Remote Sensing, 14(2), 339. (Berg, Santana Maia, et al., 2022)

**(J14)** Santana Maia, D., **Pham, M. T.**, Aptoula, E., Guiotte, F., & Lefèvre, S. (2021). Classification of remote sensing data with morphological attribute profiles: A decade of advances. IEEE Geoscience and Remote Sensing Magazine, 9(3), 43-71. (Maia et al., 2021)

**(J13)** Courtrai, L., **Pham, M. T.**, & Lefèvre, S. (2020). Small object detection in remote sensing images based on super-resolution with auxiliary generative adversarial networks. Remote Sensing, 12(19), 3152. (Courtrai, Pham, and Lefèvre, 2020)

**(J12)** **Pham, M. T.**, Courtrai, L., Friguet, C., Lefèvre, S., & Baussard, A. (2020). YOLO-Fine: One-stage detector of small objects under various backgrounds in remote sensing images. Remote Sensing, 12(15), 2501. (M.-T. Pham et al., 2020)

**(J11)** Guiotte, F., **Pham, M. T.**, Dambreville, R., Corpetti, T., & Lefèvre, S. (2020). Semantic segmentation of lidar points clouds: rasterization beyond digital elevation models. IEEE Geoscience and Remote Sensing Letters, 17(11), 2016-2019. (Guiotte et al., 2020)

**(J10)** **Pham, M. T.** (2018). Fusion of polarimetric features and structural gradient tensors for VHR PolSAR image classification. IEEE Journal of Selected Topics in Applied Earth Observations and Remote Sensing, 11(10), 3732-3742. (M.-T. Pham, 2018b)

**(J09)** **Pham, M. T.**, Lefèvre, S., & Merciol, F. (2018). Attribute profiles on derived textural features for highly textured optical image classification. IEEE Geoscience and Remote Sensing Letters, 15(7), 1125-1129. (M.-T. Pham, Lefèvre, and Merciol, 2018)

**(J08)** **Pham, M. T.**, Aptoula, E., & Lefevre, S. (2018). Feature profiles from attribute filtering for remote sensing image classification. IEEE Journal of Selected Topics in Applied Earth Observations and Remote Sensing, 11(1), 249-256. (M. Pham et al., 2018)

**(J07)** **Pham, M. T.**, Lefevre, S., & Aptoula, E. (2017). Local feature-based attribute profiles for optical remote sensing image classification. IEEE Transactions on Geoscience and Remote Sensing, 56(2), 1199-1212. (M.-T. Pham, Lefevre, and Aptoula, 2017)

**(J06)** **Pham, M. T.**, Mercier, G., & Bombrun, L. (2017). Color texture image retrieval based on local extrema features and riemannian distance, Journal of Imaging, 3(4), 43. (M.-T. Pham, Mercier, and Bombrun, 2017)

**(J05)** **Pham, M. T.**, Mercier, G., Regniers, O., & Michel J. (2016). Texture retrieval from VHR optical remote sensed images using the local extrema descriptors with application to vineyard parcel detection, Remote Sensing, 8(5), 368. (M.-T. Pham, Mercier, Regniers, et al., 2016)

**(J04) Pham, M. T.**, Mercier, G., & Michel J. (2016). PW-COG: an effective texture descriptor for VHR satellite imagery using a pointwise approach on covariance matrix of oriented gradients, IEEE Transactions on Geoscience and Remote Sensing, vol. 54(6), pp. 3345-3359. (M.-T. Pham, Mercier, and Michel, 2016)

**(J03) Pham, M. T.**, Mercier, G., & Michel J. (2015). Change detection between SAR images using a pointwise approach and graph theory, IEEE Transactions on Geoscience and Remote Sensing, vol. 54(4), pp. 2020-2032. (M.-T. Pham et al., 2015a)

**(J02) Pham, M. T.**, Mercier, G., & Michel J. (2015). Pointwise graph-based local texture characterization for very high resolution multispectral image classification, IEEE Journal of Selected Topics in Applied Earth Observations and Remote Sensing, vol. 8(5), pp. 1962-1973. (M.-T. Pham et al., 2015d)

**(J01) Pham, M. T.**, Mercier, G., & Michel J. (2014). Textural features from wavelets on graphs for very high resolution panchromatic Pléiades image classification, The French Journal of Photogrammetry and Remote Sensing, n. 208, pp. 131-136. (M. T. Pham et al., 2014)

**International conference papers**

**(C43)** Adorni, P., **Pham, M. T.**, May, S., & Lefèvre, S. (2025). Towards Efficient Benchmarking of Foundation Models in Remote Sensing: A Capabilities Encoding Approach. In Proceedings of the Computer Vision and Pattern Recognition Conference (pp. 3096-3106). (Adorni et al., 2025)

**(C42)** Rodriguez, M., **Pham, M. T.**, Sudmanns, M., Poterek, Q. & Narvaez, O. (2025). Enhancing deep learning performance on burned area delineation from SPOT-6/7 imagery for emergency management. In IGARSS 2025-2025 IEEE International Geoscience and Remote Sensing Symposium. Accepted.

**(C41)** La, T. V., **Pham, M. T.**, Li, Y., Matgen, P. & Chini, M. (2025). Adaptive federated learning for ship detection across diverse satellite imagery sources. In IGARSS 2025-2025 IEEE International Geoscience and Remote Sensing Symposium. Accepted.

**(C40)** Berg, P., Michele, B., **Pham, M. T.**, Chapel, L., & Courty, N. (2024). Horospherical Learning with Smart Prototypes. In Bristish Machine Vision Conference (BMVC). (Berg, Michele, et al., 2024)

**(C39)** Lê, H. Â., P. Berg, & **Pham, M. T.** (2024). Box for Mask and Mask for Box: weak losses for multi-task partially supervised learning. In Bristish Machine Vision Conference (BMVC). (Lê et al., 2024)

**(C38)** Duong, A. K., Lê, H. Â., & **Pham, M. T.** (2024). Leveraging feature communication in federated learning for remote sensing image classification. In IGARSS 2024-2024 IEEE International Geoscience and Remote Sensing Symposium (pp. 6890-6894). IEEE. (Duong, Lê, et al., 2024)

**(C37)** La, T. V., **Pham, M. T.**, & Chini, M. (2024). Insight Into the Collocation of Multi-Source Satellite Imagery for Multi-Scale Vessel Detection. In IGARSS 2024-2024 IEEE International Geoscience and Remote Sensing Symposium (pp. 7396-7400). IEEE. (La et al., 2024b)

**(C36)** Pande, S., Uzun, B., Guiotte, F., **Pham, M. T.**, Corpetti, T., Delerue, F., & Lefèvre, S. (2024). Plant detection from ultra high resolution remote sensing images: A semantic segmentation approach based on fuzzy loss. In IGARSS 2024-2024 IEEE International Geoscience and Remote Sensing Symposium (pp. 5213-5217). IEEE. (Pande et al., 2024)

**(C35)** Uzun, B., Pande, S., Cachin-Bernard, G., **Pham, M. T.**, Lefèvre, S., Blatrix, R., & McKey, D. (2024). Mapping Earth mounds from space. In IGARSS 2024-2024 IEEE International Geoscience and Remote Sensing Symposium (pp. 4044-4049). IEEE. (Uzun et al., 2024)

**(C34)** Lê, H. Â., & **Pham, M. T.** (2024). Leveraging knowledge distillation for partial multi-task learning from multiple remote sensing datasets. In IGARSS 2024-2024 IEEE International Geoscience and Remote Sensing Symposium (pp. 8019-8023). IEEE. (Lê and Pham, 2024)

**(C33)** Lê, H. Â., & **Pham, M. T.** (2023). Data exploitation: multi-task learning of object detection and semantic segmentation on partially annotated data. In Bristish Machine Vision Conference (BMVC). (Lê and Pham, 2023a)

**(C32)** Lê, H. Â., & **Pham, M. T.** (2023). Self-training and multi-task learning for limited data: evaluation study on object detection. In Proceedings of the IEEE/CVF International Conference on Computer Vision (pp. 1003-1009). (Lê and Pham, 2023c)

**(C31)** Luces, O. D. R. N., Pham, M. T., Poterek, Q., & Braun, R. (2023). Burnt area extraction from high-resolution satellite images based on anomaly detection. In ECML/PKDD Workshop on Machine Learning for Earth Observation (MACLEAN). (Narvaez Luces et al., 2023)

**(C30)** Gangloff, H., **Pham, M. T.**, Courtrai, L., & Lefèvre, S. (2023). Unsupervised Anomaly Detection Using Variational Autoencoder with Gaussian Random Field Prior. In 2023 IEEE International Conference on Image Processing (ICIP) (pp. 1620-1624). IEEE. (Gangloff et al., 2023)

**(C29)** Bonet, C., Berg, P., Courty, N., Septier, F., Drumetz, L., & **Pham, M. T.** (2023). Spherical Sliced-Wasserstein. In International Conference on Learning Representations (ICLR). (Bonet et al., 2022)

**(C28)** **Pham, M. T.**, Gangloff, H., & Lefèvre, S. (2023). Weakly supervised marine animal detection from remote sensing images using vector-quantized variational autoencoder. In IGARSS 2023-2023 IEEE International Geoscience and Remote Sensing Symposium (pp. 5559-5562). IEEE. (M.-T. Pham et al., 2023)

**(C27)** Hamzaoui, M., Chapel, L., **Pham, M. T.**, & Lefèvre, S. (2023). Hyperbolic Variational Auto-Encoder for Remote Sensing Scene Embeddings. In IGARSS 2023-2023 IEEE International Geoscience and Remote Sensing Symposium (pp. 5391-5394). IEEE. (Hamzaoui et al., 2023)

**(C26)** Belmouhcine, A., Burnel, J. C., Courtrai, L., **Pham, M. T.**, & Lefèvre, S. (2023). Multimodal Object Detection in Remote Sensing. In IGARSS 2023-2023 IEEE International Geoscience and Remote Sensing Symposium (pp. 1245-1248). IEEE. (Belmouhcine et al., 2023)

**(C25)** Berg, P., **Pham, M. T.**, & Courty, N. (2023). Joint multi-modal self-supervised pre-training in remote sensing: Application to methane source classification. In IGARSS 2023-2023 IEEE International Geoscience and Remote Sensing Symposium (pp. 6624-6627). IEEE. (Berg et al., 2023)

**(C24)** Lê, H. Â., & **Pham, M. T.** (2023). Knowledge distillation for object detection: from generic to remote sensing datasets. In IGARSS 2023-2023 IEEE International Geoscience and Remote Sensing Symposium (pp. 6194-6197). IEEE. (Lê and Pham, 2023b)

**(C23)** Singh, T., Gangloff, H., & **Pham, M. T.** (2023). Object counting from aerial remote sensing images: application to wildlife and marine mammals. In IGARSS 2023-2023 IEEE International Geoscience and Remote Sensing Symposium (pp. 6580-6583). IEEE. (Singh et al., 2023)

**(C22)** Hamzaoui, M., Chapel, L., **Pham, M. T.**, & Lefèvre, S. (2022). A hierarchical prototypical network for few-shot remote sensing scene classification. In International Conference on Pattern Recognition and Artificial Intelligence (pp. 208-220). Cham: Springer International Publishing. (Hamzaoui et al., 2022)

**(C21)** Gangloff, H., **Pham, M. T.**, Courtrai, L., & Lefèvre, S. (2022). Leveraging vector-quantized variational autoencoder inner metrics for anomaly detection. In 2022 26th International Conference on Pattern Recognition (ICPR) (pp. 435-441). IEEE. (Gangloff et al., 2022b)

**(C20)** Santana Maia, D., **Pham, M. T.**, & Lefèvre, S. (2021). Watershed-based attribute profiles for pixel classification of remote sensing data. In International Conference on Discrete Geometry and Mathematical Morphology (pp. 120-133). Cham: Springer International Publishing. (Santana Maia et al., 2021)

**(C19)** **Pham, M. T.**, & Lefevre, S. (2021). Very high resolution airborne polsar image classification using convolutional neural networks. In EUSAR 2021; 13th European Conference on Synthetic Aperture Radar (pp. 1-4). VDE. (M.-T. Pham and Lefevre, 2021)

**(C18)** Courtrai, L., **Pham, M. T.**, Friguet, C., & Lefèvre, S. (2020). Small object detection from remote sensing images with the help of object-focused super-resolution using Wasserstein GANs. In IGARSS 2020-2020 IEEE International Geoscience and Remote Sensing Symposium (pp. 260-263). IEEE. (Courtrai, Pham, and Lefèvre, 2020)

**(C17)** Pirrone, D., & **Pham, M. T.** (2020). A Compound Polarimetric-Textural Approach for Unsupervised Change Detection in Multi-Temporal Full-Pol SAR Imagery. In IGARSS 2020-2020 IEEE International Geoscience and Remote Sensing Symposium (pp. 316-319). IEEE. (Pirrone and Pham, 2020)

**(C16)** Froidevaux, A., Julier, A., Lifschitz, A., **Pham, M. T.**, Dambreville, R., Lefèvre, S., Lassalle, P., & Huynh, T. L. (2020). Vehicle detection and counting from VHR satellite images: efforts and open issues. In IGARSS 2020-2020 IEEE International Geoscience and Remote Sensing Symposium (pp. 256-259). IEEE. (Froidevaux et al., 2020)

**(C15)** Osio, A., **Pham, M. T.**, & Lefèvre, S. (2020). Spatial processing of sentinel imagery for monitoring of acacia forest degradation in lake nakuru riparian reserve. ISPRS Annals of Photogrammetry, Remote Sensing and Spatial Information Sciences, 3, 525-532. (Osio et al., 2020)

**(C14)** Merciol, F., **Pham, M. T.**, Santana Maia, D., Masse, A., & Sannier, C. (2020). BROCELIANDE: a comparative study of attribute profiles and feature profiles from different attributes. The International Archives of the Photogrammetry, Remote Sensing and Spatial Information Sciences, 43, 1371-1377. (Merciol et al., 2020)

**(C13)** **Pham, M. T.** (2018). Efficient texture retrieval using multiscale local extrema descriptors and covariance embedding. In Proceedings of the European Conference on Computer Vision (ECCV) Workshops. (M.-T. Pham, 2018a)

**(C12)** **Pham, M. T.**, Aptoula, E., & Lefèvre, S. (2018). Classification of remote sensing images using attribute profiles and feature profiles from different trees: a comparative study. In IGARSS 2018-2018 IEEE International Geoscience and Remote Sensing Symposium (pp. 4511-4514). IEEE. (M.-T. Pham, Aptoula, et al., 2018)

**(C11)** **Pham, M. T.**, & Lefèvre, S. (2018). Buried object detection from B-scan ground penetrating radar data using Faster-RCNN. In IGARSS 2018-2018 IEEE international geoscience and remote sensing symposium (pp. 6804-6807). IEEE. (M.-T. Pham and Lefèvre, 2018a)

**(C10)** **Pham, M. T.**, Lefèvre, S., Aptoula, E., & Bruzzone, L. (2018). Recent developments from attribute profiles for remote sensing image classification. In Proceedings of the International Conference on Pattern Recognition and Artificial Intelligence (pp. 102-107). CENPARMI. (M.-T. Pham, Lefèvre, Aptoula, and Bruzzone, 2018)

**(C09)** **Pham, M. T.**, Mercier, G., Trouvé, E., & Lefèvre, S. (2017). SAR image texture tracking using a pointwise graph-based model for glacier displacement measurement. In 2017 IEEE International Geoscience and Remote Sensing Symposium (IGARSS) (pp. 1083-1086). IEEE. (M.-T. Pham, Mercier, Trouvé, et al., 2017)

**(C08)** **Pham, M. T.**, Lefèvre, S., Aptoula, E., & Damodaran, B. B. (2017). Classification of VHR remote sensing images using local feature-based attribute profiles. In 2017 IEEE International Geoscience and Remote Sensing Symposium (IGARSS) (pp. 747-750). IEEE. (M.-T. Pham, Lefèvre, Aptoula, and Damodaran, 2017)

**(C07)** Aptoula, E., **Pham, M. T.**, & Lefèvre, S. (2017). Quasi-flat zones for angular data simplification. In International Symposium on Mathematical Morphology and Its Applications to Signal and Image Processing (pp. 342-354). Cham: Springer International Publishing. (Aptoula et al., 2017)

**(C06)** **Pham, M. T.**, Mercier, G., Regniers, O., Bombrun, L., & Michel, J. (2016). Texture retrieval from very high resolution remote sensing images using the local extrema-based descriptors. In 2016 IEEE International Geoscience and Remote Sensing Symposium (IGARSS), pp. 1839-1842. IEEE. (M.-T. Pham, Mercier, Regniers, et al., 2016)

**(C05)** **Pham, M. T.**, Mercier, G., & Michel, J. (2015). Pointwise approach on covariance matrix of oriented gradients for very high resolution image texture segmentation. In 2015 IEEE International Geoscience and Remote Sensing Symposium (IGARSS), pp. 1008-1011. IEEE. (M.-T. Pham et al., 2015d)

**(C04)** **Pham, M. T.**, Mercier, G., & Michel, J. (2015). Covariance-based texture description from weighted coherence matrix and gradient tensors for polarimetric SAR image classification. In 2015 IEEE International Geoscience and Remote Sensing Symposium (IGARSS), pp. 2469-2472. IEEE. (M.-T. Pham et al., 2015b)

**(C03)** **Pham, M. T.**, Mercier, G., & Michel, J. (2014). A keypoint approach for change detection between SAR images based on graph theory. In 2015 8th International Workshop on the Analysis of Multitemporal Remote Sensing Images (Multi-Temp) (pp. 1-4). IEEE. (M.-T. Pham et al., 2015c)

**(C02)** **Pham, M. T.**, Mercier, G., & Michel, J. (2014). Wavelets on graphs for very high resolution multispectral image texture segmentation. In 2014 IEEE International Geoscience and Remote Sensing Symposium (IGARSS), pp. 2273-2276. IEEE. (M.-T. Pham et al., 2014)

**(C01)** **Pham, M. T.**, & Gueriot, D. (2013). Guided block-matching for sonar image registration using unsupervised kohonen neural networks. In 2013 OCEANS-San Diego (pp. 1-5). IEEE. (M. T. Pham and Gueriot, 2013)

**French conference papers**

**(C05fr)** Berg, P., **Pham, M. T.**, & Courty, N. (2024). Apprentissage contrastif multi-modal: Du pré-entraînement auto-supervisé à la classification supervisée. In RFIAP. (Berg, Pham, et al., 2024)

**(C04fr)** Gangloff, H., **Pham, M. T.**, Courtrai, L., & Lefèvre, S. (2022). Autoencodeurs variationnels à registre de vecteurs pour la détection d'anomalies. In RFIAP. (Gangloff et al., 2022a)

**(C03fr)** Hamzaoui, M., Chapel, L., **Pham, M. T.**, & Lefèvre, S. (2021). Hyperbolic variational auto-encoder for remote sensing scene classification. In ORASIS. (Hamzaoui et al., 2021)

**(C02fr)** Courtrai, L., **Pham, M. T.**, Burnel, J. C., & Lefèvre, S. (2020). Apprentissage de réseaux de neurones de super-résolution pour la détection d'objets de petite taille dans les images de télédétection. In RFIAP. (Courtrai, Pham, Burnel, et al., 2020)

**(C01fr)** **Pham, M. T.**, & Lefèvre, S. (2018). Détection d'objets enterrés par apprentissage profond sur imagerie géoradar. In RFIAP. (M.-T. Pham and Lefèvre, 2018b)

## Communications without proceedings

**(Com05)** La, T. V., **Pham, M. T.**, & Chini, M. (2024). Collocation of multi-source satellite imagery for ship detection based on deep learning models, The European Geosciences Union (EGU) General Assembly. (La et al., 2024a)

**(Com04)** Nguyen, H.H., Nguyen, C.N., Dao, X.T., Duong, Q.T., Pham, T.K.D. & **Pham, M. T.** (2024). Variational autoencoder for anomaly detection: A comparative study, IEEE Int. Conf. on Communications and Electronics (ICCE). (Nguyen et al., 2024)

**(Com03)** Hamard, Q., **Pham, M. T.**, Heerah, K., & Cazau, D. (2023). Deep learning for marine mammal monitoring from underwater acoustic data at offshore windfarm scale, The 7th Conference on Wind energy and Wildlife impacts (CWW).

**(Com02)** Viain, A., Michel, S., Duclos, G., Allain, P., Lefèvre, S., **Pham, M. T.**, Heerah, K., & Rouyer, T. (2023). SEMMACAPE: aerial survey of the marine megafauna in offshore windfarms by automatic characterisation, The 7th CWW.

**(Com01)** Lê, H. Â. **Pham, M. T.**, Lefèvre, S., Heerah, K., & Cazau, D. (2022). Performance of off-the-shelf computer vision methods on spectrogram-based marine mammal detection, The DLDCE Workshop.

## Chapters

**(Ch02)** Lefèvre., S., Courtrai, L., **Pham, M. T.**, Friguet, C., Burnel, J. C. (2021). Observation de la mer par apprentissage profond : quelques exemples d'applications pour protéger notre bien commun. (Lefèvre et al., 2021)

**(Ch01)** **Pham, M. T.**, Mercier, G. (2021). Graph of Characteristic Points for Texture Tracking: Application to Change Detection and Glacier Flow Measurement from SAR Images. Change Detection and Image Time Series Analysis 1 : Unsupervised Methods (2021) : 167-200, Wiley Online Library. (M.-T. Pham and Mercier, 2021)

## Theses

**(T02)** **Pham, M. T.** (2016). Pointwise approach for texture analysis and characterization from very high resolution images, PhD. Thesis. (M. T. Pham, 2016)

**(T01)** **Pham, M. T.** (2013). Study of the joint behavior of 2D/3D descriptors for heterogeneous registration of LiDAR point clouds and optical images, Master thesis.

# List of Figures

# List of Tables

# Bibliography

Adorni, P., Pham, M.-T., May, S., & Lefèvre, S. (2025). Towards efficient benchmarking of foundation models in remote sensing: A capabilities encoding approach. *Proceedings of the Computer Vision and Pattern Recognition Conference*, 3096–3106.

Aptoula, E., Pham, M.-T., & Lefèvre, S. (2017). Quasi-flat zones for angular data simplification. *International Symposium on Mathematical Morphology and Its Applications to Signal and Image Processing*, 342–354.

Bai, L., Hu, H., Ye, Q., Li, H., Wang, L., & Xu, J. (2024). Membership inference attacks and defenses in federated learning: A survey. *ACM Computing Surveys*, *57*(4), 1–35.

Bardes, A., Ponce, J., & LeCun, Y. (2022). Vicreg: Variance-invariance-covariance regularization for self-supervised learning. *International Conference on Learning Representations*.

Baur, C., Denner, S., Wiestler, B., Navab, N., & Albarqouni, S. (2021). Autoencoders for unsupervised anomaly segmentation in brain MR images: A comparative study. *Medical Image Analysis*, 101952.

Belmouhcine, A., Burnel, J.-C., Courtrai, L., Pham, M.-T., & Lefèvre, S. (2023). Multimodal object detection in remote sensing. *IGARSS 2023-2023 IEEE International Geoscience and Remote Sensing Symposium*, 1245–1248.

Belmouhcine, A., Pham, M.-T., & Lefèvre, S. (2025). Yolo-g3cf: Gaussian contrastive cross-channel fusion for multimodal object detection. *IEEE Geoscience and Remote Sensing Letters*.

Berg, P., Michele, B., Pham, M.-T., Chapel, L., & Courty, N. (2024). Horospherical learning with smart prototypes. *British Machine Vision Conference (BMVC)*.

Berg, P., Pham, M.-T., & Courty, N. (2024). Apprentissage contrastif multi-modal: Du préentrainement auto-supervisé à la classification supervisée. *Joint CAP and RFIAP 2024 Conferences*.

Berg, P., Pham, M.-T., & Courty, N. (2023). Joint multi-modal self-supervised pre-training in remote sensing: Application to methane source classification. *IGARSS 2023-2023 IEEE International Geoscience and Remote Sensing Symposium*, 6624–6627.

Berg, P., Pham, M.-T., & Courty, N. (2022). Self-supervised learning for scene classification in remote sensing: Current state of the art and perspectives. *Remote Sensing*, *14*(16), 3995.

Berg, P., Santana Maia, D., Pham, M.-T., & Lefèvre, S. (2022). Weakly supervised detection of marine animals in high resolution aerial images. *Remote Sensing*, *14*(2), 339.

Berg, P., Uzun, B., Pham, M.-T., & Courty, N. (2024). Multimodal supervised contrastive learning in remote sensing downstream tasks. *IEEE Geoscience and Remote Sensing Letters*, *21*, 1–5.

Bergmann, P., Fauser, M., Sattlegger, D., & Steger, C. (2019). MVTec AD–A comprehensive real-world dataset for unsupervised anomaly detection. *IEEE/CVF Conference on Computer Vision and Pattern Recognition*, 9592–9600.

Bertinetto, L., Mueller, R., Tertikas, K., Samangooei, S., & Lord, N. A. (2020). Making better mistakes: Leveraging class hierarchies with deep networks. *Proceedings of the IEEE/CVF Conference on Computer Vision and Pattern Recognition*, 12506–12515.

Beyer, L., Zhai, X., Royer, A., Markeeva, L., Anil, R., & Kolesnikov, A. (2022). Knowledge distillation: A good teacher is patient and consistent. *Proceedings of the IEEE/CVF Conference on Computer Vision and Pattern Recognition*, 10925–10934.

Bond-Taylor, S., Leach, A., Long, Y., & Willcocks, C. G. (2021). Deep generative modelling: A comparative review of VAEs, GANs, normalizing flows, energy-based and autoregressive models. *IEEE Transactions on Pattern Analysis and Machine Intelligence*, *44*(11), 7327–7347.

Bonet, C., Berg, P., Courty, N., Septier, F., Drumetz, L., & Pham, M.-T. (2022). Spherical sliced-wasserstein. *arXiv preprint arXiv:2206.08780.*

Bonnabel, S. (2013). Stochastic gradient descent on Riemannian manifolds. *IEEE Transactions on Automatic Control*, *58*(9), 2217–2229.

Caron, M., Misra, I., Mairal, J., Goyal, P., Bojanowski, P., & Joulin, A. (2020). Unsupervised learning of visual features by contrasting cluster assignments. *Advances in Neural Information Processing Systems*, *33*, 9912–9924.

Caron, M., Touvron, H., Misra, I., Jégou, H., Mairal, J., Bojanowski, P., & Joulin, A. (2021). Emerging properties in self-supervised vision transformers. *Proceedings of the IEEE/CVF International Conference on Computer Vision*, 9650–9660.

Chai, J., Zeng, H., Li, A., & Ngai, E. W. (2021). Deep learning in computer vision: A critical review of emerging techniques and application scenarios. *Machine Learning with Applications*, *6*, 100134.

Chen, T., Kornblith, S., Norouzi, M., & Hinton, G. (2020). A simple framework for contrastive learning of visual representations. *International Conference on Machine Learning*, 1597–1607.

Chen, Z., Zhu, L., Wan, L., Wang, S., Feng, W., & Heng, P.-A. (2020). A multi-task mean teacher for semi-supervised shadow detection. *Proceedings of the IEEE/CVF Conference on Computer Vision and Pattern Recognition*, 5611–5620.

Cheng, G., Cai, L., Lang, C., Yao, X., Chen, J., Guo, L., & Han, J. (2021). SPNet: Siamese-prototype network for few-shot remote sensing image scene classification. *IEEE Transactions on Geoscience and Remote Sensing*, *60*, 1–11.

Cheng, G., Han, J., & Lu, X. (2017). Remote sensing image scene classification: Benchmark and state of the art. *Proceedings of the IEEE*, *105*(10), 1865–1883.

Courtrai, L., Pham, M.-T., Burnel, J.-C., & Lefèvre, S. (2020). Apprentissage de réseaux de neurones de super-résolution pour la détection d'objets de petite taille dans les images de télédétection. *RFIAP 2020-Reconnaissance des Formes, Image, Apprentissage et Perception.*

Courtrai, L., Pham, M.-T., & Lefèvre, S. (2020). Small object detection in remote sensing images based on super-resolution with auxiliary generative adversarial networks. *Remote Sensing*, *12*(19), 3152.

Dalla Mura, M., Benediktsson, J. A., Waske, B., & Bruzzone, L. (2010). Morphological attribute profiles for the analysis of very high resolution images. *IEEE Transactions on Geoscience and Remote Sensing, 48*(10), 3747–3762.

Defard, T., Setkov, A., Loesch, A., & Audigier, R. (2021). PaDiM: A patch distribution modeling framework for anomaly detection and localization. *International Conference on Pattern Recognition*, 475–489.

Dehaene, D., Frigo, O., Combrexelle, S., & Eline, P. (2020). Iterative energy-based projection on a normal data manifold for anomaly localization. *arXiv preprint arXiv:2002.03734.*

Demir, B., & Bruzzone, L. (2015). Histogram-based attribute profiles for classification of very high resolution remote sensing images. *IEEE Transactions on Geoscience and Remote Sensing, 54*(4), 2096–2107.

Dimitrovski, I., Kitanovski, I., Kocev, D., & Simidjievski, N. (2023). Current trends in deep learning for earth observation: An open-source benchmark arena for image classification. *ISPRS Journal of Photogrammetry and Remote Sensing, 197*, 18–35.

Doersch, C., Gupta, A., & Efros, A. A. (2015). Unsupervised visual representation learning by context prediction. *Proceedings of the IEEE International Conference on Computer Vision*, 1422–1430.

Dong, X., & Shen, J. (2018). Triplet loss in siamese network for object tracking. *European Conference on Computer Vision*, 459–474.

Dosovitskiy, A., Beyer, L., Kolesnikov, A., Weissenborn, D., Zhai, X., Unterthiner, T., Dehghani, M., Minderer, M., Heigold, G., Gelly, S., et al. (2020). An image is worth 16x16 words: Transformers for image recognition at scale. *arXiv preprint arXiv:2010.11929.*

Dosovitskiy, A., Springenberg, J. T., Riedmiller, M., & Brox, T. (2014). Discriminative unsupervised feature learning with convolutional neural networks. *Advances in Neural Information Processing Systems, 27.*

Duong, A.-K., La, T.-V., Lê, H.-Â., & Pham, M.-T. (2024). Fedship: Federated learning for ship detection from multi-source satellite images. *IEEE Geoscience and Remote Sensing Letters.*

Duong, A.-K., Lê, H.-Â., & Pham, M.-T. (2024). Leveraging feature communication in federated learning for remote sensing image classification. *IGARSS 2024-2024 IEEE International Geoscience and Remote Sensing Symposium*, 6890–6894.

Ericsson, L., Gouk, H., & Hospedales, T. M. (2021). How well do self-supervised models transfer? *Proceedings of the IEEE/CVF Conference on Computer Vision and Pattern Recognition*, 5414–5423.

Ericsson, L., Gouk, H., Loy, C. C., & Hospedales, T. M. (2022). Self-supervised representation learning: Introduction, advances, and challenges. *IEEE Signal Processing Magazine, 39*(3), 42–62.

Everingham, M., Eslami, S. A., Van Gool, L., Williams, C. K., Winn, J., & Zisserman, A. (2015). The Pascal visual object classes challenge: A retrospective. *International Journal of Computer Vision, 111*, 98–136.

Fontana, M., Spratling, M., & Shi, M. (2024). When multitask learning meets partial supervision: A computer vision review. *Proceedings of the IEEE, 112*(6), 516–543.

Froidevaux, A., Julier, A., Lifschitz, A., Pham, M.-T., Dambreville, R., Lefèvre, S., Lassalle, P., & Huynh, T.-L. (2020). Vehicle detection and counting from vhr satellite images: Efforts and open issues. *IGARSS 2020-2020 IEEE International Geoscience and Remote Sensing Symposium*, 256–259.

Gangloff, H., Pham, M.-T., Courtrai, L., & Lefèvre, S. (2022a). Autoencodeurs variationnels à registre de vecteurs pour la détection d'anomalies. *RFIAP 2022-(Congrès Reconnaissance des Formes, Image, Apprentissage et Perception)*.

Gangloff, H., Pham, M.-T., Courtrai, L., & Lefèvre, S. (2022b). Leveraging vector-quantized variational autoencoder inner metrics for anomaly detection. *2022 26th International Conference on Pattern Recognition (ICPR)*, 435–441.

Gangloff, H., Pham, M.-T., Courtrai, L., & Lefèvre, S. (2023). Unsupervised anomaly detection using variational autoencoder with gaussian random field prior. *2023 IEEE International Conference on Image Processing (ICIP)*, 1620–1624.

Gangloff, H., Pham, M.-T., Courtrai, L., & Lefèvre, S. (2024). Variational autoencoder with gaussian random field prior: Application to unsupervised animal detection in aerial images. *ISPRS Journal of Photogrammetry and Remote Sensing*, *218*, 600–609.

Gidaris, S., Singh, P., & Komodakis, N. (2018). Unsupervised representation learning by predicting image rotations. *arXiv preprint arXiv:1803.07728*.

Grill, J.-B., Strub, F., Altché, F., Tallec, C., Richemond, P., Buchatskaya, E., Doersch, C., Avila Pires, B., Guo, Z., Gheshlaghi Azar, M., et al. (2020). Bootstrap your own latent-a new approach to self-supervised learning. *Advances in Neural Information Processing Systems*, *33*, 21271–21284.

Guiotte, F., Pham, M.-T., Dambreville, R., Corpetti, T., & Lefevre, S. (2020). Semantic segmentation of lidar points clouds: Rasterization beyond digital elevation models. *IEEE Geoscience and Remote Sensing Letters*, *17*(11), 2016–2019.

Guo, J., Han, K., Wang, Y., Wu, H., Chen, X., Xu, C., & Xu, C. (2021). Distilling object detectors via decoupled features. *Proceedings of the IEEE/CVF Conference on Computer Vision and Pattern Recognition*, 2154–2164.

Guo, X., Lao, J., Dang, B., Zhang, Y., Yu, L., Ru, L., Zhong, L., Huang, Z., Wu, K., Hu, D., et al. (2024). SkySense: a multi-modal remote sensing foundation model towards universal interpretation for Earth observation imagery. *Proceedings of the IEEE/CVF Conference on Computer Vision and Pattern Recognition*, 27672–27683.

Guo, Y., Wang, X., Chen, Y., & Yu, S. X. (2022). Clipped hyperbolic classifiers are super-hyperbolic classifiers. *Proceedings of the IEEE/CVF Conference on Computer Vision and Pattern Recognition*, 11–20.

Hamard, Q., Pham, M.-T., Cazau, D., & Heerah, K. (2024). A deep learning model for detecting and classifying multiple marine mammal species from passive acoustic data. *Ecological Informatics*, *84*, 102906.

Hamzaoui, M., Chapel, L., Pham, M.-T., & Lefevre, S. (2024). Hyperbolic prototypical network for few shot remote sensing scene classification. *Pattern Recognition Letters*, *177*, 151–156.

Hamzaoui, M., Chapel, L., Pham, M.-T., & Lefèvre, S. (2022). A hierarchical prototypical network for few-shot remote sensing scene classification. *International Conference on Pattern Recognition and Artificial Intelligence*, 208–220.

Hamzaoui, M., Chapel, L., Pham, M.-T., & Lefèvre, S. (2021). Hyperbolic variational auto-encoder for remote sensing scene classification. *ORASIS 2021*.

Hamzaoui, M., Chapel, L., Pham, M.-T., & Lefèvre, S. (2023). Hyperbolic variational auto-encoder for remote sensing scene embeddings. *IGARSS 2023-2023 IEEE International Geoscience and Remote Sensing Symposium*, 5391–5394.

Han, L., Tao, P., & Martin, R. R. (2019). Livestock detection in aerial images using a fully convolutional network. *Computational Visual Media*, 5(2), 221–228.

Hariharan, B., Arbeláez, P., Bourdev, L., Maji, S., & Malik, J. (2011). Semantic contours from inverse detectors. *International Conference on Computer Vision*, 991–998.

He, K., Chen, X., Xie, S., Li, Y., Dollár, P., & Girshick, R. (2022). Masked autoencoders are scalable vision learners. *Proceedings of the IEEE/CVF Conference on Computer Vision and Pattern Recognition*, 16000–16009.

He, K., Fan, H., Wu, Y., Xie, S., & Girshick, R. (2020). Momentum contrast for unsupervised visual representation learning. *Proceedings of the IEEE/CVF Conference on Computer Vision and Pattern Recognition*, 9729–9738.

Helber, P., Bischke, B., Dengel, A., & Borth, D. (2019). EuroSAT: a novel dataset and deep learning benchmark for land use and land cover classification. *IEEE Journal of Selected Topics in Applied Earth Observations and Remote Sensing*, 12(7), 2217–2226.

Higgins, I., Matthey, L., Pal, A., Burgess, C. P., Glorot, X., Botvinick, M. M., Mohamed, S., & Lerchner, A. (2017). Beta-VAE: learning basic visual concepts with a constrained variational framework. *International Conference on Learning Representations*, 3.

Hinton, G., Vinyals, O., & Dean, J. (2015). Distilling the knowledge in a neural network. *arXiv preprint arXiv:1503.02531*.

Hosseiny, B., Mahdianpari, M., Hemati, M., Radman, A., Mohammadimanesh, F., & Chanussot, J. (2023). Beyond supervised learning in remote sensing: A systematic review of deep learning approaches. *IEEE Journal of Selected Topics in Applied Earth Observations and Remote Sensing*, 17, 1035–1052.

Huang, C., Tang, H., Fan, W., Xiao, Y., Hao, D., Qian, Z., Terzopoulos, D., et al. (2020). Partly supervised multi-task learning. *2020 19th IEEE International Conference on Machine Learning and Applications (ICMLA)*, 769–774.

Jazbec, M., Pearce, M., & Fortuin, V. (2020). Factorized Gaussian process variational autoencoders. *arXiv preprint arXiv:2011.07255*.

Jing, L., & Tian, Y. (2020). Self-supervised visual feature learning with deep neural networks: A survey. *IEEE Transactions on Pattern Analysis and Machine Intelligence*, 43(11), 4037–4058.

Khosla, P., Teterwak, P., Wang, C., Sarna, A., Tian, Y., Isola, P., Maschinot, A., Liu, C., & Krishnan, D. (2020). Supervised contrastive learning. *Advances in Neural Information Processing Systems*, 33, 18661–18673.

Khrulkov, V., Mirvakhabova, L., Ustinova, E., Oseledets, I., & Lempitsky, V. (2020). Hyperbolic image embeddings. *Proceedings of the IEEE/CVF Conference on Computer Vision and Pattern Recognition*, 6418–6428.

Kim, J.-H., Kim, D.-H., Yi, S., & Lee, T. (2021). Semi-orthogonal embedding for efficient unsupervised anomaly segmentation. *arXiv preprint arXiv:2105.14737*.

Kingma, D. P., & Welling, M. (2019). An introduction to variational autoencoders. *Foundations and Trends® in Machine Learning*, *12*(4), 307–392.

Krähenbühl, P., & Koltun, V. (2011). Efficient inference in fully connected CRFs with Gaussian edge potentials. *Advances in Neural Information Processing Systems*, *24*.

Kulharia, V., Chandra, S., Agrawal, A., Torr, P., & Tyagi, A. (2020). Box2seg: Attention weighted loss and discriminative feature learning for weakly supervised segmentation. *European Conference on Computer Vision*, 290–308.

La, T.-V., Pham, M.-T., & Chini, M. (2024a). Collocation of multi-source satellite imagery for ship detection based on deep learning models. *EGU General Assembly Conference Abstracts*, 3954.

La, T.-V., Pham, M.-T., & Chini, M. (2024b). Insight into the collocation of multi-source satellite imagery for multi-scale vessel detection. *IGARSS 2024-2024 IEEE International Geoscience and Remote Sensing Symposium*, 7396–7400.

Le, H.-A., Guiotte, F., Pham, M.-T., Lefevre, S., & Corpetti, T. (2022). Learning digital terrain models from point clouds: Als2dtm dataset and rasterization-based gan. *IEEE Journal of Selected Topics in Applied Earth Observations and Remote Sensing*, *15*, 4980–4989.

Lê, H.-Â., Berg, P., & Pham, M.-T. (2024). Box for mask and mask for box: Weak losses for multi-task partially supervised learning. *arXiv preprint arXiv:2411.17536*.

Lê, H.-Â., & Pham, M.-T. (2023a). Data exploitation: Multi-task learning of object detection and semantic segmentation on partially annotated data. *arXiv preprint arXiv:2311.04040*.

Lê, H.-Â., & Pham, M.-T. (2023b). Knowledge distillation for object detection: From generic to remote sensing datasets. *IGARSS 2023-2023 IEEE International Geoscience and Remote Sensing Symposium*, 6194–6197.

Lê, H.-Â., & Pham, M.-T. (2024). Leveraging knowledge distillation for partial multi-task learning from multiple remote sensing datasets. *IGARSS 2024-2024 IEEE International Geoscience and Remote Sensing Symposium*, 8019–8023.

Lê, H.-Â., & Pham, M.-T. (2023c). Self-training and multi-task learning for limited data: Evaluation study on object detection. *Proceedings of the IEEE/CVF International Conference on Computer Vision*, 1003–1009.

Lê, H.-Â., Zhang, H., Pham, M.-T., & Lefèvre, S. (2022). Mutual guidance meets supervised contrastive learning: Vehicle detection in remote sensing images. *Remote Sensing*, *14*(15), 3689.

Lefèvre, S., Courtrai, L., Pham, M.-T., Friguet, C., & Burnel, J.-C. (2021). Observation de la mer par apprentissage profond: Quelques exemples d'applications pour protéger notre bien commun. *Mer et Littoral: un bien commun? Une approche pluridisciplinaire. Actes du Séminaire International de l'Institut Archipel. Université Bretagne Sud, 17-19 juin 2019.*, 215–224.

Li, W.-H., Liu, X., & Bilen, H. (2022). Learning multiple dense prediction tasks from partially annotated data. *Proceedings of the IEEE/CVF Conference on Computer Vision and Pattern Recognition*, 18879–18889.

Li, X., Li, H., Yu, R., & Wang, F. (2021). Few-shot scene classification with attention mechanism in remote sensing. *Journal of Physics: Conference Series*, *1961*(1), 012015.

Liang, P. P., Zadeh, A., & Morency, L.-P. (2024). Foundations & trends in multimodal machine learning: Principles, challenges, and open questions. *ACM Computing Surveys*, *56*(10), 1–42.

Lin, T.-Y., Goyal, P., Girshick, R., He, K., & Dollár, P. (2017). Focal loss for dense object detection. *Proceedings of the IEEE International Conference on Computer Vision*, 2980–2988.

Lin, T.-Y., Maire, M., Belongie, S., Hays, J., Perona, P., Ramanan, D., Dollár, P., & Zitnick, C. L. (2014). Microsoft COCO: Common objects in context. *European Conference on Computer Vision*, 740–755.

Liu, W., Li, R., Zheng, M., Karanam, S., Wu, Z., Bhanu, B., Radke, R. J., & Camps, O. (2020). Towards visually explaining variational autoencoders. *Proceedings of the IEEE/CVF Conference on Computer Vision and Pattern Recognition*, 8642–8651.

Liu, Y., Li, J., Sun, S., & Yu, B. (2019). Advances in Gaussian random field generation: A review. *Computational Geosciences*, *23*, 1011–1047.

Liu, Z., Hu, H., Lin, Y., Yao, Z., Xie, Z., Wei, Y., Ning, J., Cao, Y., Zhang, Z., Dong, L., et al. (2022). Swin transformer v2: Scaling up capacity and resolution. *Proceedings of the IEEE/CVF Conference on Computer Vision and Pattern Recognition*, 12009–12019.

Liznerski, P., Ruff, L., Vandermeulen, R. A., Franks, B. J., Kloft, M., & Müller, K.-R. (2020). Explainable deep one-class classification. *arXiv preprint arXiv:2007.01760*.

Loaiza-Ganem, G., & Cunningham, J. P. (2019). The continuous Bernoulli: Fixing a pervasive error in variational autoencoders. *Advances in Neural Information Processing Systems*, *32*.

Lu, S., Guo, J., Zimmer-Dauphinee, J. R., Nieusma, J. M., Wang, X., VanValkenburgh, P., Wernke, S. A., & Huo, Y. (2024). AI foundation models in remote sensing: A survey. *arXiv preprint arXiv:2408.03464*.

Maia, D. S., Pham, M.-T., Aptoula, E., Guiotte, F., & Lefèvre, S. (2021). Classification of remote sensing data with morphological attribute profiles: A decade of advances. *IEEE geoscience and remote sensing magazine*, *9*(3), 43–71.

Maia, D. S., Pham, M.-T., & Lefèvre, S. (2022). Watershed-based attribute profiles with semantic prior knowledge for remote sensing image analysis. *IEEE Journal of Selected Topics in Applied Earth Observations and Remote Sensing*, *15*, 2574–2591.

Merciol, F., Pham, M.-T., Santana Maia, D., Masse, A., & Sannier, C. (2020). Broceliande: A comparative study of attribute profiles and feature profiles from different attributes. *The International Archives of the Photogrammetry, Remote Sensing and Spatial Information Sciences*, *43*, 1371–1377.

Miller, G. A. (1998). *WordNet: An electronic lexical database*.

Mirzadeh, S. I., Farajtabar, M., Li, A., Levine, N., Matsukawa, A., & Ghasemzadeh, H. (2020). Improved knowledge distillation via teacher assistant. *Proceedings of the AAAI Conference on Artificial Intelligence*, *34*(04), 5191–5198.

Narvaez Luces, O. D. R., Pham, M.-T., Poterek, Q., & Braun, R. (2023). Burnt area extraction from high-resolution satellite images based on anomaly detection. *Joint European Conference on Machine Learning and Knowledge Discovery in Databases*, 448–457.

Nguyen, H. H., Nguyen, C. N., Dao, X. T., Duong, Q. T., Kim, D. P. T., & Pham, M.-T. (2024). Variational autoencoder for anomaly detection: A comparative study. *arXiv preprint arXiv:2408.13561*.

Nickel, M., & Kiela, D. (2017). Poincaré embeddings for learning hierarchical representations. *Advances in Neural Information Processing Systems*, *30*.

Noroozi, M., & Favaro, P. (2016). Unsupervised learning of visual representations by solving jigsaw puzzles. *European Conference on Computer Vision*, 69–84.

Ohri, K., & Kumar, M. (2021). Review on self-supervised image recognition using deep neural networks. *Knowledge-Based Systems*, *224*, 107090.

Osio, A., Pham, M.-T., & Lefèvre, S. (2020). Spatial processing of sentinel imagery for monitoring of acacia forest degradation in lake nakuru riparian reserve. *ISPRS Annals of the Photogrammetry, Remote Sensing and Spatial Information Sciences*, *3*, 525–532.

Pande, S., Uzun, B., Guiotte, F., Pham, M.-T., Corpetti, T., Delerue, F., & Lefèvre, S. (2024). Plant detection from ultra high resolution remote sensing images: A semantic segmentation approach based on fuzzy loss. *IGARSS 2024-2024 IEEE International Geoscience and Remote Sensing Symposium*, 5213–5217.

Pang, J., Chen, K., Shi, J., Feng, H., Ouyang, W., & Lin, D. (2019). Libra R-CNN: Towards balanced learning for object detection. *Proceedings of the IEEE/CVF Conference on Computer Vision and Pattern Recognition*, 821–830.

Peng, W., Varanka, T., Mostafa, A., Shi, H., & Zhao, G. (2021). Hyperbolic deep neural networks: A survey. *IEEE Transactions on Pattern Analysis and Machine Intelligence*, *44*(12), 10023–10044.

Persello, C., Wegner, J. D., Hänsch, R., Tuia, D., Ghamisi, P., Koeva, M., & Camps-Valls, G. (2022). Deep learning and Earth observation to support the sustainable development goals: Current approaches, open challenges, and future opportunities. *IEEE Geoscience and Remote Sensing Magazine*, *10*(2), 172–200.

Pham, M. T. (2016). *Pointwise approach for texture analysis and characterization from very high resolution remote sensing images* (Doctoral dissertation). Ecole Nationale Supérieure des Télécommunications de Bretagne-ENSTB.

Pham, M. T., Mercier, G., & Michel, J. (2014). Textural features from wavelets on graphs for very high resolution panchromatic pléiades image classification. *Revue française de photogrammétrie et de télédétection*, (208), 131–136.

Pham, M. T., & Gueriot, D. (2013). Guided block-matching for sonar image registration using unsupervised kohonen neural networks. *2013 OCEANS-San Diego*, 1–5.

Pham, M.-T. (2018a). Efficient texture retrieval using multiscale local extrema descriptors and covariance embedding. *Proceedings of the European Conference on Computer Vision (ECCV) Workshops*, 0–0.

Pham, M.-T. (2018b). Fusion of polarimetric features and structural gradient tensors for vhr polsar image classification. *IEEE Journal of Selected Topics in Applied Earth Observations and Remote Sensing*, *11*(10), 3732–3742.

Pham, M.-T., Aptoula, E., & Lefèvre, S. (2018). Classification of remote sensing images using attribute profiles and feature profiles from different trees: A comparative study. *IGARSS 2018-2018 IEEE International Geoscience and Remote Sensing Symposium*, 4511–4514.

Pham, M.-T., Courtrai, L., Friguet, C., Lefèvre, S., & Baussard, A. (2020). Yolo-fine: One-stage detector of small objects under various backgrounds in remote sensing images. *Remote Sensing*, *12*(15), 2501.

Pham, M.-T., Gangloff, H., & Lefèvre, S. (2023). Weakly supervised marine animal detection from remote sensing images using vector-quantized variational autoencoder. *IGARSS 2023-2023 IEEE International Geoscience and Remote Sensing Symposium*, 5559–5562.

Pham, M.-T., & Lefevre, S. (2021). Very high resolution airborne polsar image classification using convolutional neural networks. *EUSAR 2021; 13th European Conference on Synthetic Aperture Radar*, 1–4.

Pham, M.-T., Lefevre, S., & Aptoula, E. (2017). Local feature-based attribute profiles for optical remote sensing image classification. *IEEE Transactions on Geoscience and Remote Sensing*, *56*(2), 1199–1212.

Pham, M.-T., & Lefèvre, S. (2018a). Buried object detection from b-scan ground penetrating radar data using faster-rcnn. *IGARSS 2018-2018 IEEE international geoscience and remote sensing symposium*, 6804–6807.

Pham, M.-T., & Lefèvre, S. (2018b). Détection d'objets enterrés par apprentissage profond sur imagerie géoradar. *RFIAP*.

Pham, M.-T., Lefèvre, S., Aptoula, E., & Bruzzone, L. (2018). Recent developments from attribute profiles for remote sensing image classification. *arXiv preprint arXiv:1803.10036*.

Pham, M.-T., Lefèvre, S., Aptoula, E., & Damodaran, B. B. (2017). Classification of vhr remote sensing images using local feature-based attribute profiles. *2017 IEEE International Geoscience and Remote Sensing Symposium (IGARSS)*, 747–750.

Pham, M.-T., Lefèvre, S., & Merciol, F. (2018). Attribute profiles on derived textural features for highly textured optical image classification. *IEEE Geoscience and Remote Sensing Letters*, *15*(7), 1125–1129.

Pham, M.-T., & Mercier, G. (2021). Graph of characteristic points for texture tracking: Application to change detection and glacier flow measurement from sar images. *Change Detection and Image Time Series Analysis 1: Unsupervised Methods*, 167–200.

Pham, M.-T., Mercier, G., & Michel, J. (2016). Pw-cog: An effective texture descriptor for vhr satellite imagery using a pointwise approach on covariance matrix of oriented gradients. *IEEE Transactions on Geoscience and Remote Sensing*, *54*(6), 3345–3359.

Pham, M.-T., Mercier, G., & Bombrun, L. (2017). Color texture image retrieval based on local extrema features and riemannian distance. *Journal of Imaging*, *3*(4), 43.

Pham, M.-T., Mercier, G., & Michel, J. (2015a). Change detection between sar images using a pointwise approach and graph theory. *IEEE Transactions on Geoscience and Remote Sensing*, *54*(4), 2020–2032.

Pham, M.-T., Mercier, G., & Michel, J. (2015b). Covariance-based texture description from weighted coherency matrix and gradient tensors for polarimetric sar image classification. *2015 IEEE International Geoscience and Remote Sensing Symposium (IGARSS)*, 2469–2472.

Pham, M.-T., Mercier, G., & Michel, J. (2015c). A keypoint approach for change detection between sar images based on graph theory. *2015 8th International Workshop on the Analysis of Multitemporal Remote Sensing Images (Multi-Temp)*, 1–4.

Pham, M.-T., Mercier, G., & Michel, J. (2015d). Pointwise graph-based local texture characterization for very high resolution multispectral image classification. *IEEE Journal of Selected Topics in Applied Earth Observations and Remote Sensing*, *8*(5), 1962–1973.

Pham, M.-T., Mercier, G., & Michel, J. (2014). Wavelets on graphs for very high resolution multispectral image texture segmentation. *2014 IEEE Geoscience and Remote Sensing Symposium*, 2273–2276.

Pham, M.-T., Mercier, G., Regniers, O., & Michel, J. (2016). Texture retrieval from vhr optical remote sensed images using the local extrema descriptor with application to vineyard parcel detection. *Remote Sensing*, *8*(5), 368.

Pham, M.-T., Mercier, G., Trouvé, E., & Lefèvre, S. (2017). Sar image texture tracking using a pointwise graph-based model for glacier displacement measurement. *2017 IEEE International Geoscience and Remote Sensing Symposium (IGARSS)*, 1083–1086.

Pham, M., Aptoula, E., & Lefevre, S. (2018). Feature profiles from attribute filtering for remote sensing image classification. *IEEE J. Sel. Topics Appl. Earth Observations Remote Sens*, *11*(1), 249–256.

Pirrone, D., & Pham, M.-T. (2020). A compound polarimetric-textural approach for unsupervised change detection in multi-temporal full-pol sar imagery. *IGARSS 2020-2020 IEEE International Geoscience and Remote Sensing Symposium*, 316–319.

Rolf, E., Klemmer, K., Robinson, C., & Kerner, H. (2024). Position: Mission critical–satellite data is a distinct modality in machine learning. *Forty-first International Conference on Machine Learning*.

Rother, C., Kolmogorov, V., & Blake, A. (2004). "GrabCut" interactive foreground extraction using iterated graph cuts. *ACM Transactions on Graphics (TOG)*, *23*(3), 309–314.

Rottensteiner, F., Sohn, G., Jung, J., Gerke, M., Baillard, C., Benitez, S., & Breitkopf, U. (2012). The ISPRS benchmark on urban object classification and 3D building reconstruction. *ISPRS Annals of the Photogrammetry, Remote Sensing and Spatial Information Sciences; I-3*, *1*(1), 293–298.

Santana Maia, D., Pham, M.-T., & Lefèvre, S. (2021). Watershed-based attribute profiles for pixel classification of remote sensing data. *International Conference on Discrete Geometry and Mathematical Morphology*, 120–133.

Sarker, I. H. (2021). Deep learning: A comprehensive overview on techniques, taxonomy, applications and research directions. *SN Computer Science*, *2*(6), 420.

Scheibenreif, L., Hanna, J., Mommert, M., & Borth, D. (2022). Self-supervised vision transformers for land-cover segmentation and classification. *Proceedings of the IEEE/CVF Conference on Computer Vision and Pattern Recognition*, 1422–1431.

Schmitt, M., Ahmadi, S. A., Xu, Y., Taşkin, G., Verma, U., Sica, F., & Hänsch, R. (2023). There are no data like more data: Datasets for deep learning in earth observation. *IEEE Geoscience and Remote Sensing Magazine*, *11*(3), 63–97.

Schmitt, M., Hughes, L. H., Qiu, C., & Zhu, X. X. (2019). SEN12MS-A curated dataset of georeferenced multi-spectral sentinel-1/2 imagery for deep learning and data fusion. *arXiv preprint arXiv:1906.07789*.

Singh, T., Gangloff, H., & Pham, M.-T. (2023). Object counting from aerial remote sensing images: Application to wildlife and marine mammals. *IGARSS 2023-2023 IEEE International Geoscience and Remote Sensing Symposium*, 6580–6583.

Snell, J., Swersky, K., & Zemel, R. (2017). Prototypical networks for few-shot learning. *Advances in Neural Information Processing Systems*, *30*.

Song, C., Ouyang, W., & Zhang, Z. (2023). Weakly supervised semantic segmentation via box-driven masking and filling rate shifting. *IEEE Transactions on Pattern Analysis and Machine Intelligence*, *45*(12), 15996–16012.

Sun, X., Wang, B., Wang, Z., Li, H., Li, H., & Fu, K. (2021). Research progress on few-shot learning for remote sensing image interpretation. *IEEE Journal of Selected Topics in Applied Earth Observations and Remote Sensing*, *14*, 2387–2402.

Tuia, D., Schindler, K., Demir, B., Zhu, X. X., Kochupillai, M., Džeroski, S., van Rijn, J. N., Hoos, H. H., Del Frate, F., Datcu, M., et al. (2024). Artificial intelligence to advance Earth observation: A review of models, recent trends, and pathways forward. *IEEE Geoscience and Remote Sensing Magazine*.

Uzun, B., Pande, S., Cachin-Bernard, G., Pham, M.-T., Lefèvre, S., Blatrix, R., & Mckey, D. (2024). Mapping earth mounds from space. *IGARSS 2024-2024 IEEE International Geoscience and Remote Sensing Symposium*, 4044–4049.

Van Den Oord, A., & Vinyals, O. (2017). Neural discrete representation learning. *Advances in Neural Information Processing Systems*, *30*.

Vandenhende, S., Georgoulis, S., Van Gansbeke, W., Proesmans, M., Dai, D., & Van Gool, L. (2021). Multi-task learning for dense prediction tasks: A survey. *IEEE Transactions on Pattern Analysis and Machine Intelligence*, *44*(7), 3614–3633.

Venkataramanan, S., Peng, K.-C., Singh, R. V., & Mahalanobis, A. (2020). Attention guided anomaly localization in images. *European Conference on Computer Vision*, 485–503.

Vinyals, O., Blundell, C., Lillicrap, T., Wierstra, D., et al. (2016). Matching networks for one shot learning. *Advances in Neural Information Processing Systems*, *29*.

Wang, L., Zhang, D., Guo, J., & Han, Y. (2020). Image anomaly detection using normal data only by latent space resampling. *Applied Sciences*, *10*(23), 8660.

Wang, T., & Isola, P. (2020). Understanding contrastive representation learning through alignment and uniformity on the hypersphere. *International Conference on Machine Learning*, 9929–9939.

Wang, X., Zhang, R., Shen, C., Kong, T., & Li, L. (2021). Dense contrastive learning for self-supervised visual pre-training. *Proceedings of the IEEE/CVF Conference on Computer Vision and Pattern Recognition*, 3024–3033.

Wang, Y., Albrecht, C. M., Braham, N. A. A., Mou, L., & Zhu, X. X. (2022). Self-supervised learning in remote sensing: A review. *IEEE Geoscience and Remote Sensing Magazine*, *10*(4), 213–247.

Wang, Z., Bovik, A. C., Sheikh, H. R., & Simoncelli, E. P. (2004). Image quality assessment: From error visibility to structural similarity. *IEEE transactions on Image Processing*, *13*(4), 600–612.

Xiao, A., Xuan, W., Wang, J., Huang, J., Tao, D., Lu, S., & Yokoya, N. (2024). Foundation models for remote sensing and Earth observation: A survey. *arXiv preprint arXiv:2410.16602*.

Xie, Z., Zhang, Z., Cao, Y., Lin, Y., Bao, J., Yao, Z., Dai, Q., & Hu, H. (2022). SimMIM: A simple framework for masked image modeling. *Proceedings of the IEEE/CVF Conference on Computer Vision and Pattern Recognition*, 9653–9663.

Xu, G., Liu, Z., Li, X., & Loy, C. C. (2020). Knowledge distillation meets self-supervision. *European Conference on Computer Vision*, 588–604.

Yin, X., Zhu, Y., & Hu, J. (2021). A comprehensive survey of privacy-preserving federated learning: A taxonomy, review, and future directions. *ACM Computing Surveys (CSUR)*, *54*(6), 1–36.

Yokoya, N., Ghamisi, P., Hänsch, R., & Schmitt, M. (2020). 2020 IEEE GRSS Data Fusion Contest: Global land cover mapping with weak supervision [technical committees]. *IEEE Geoscience and Remote Sensing Magazine*, *8*(1), 154–157.

Zbontar, J., Jing, L., Misra, I., LeCun, Y., & Deny, S. (2021). Barlow twins: Self-supervised learning via redundancy reduction. *International Conference on Machine Learning*, 12310–12320.

Zhang, H., Fromont, E., Lefevre, S., & Avignon, B. (2021). PDF-Distil: including prediction disagreements in feature-based distillation for object detection. *BMVC 2021-32nd British Machine Vision Conference*, 1–13.

Zhang, J., Huang, J., Jin, S., & Lu, S. (2024). Vision-language models for vision tasks: A survey. *IEEE Transactions on Pattern Analysis and Machine Intelligence*, *46*(8), 5625–5644.

Zhang, P., Bai, Y., Wang, D., Bai, B., & Li, Y. (2020). Few-shot classification of aerial scene images via meta-learning. *Remote Sensing*, *13*(1), 108.

Zhang, P., Fan, G., Wu, C., Wang, D., & Li, Y. (2021). Task-adaptive embedding learning with dynamic kernel fusion for few-shot remote sensing scene classification. *Remote Sensing*, *13*(21), 4200.

Zhang, X., Li, N., Li, J., Dai, T., Jiang, Y., & Xia, S.-T. (2023). Unsupervised surface anomaly detection with diffusion probabilistic model. *Proceedings of the IEEE/CVF International Conference on Computer Vision*, 6782–6791.

Zhu, B., Lui, N., Irvin, J., Le, J., Tadwalkar, S., Wang, C., Ouyang, Z., Liu, F. Y., Ng, A. Y., & Jackson, R. B. (2022). Meter-ML: A multi-sensor earth observation benchmark for automated methane source mapping. *arXiv preprint arXiv:2207.11166*.

Zhu, X. X., Tuia, D., Mou, L., Xia, G.-S., Zhang, L., Xu, F., & Fraundorfer, F. (2017). Deep learning in remote sensing: A comprehensive review and list of resources. *IEEE Geoscience and Remote Sensing Magazine*, *5*(4), 8–36.

Zimmerer, D., Isensee, F., Petersen, J., Kohl, S., & Maier-Hein, K. (2019). Unsupervised anomaly localization using variational auto-encoders. *International Conference on Medical Image Computing and Computer-Assisted Intervention (MICCAI)*, 289–297.